\documentclass[10pt,twocolumn,letterpaper]{article}
\usepackage{iccv_format/iccv}
\usepackage[pagebackref=true,breaklinks=true,letterpaper=true,colorlinks,bookmarks=false]{hyperref}
\usepackage{times}
\usepackage{epsfig}
\usepackage{graphicx}
\usepackage{amsmath,amssymb}
\usepackage{float}
\usepackage{caption}
\usepackage{subcaption}
\usepackage{mathtools}
\usepackage{colortbl}
\definecolor{Black}{rgb}{0,0,0}
\usepackage{adjustbox}

\usepackage{pifont}
\newcommand{\cmark}{\ding{51}}

\usepackage{csvsimple}
\usepackage{tabularx}
\usepackage{pgfplotstable,filecontents,booktabs}

\def\iccvPaperID{1404} 

\iccvfinalcopy
\ificcvfinal\fi

\usepackage[round-mode=places, round-integer-to-decimal, round-precision=4,
    table-format = 1.4, 
    table-number-alignment=center,
    round-integer-to-decimal]{siunitx}

\begin{document}

\title{SID4VAM: A Benchmark Dataset with Synthetic Images \\ for Visual Attention Modeling}

\author{David Berga\textsuperscript{1}
\and Xos\'e R. Fdez-Vidal\textsuperscript{2}
\and Xavier Otazu\textsuperscript{1}
\and Xos\'e M. Pardo\textsuperscript{2}
\and \textsuperscript{1}Computer Vision Center, Universitat Aut\`onoma de Barcelona, Spain\\ 
\textsuperscript{2}CiTIUS, Universidade de Santiago de Compostela, Spain\\ 
{\tt\small \{dberga,xotazu\}@cvc.uab.es} \hspace{1cm} {\tt\small \{xose.vidal,xose.pardo\}@usc.es}
}

\maketitle

\ificcvfinal\thispagestyle{empty}\fi

\begin{abstract}
A benchmark of saliency models performance with a synthetic image dataset is provided. Model performance is evaluated through saliency metrics as well as the influence of model inspiration and consistency with human psychophysics. SID4VAM is composed of 230 synthetic images, with known salient regions. Images were generated with 15 distinct types of low-level features (e.g. orientation, brightness, color, size...) with a target-distractor pop-out type of synthetic patterns. We have used Free-Viewing and Visual Search task instructions and 7 feature contrasts for each feature category. Our study reveals that state-of-the-art Deep Learning saliency models do not perform well with synthetic pattern images, instead, models with Spectral/Fourier inspiration outperform others in saliency metrics and are more consistent with human psychophysical experimentation. This study proposes a new way to evaluate saliency models in the forthcoming literature, accounting for synthetic images with uniquely low-level feature contexts, distinct from previous eye tracking image datasets.
\end{abstract}

\section{Introduction}

Although eye movements are indicators of  \textit{``where people look at"}, a more complex question arises as a consequence for understanding bottom-up visual attention: \textit{Are all eye movements equally valuable for determining saliency?} According to the initial hypotheses in visual attention \cite{Treisman1980,Wolfe2010a}, we could define visual saliency as the perceptual quality that makes our human visual system (HVS) to gaze towards certain areas that pop-out on a scene due to their distinctive visual characteristics. Therefore, this capacity (saliency) cannot be influenced by top-down factors, which seemingly guide eye movements regardless of stimulus characteristics \cite{Yantis1999}. Accounting for prior knowledge of whether a stimulus area is salient or not, when it becomes salient, and why, are issues that need to be accounted for saliency evaluation \cite{Bruce2015,Berga2018a}.

Common frameworks for predicting saliency have been created since Koch \& Ullman's seminal work \cite{Koch1987}. This framework defined a theoretical basis for modeling the early visual stages of the HVS in order to obtain a representation of the saliency map. By extracting sensory signals as feature maps, processing the conspicuous objects and selecting the maximally-active locations through winner-take-all (WTA) mechanisms, it is possible to obtain a unique/master saliency map. However, it was hypothesized that visual attention combines both bottom-up (saliency) and top-down (relevance) mechanisms in a central representation (priority) \cite{Egeth1997,FECTEAU2006}. These top-down specificities (e.g. world, object, task, etc.) were later accounted in the selective tuning model as a hierarchy of WTA-like processes \cite{Tsotsos1995}. Despite the neural correlates simultaneously involved in saliency have been investigated \cite{Veale2017}, the direct relation between saliency and eye movements defined in a unique computational framework requires further study. Itti et al. initially introduced a computational biologically-inspired model \cite{Itti1998} composed of 3 main steps: First, feature maps are extracted using oriented linear DoG filters for each chromatic channel. Second, feature conspicuity is computed using center-surround differences. Third, conspicuity maps are integrated with linear WTA mechanisms. This architecture has been the main inspiration for current saliency models \cite{Zhang2013a,Riche2016a}, that alternatively use distinct mechanisms (accounting for different levels of processing, context or tuning depending on the scene) but preserving same or similar structure for these steps. Although current state-of-the-art models precisely resemble eye-tracking fixation data \cite{Borji2013b,Bylinskii2015}, we question if these models represent saliency. We will test this hypothesis with a novel synthetic image dataset. 


\subsection{Related Work} 


In order to determine whether an object or a feature attracts attention, initial experimentation was assessing feature discriminability upon display characteristics (e.g. display size, feature contrast...) during visual search tasks \cite{Treisman1980,Wolfe2010a}. Parallel search occurs when features are processed preattentively, therefore search targets are found efficiently regardless of distractor properties. Instead, serial search happens when attention is directed to one item at a time, requiring a ``binding" process to allow each object to be discriminated. For this case, search time decrease with higher target-distractor contrast and/or lower set size (following the Weber Law \cite{fechner1966}). More recent studies replicated these experiments by providing real images with parametrization of feature contrast and/or set size (iLab USC, UCL, VAL Hardvard, ADA KCL), combining visual search or visual segmentation tasks, however not providing eye tracking data (\hyperref[fig:datasets1]{Table \ref*{fig:datasets1}B}). Rather, current eye movement datasets provide fixations and scanpaths from real scenes during free-viewing tasks. These image datasets are usually composed of real image scenes (\hyperref[fig:datasets1]{Table \ref*{fig:datasets1}A}), either from indoor / outdoor scenes (Toronto, MIT1003, MIT300), nature scenes (KTH) or semantically-specific categories such as faces (NUSEF) and several others (CAT2000). A complete list of eye tracking datasets is in Winkler \& Subramanian's overview \cite{Winkler2013}. CAT2000 training subset of ``Pattern" images (CAT2000$_p$) provides eye movement data with psychophysical / synthetic image patterns during 5 sec of free-viewing. However, no parametrization of feature contrast nor stimulus properties is given. A synthetic image dataset could provide information of how attention is dependent on feature contrast and other stimulus properties with distinct tasks. We describe in \hyperref[sec:dataset]{Section \ref*{sec:dataset}} how we do so with our novel SID4VAM's dataset.

\begin{table}[h!]
\scriptsize
\centering
\caption{Characteristics of eye tracking datasets}
\subcaption*{A: Real Images} \label{fig:datasets1}
\begin{tabular}{ |c|c|c|c|c|c| } 
\hline
Dataset & Task & \# TS & \# PP & PM & DO\\ 
\hline
Toronto \cite{Bruce2005} & FV & 120 & 20 &   & \cmark\\ 
MIT1003 \cite{Judd2009} & FV & 1003 & 15 &    & \cmark \\
NUSEF \cite{Ramanathan2010} & FV & 758 & 25 &   & \cmark \\ 
KTH \cite{Kootstra2011} & FV & 99 & 31 &   & \cmark \\ 
MIT300 \cite{Judd2012} & FV & 300 & 39 &    & \cmark \\ 
CAT2000 \cite{CAT2000} & FV & 4000 & 24 &   & \cmark \\ 
\hline
\end{tabular}
\bigskip
\subcaption*{B: Psychophysical Pattern / Synthetic Images} \label{fig:datasets2}
\begin{tabular}{ |c|c|c|c|c|c| } 
\hline
Dataset & Task & \# TS & \# PP & PM & DO\\ 
\hline
iLab USC \cite{Itti2000} & - & \textasciitilde 540 & - & \cmark &   \\ 
UCL \cite{Zhaoping2007} & VS \& SG & 2784 & 5 & \cmark &  \\
VAL Harvard \cite{Wolfe2010b} & VS & 4000 & 30 & \cmark &  \\ 
ADA KCL \cite{Spratling2012} & - & \textasciitilde 430 & - & \cmark &  \\ 
CAT2000$_p$ \cite{CAT2000} & FV & 100 & 18 &   & \cmark\\ 
SID4VAM (Ours) & FV \& VS & 230 & 34 & \cmark & \cmark\\ 
\hline
\end{tabular}
\\\hspace{2mm}\small TS: total number of stimuli, PP: participants, PM: Parametrization, DO: Fixation data is available online, FV: Free-Viewing, VS: Visual Search, SG: visual segmentation
\label{table:datasets}
\end{table}

Being inspired by Itti et al's architecture, a myiriad of computational models has been proposed with distinct computational approaches, from biological, mathematical and physical inspiration \cite{Zhang2013a,Riche2016a}. By processing global and/or local image fatures for calculating feature conspicuity, these models are able to generate a master saliency map to predict human fixations (\hyperref[table:models]{Table \ref*{table:models}}). Taking up Judd et al. \cite{Judd2012} and Borji \& Borji's \cite{Borji2013c} reviews, we have grouped saliency model inspiration in five general categories according to its saliency computation endeavour:
\begin{itemize}
    \item Cognitive/Biological (C): Saliency is usually generated by mimicking HVS neuronal mechanisms or either specific patterns found in human eye movement behavior. Feature extraction is generally based on Gabor-like filters and its integration with WTA-like mechanisms.
    \item Information-Theoretic (I): These models compute saliency by selecting the regions that maximize visual information of scenes.
    \item Probabilistic (P): Probabilistic models generate saliency by optimizing the probability of performing certain tasks and/or finding certain patterns. These models use graphs, bayesian, decision-theoretic and other  approaches for their computations.
    \item Spectral/Fourier-based (F): Spectral Analysis or Fourier-based models derive saliency by extracting or manipulating features in the frequency domain (e.g. spectral frequency or phase).
    \item Machine/Deep Learning (D): These techniques are based on training existing machine/deep learning architectures (e.g. CNN, RNN, GAN...) by minimizing the error of predicting fixations of images from existing eye tracking data or labeled salient regions.
\end{itemize}

\begin{table}[h!]
\centering
\scriptsize
\setlength{\tabcolsep}{5pt}
\caption{Description of saliency models}
\begin{tabular}{ |c|c|c|ccccc|cc| } 
\hline
Model & Authors & Year & \multicolumn{5}{ c| }{Inspiration} & \multicolumn{2}{ c| }{Type}\\ 
 & & & \cellcolor{red!20} C & \cellcolor{green!20} I & \cellcolor{blue!20} P & \cellcolor{orange!20} F & \cellcolor{cyan!20} D & \cellcolor{white} G & L \\
\hline
IKN & Itti et al.\cite{Itti1998,Itti2000} & 1998 & \cmark & & & & & \cmark & \cmark \\
AIM & Bruce \& Tsotsos \cite{Bruce2005} & 2005 & \cmark & \cmark & & & & & \cmark \\ 
GBVS & Harel et al.\cite{harel2006} & 2006 & & & \cmark & & & \cmark & \cmark \\
SDLF & Torralba et al. \cite{Torralba2006} & 2006 & &  & \cmark & & & \cmark & \cmark \\
SR \& PFT & Hou \& Zhang\cite{Hou2007} & 2007 &  & & & \cmark & & \cmark & \\
PQFT & Guo \& Zhang\cite{ChenleiGuo2008} & 2008 &  & & & \cmark & & \cmark & \\
ICL & Hou \& Zhang \cite{Hou2009} & 2008 & & \cmark & \cmark & & & \cmark & \cmark \\
SUN & Zhang et al. \cite{Zhang2008} & 2008 & &  & \cmark & & &  & \cmark \\
SDSR & Seo \& Milanfar \cite{Seo2009} & 2009 & \cmark & & \cmark & & & \cmark & \cmark \\
FT & Achanta et al.\cite{Achanta2009} & 2009 &  & & & \cmark & & \cmark & \\
DCTS/SIGS & Hou et al.\cite{XiaodiHou2012} & 2011 & & & & \cmark & & \cmark & \\
SIM & Murray et al.\cite{Murray2011} & 2011 & \cmark & & & & & \cmark & \cmark \\
WMAP & Lopez-Garcia et al.\cite{LopezGarcia2011} & 2011 & \cmark & & & \cmark & & \cmark & \cmark \\
AWS & Garcia-Diaz et al.\cite{GarciaDiaz2012} & 2012 & \cmark & & & & & \cmark & \cmark \\
CASD & Goferman et al.\cite{Goferman2012} & 2012 & \cmark & \cmark & \cmark & & \cmark & \cmark & \cmark \\
RARE & Riche et al.\cite{Riche2012}& 2012 & & \cmark &  & & & \cmark & \cmark \\ 
QDCT & Schauerte et al.\cite{Schauerte2012b} & 2012 &  & & & \cmark & & \cmark & \\ 
HFT & Li et al.\cite{Li2013} & 2013 &  & & & \cmark & & \cmark & \\ 
BMS & Zhang \& Sclaroff \cite{Zhang2013b} & 2013 & & & \cmark & & & \cmark &  \\
SALICON & Jiang et al.\cite{Jiang2015,christopherleethomas2016} & 2015 &  & & & & \cmark &  & \cmark \\
ML-Net & Cornia et al.\cite{mlnet2016} & 2016 &  & & & & \cmark &  & \cmark \\
DeepGazeII & K\"ummerer et al.\cite{Kummerer2017} & 2016 &  & & & & \cmark &  & \cmark \\
SalGAN & Pan et al.\cite{Pan_2017_SalGAN} & 2017 & & & & & \cmark &  & \cmark \\
ICF & K\"ummerer et al.\cite{Kummerer2017} & 2017 & & & \cmark & & & \cmark & \cmark \\
SAM & Cornia et al.\cite{cornia2018predicting} & 2018 &  & & & & \cmark &  & \cmark \\ 
NSWAM & Berga \& Otazu \cite{Berga2018b} & 2018 & \cmark & & & & & \cmark & \cmark \\
Sal-DCNN & Jiang et al. \cite{jiang2019saldcnn} & 2019 & & & & \cmark & \cmark & \cmark & \cmark \\
\hline
\end{tabular}
\\
Inspiration: \{\colorbox{red!20}{ C }: Cognitive/Biological, \colorbox{green!20}{ I }: Information-Theoretic, \colorbox{blue!20}{ P }: Probabilistic, \colorbox{orange!20}{ F }: Fourier/Spectral, \colorbox{cyan!20}{ D }: Machine/Deep Learning\}
Type: \{G: Global, L: Local\}
\label{table:models}
\end{table}

\subsection{Problem formulation}

Visual saliency is a term coined on a perceptual basis. According to this principle, a correct modelization of saliency should consider specific experimental conditions upon a visual attention task. The output of such a model can vary for stimulus or task, but must arise as a common behavioral phenomena in order to validate the general hypothesis definition from Treisman, Wolfe, Itti and colleagues \cite{Treisman1980,Wolfe2010a,Itti2000}. Eye movements have been considered the main behavioral markers of visual attention. But understanding saliency means not only to prove how visual fixations can be predicted, but to simulate which patterns of eye movements are gathered from vision and its sensory signals (here avoiding any top-down influences). This challenge offers eye tracking researchers to consider several experimental issues (with respect contextual, contrast, temporal, oculomotor and task-related biases) when capturing bottom-up attention, largely explained by Borji et al. \cite{Borji2013c}, Bruce et al. \cite{Bruce2015} and lately by Berga et al. \cite{Berga2018a}. Computational models advance several ways to predict, to some extent, human visual fixations. However, the limits of the prediction capability of these saliency models arise as a consequence of the validity of the evaluation from eye tracking experimentation. We aim to to provide a new dataset with uniquely synthetic images and a benchmark, studying for each saliency model:
\begin{enumerate}
    \item How model inspiration and feature processing influences model predictions?
    \item How does temporality of fixations affect model predictions?
    \item How low-level feature type and contrast influences model's psychophysical measurements?
\end{enumerate}


\section{SID4VAM: Synthetic Image Dataset for Visual Attention Modeling} \label{sec:dataset}

\begin{figure}[!] 
\centering 
\makebox[1em]{1)}
\fbox{\includegraphics[width=.12\linewidth]{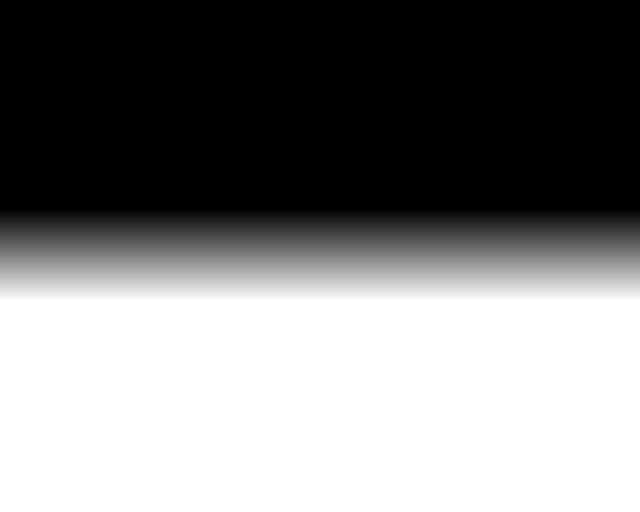}}
\fbox{\includegraphics[width=.12\linewidth]{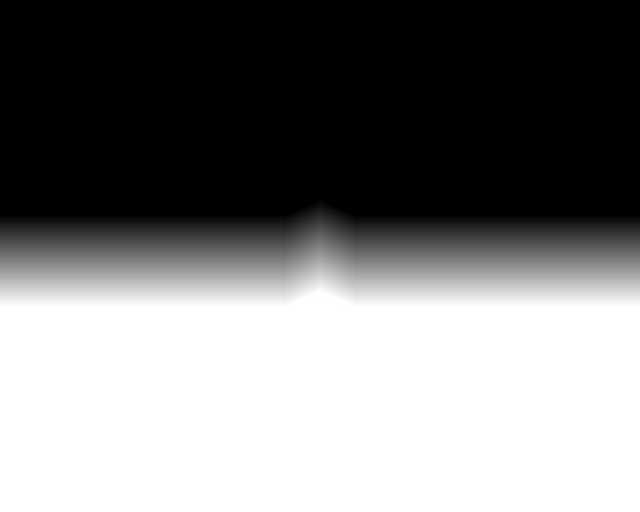}}
\fbox{\includegraphics[width=.12\linewidth]{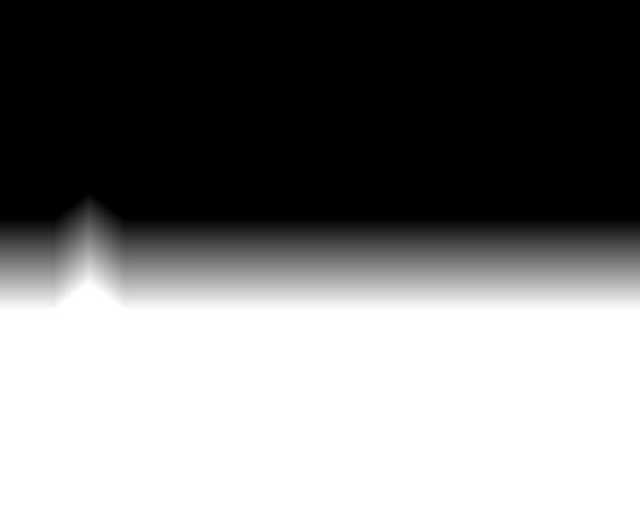}}
\fbox{\includegraphics[width=.12\linewidth]{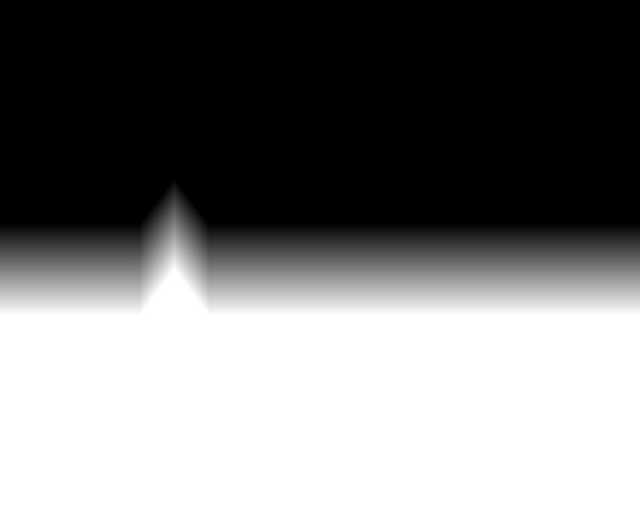}}
\fbox{\includegraphics[width=.12\linewidth]{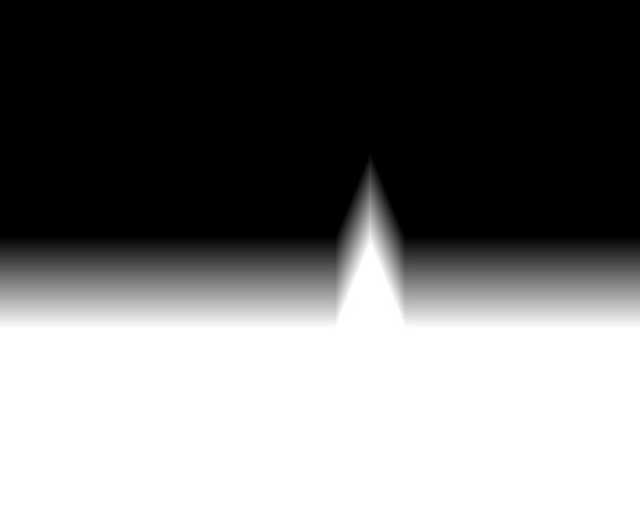}}
\fbox{\includegraphics[width=.12\linewidth]{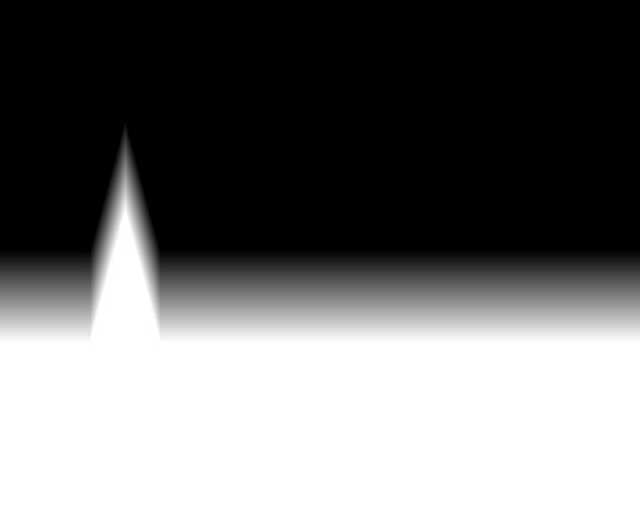}}
\fbox{\includegraphics[width=.12\linewidth]{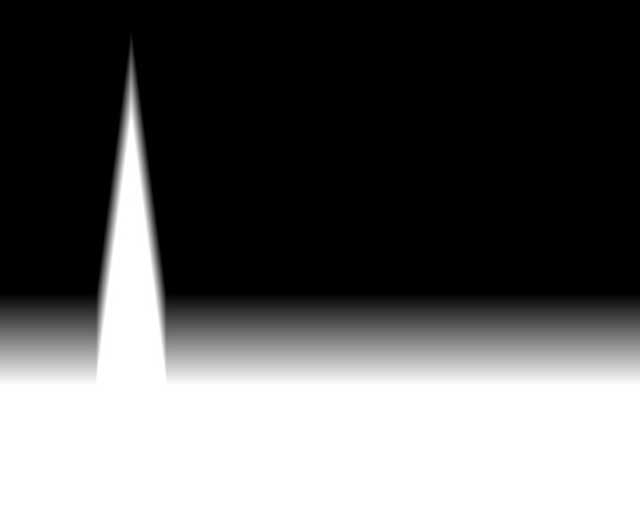}}
\\\makebox[1em]{2)}
\fbox{\includegraphics[width=.12\linewidth]{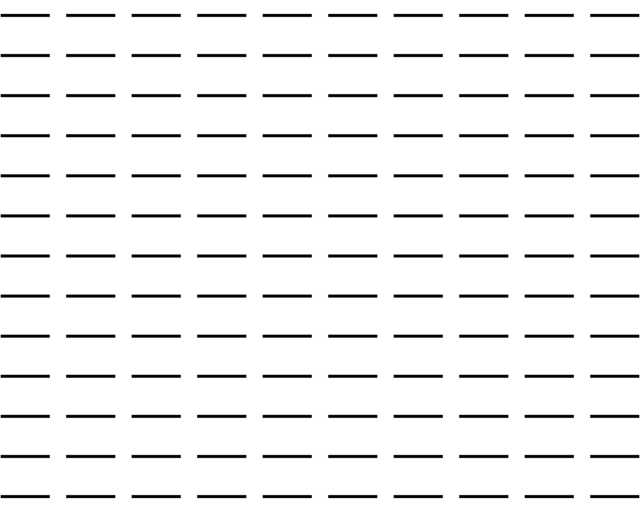}}
\fbox{\includegraphics[width=.12\linewidth]{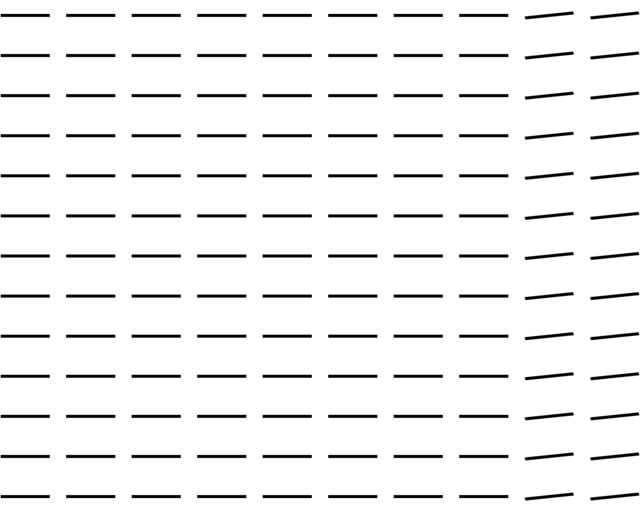}}
\fbox{\includegraphics[width=.12\linewidth]{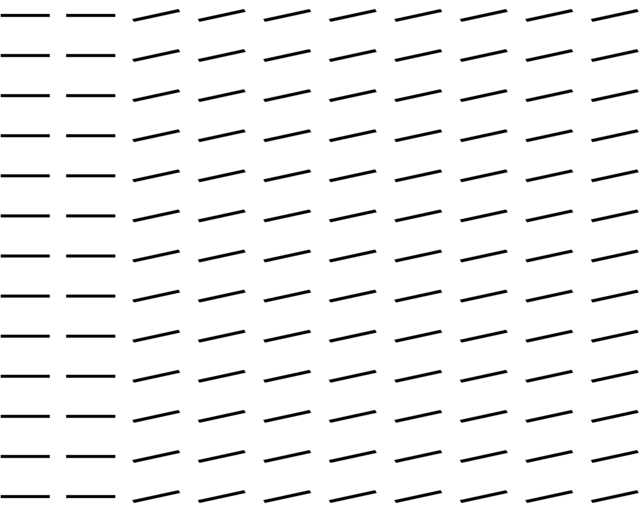}}
\fbox{\includegraphics[width=.12\linewidth]{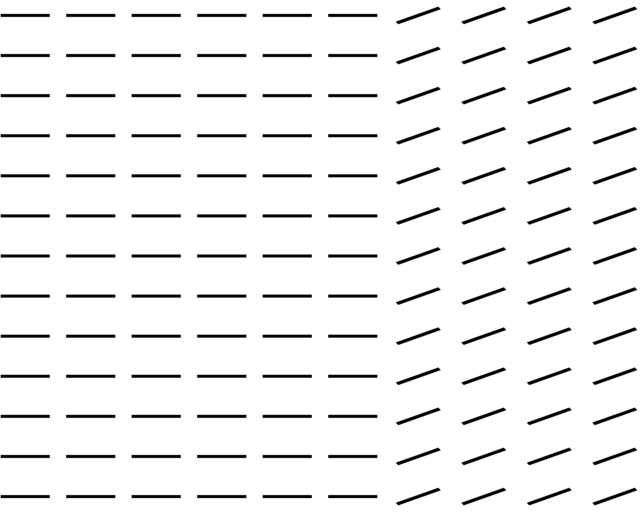}}
\fbox{\includegraphics[width=.12\linewidth]{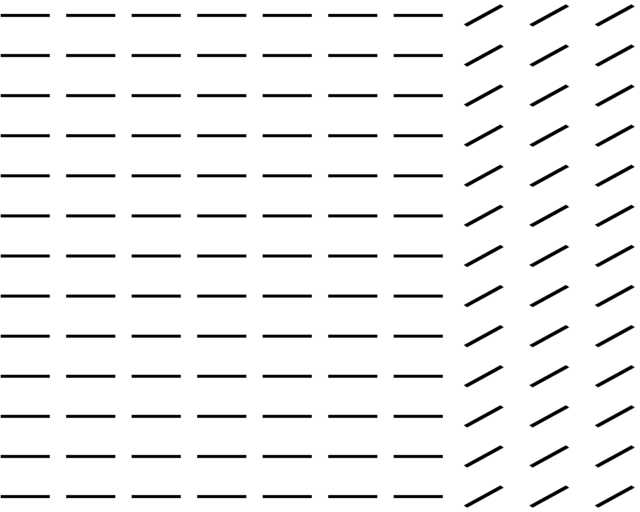}}
\fbox{\includegraphics[width=.12\linewidth]{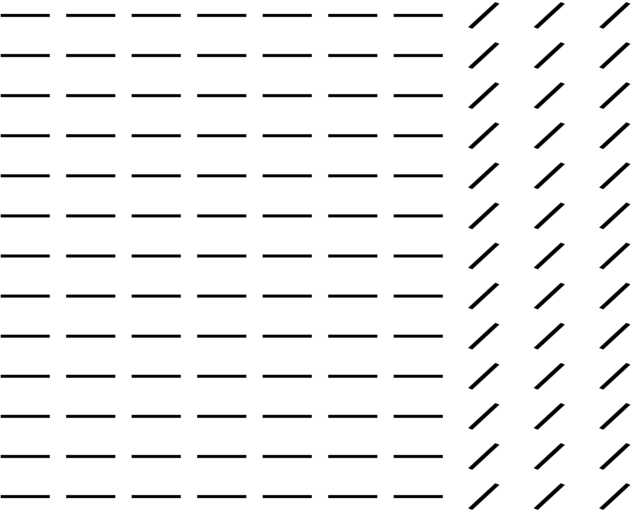}}
\fbox{\includegraphics[width=.12\linewidth]{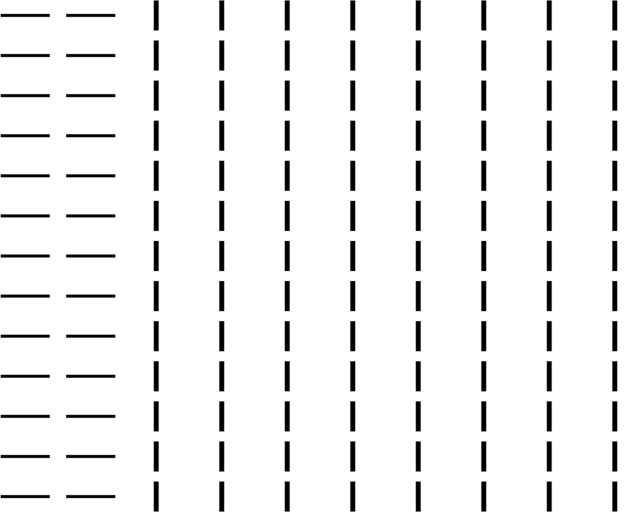}}
\\\hspace{1em}
\fbox{\includegraphics[width=.12\linewidth]{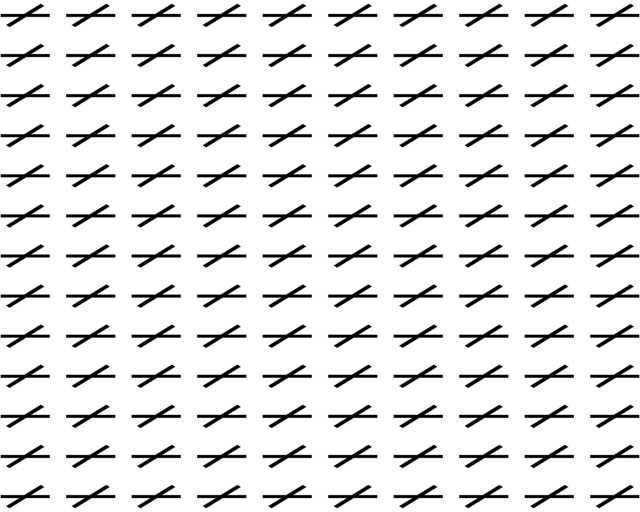}}
\fbox{\includegraphics[width=.12\linewidth]{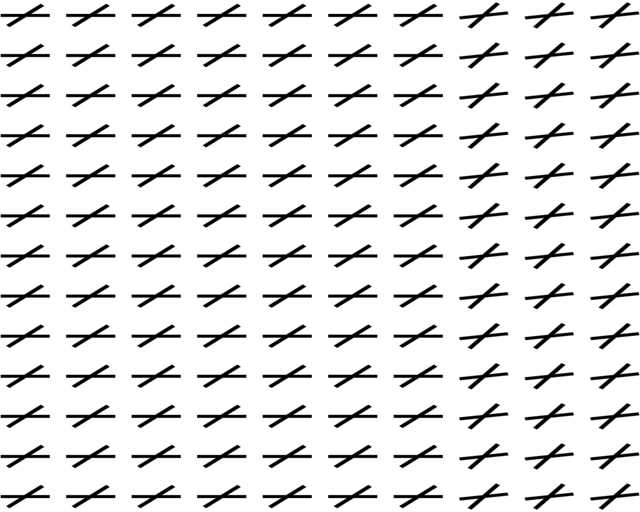}}
\fbox{\includegraphics[width=.12\linewidth]{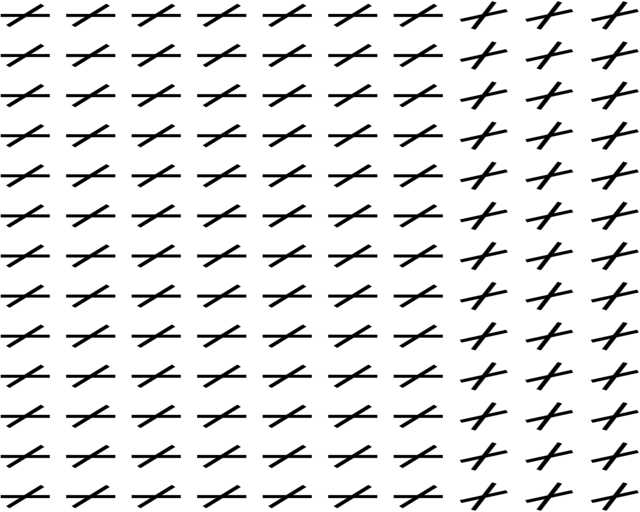}}
\fbox{\includegraphics[width=.12\linewidth]{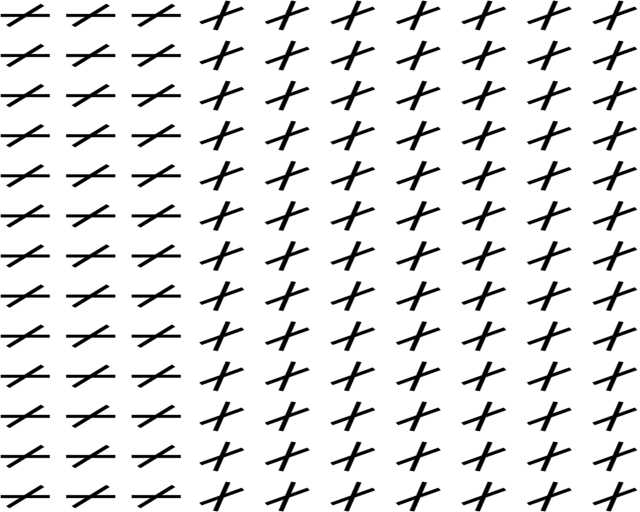}}
\fbox{\includegraphics[width=.12\linewidth]{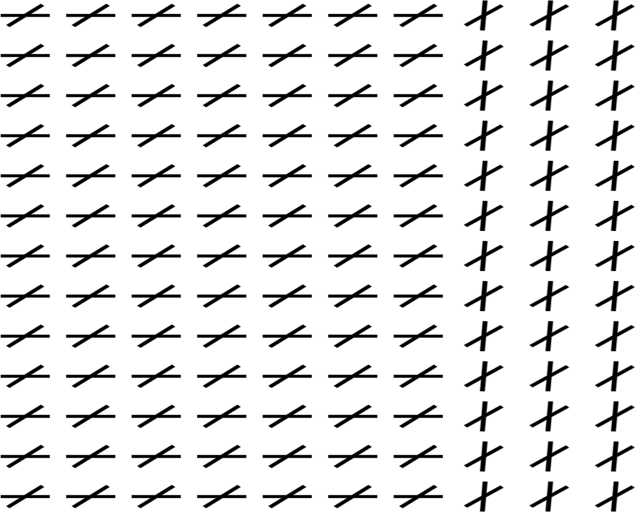}}
\fbox{\includegraphics[width=.12\linewidth]{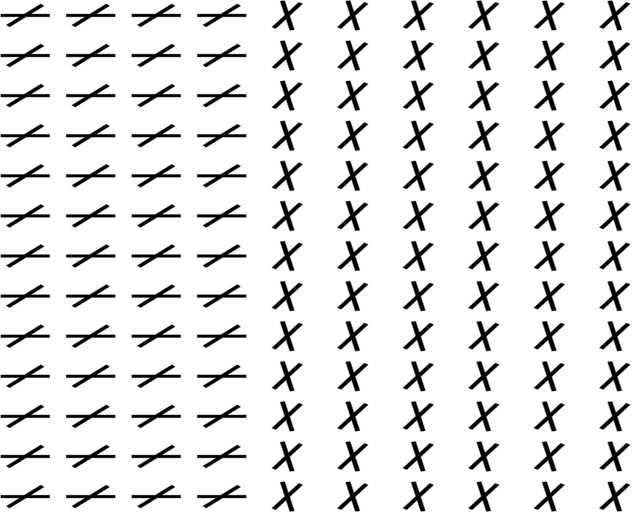}}
\fbox{\includegraphics[width=.12\linewidth]{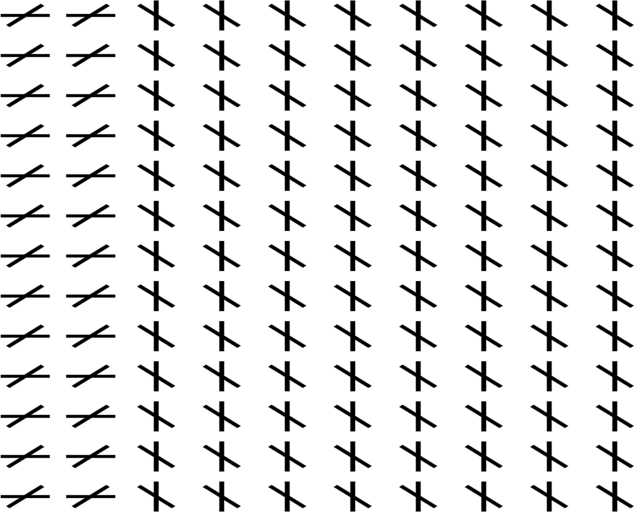}}
\\\makebox[1em]{3)}
\fbox{\includegraphics[width=.12\linewidth]{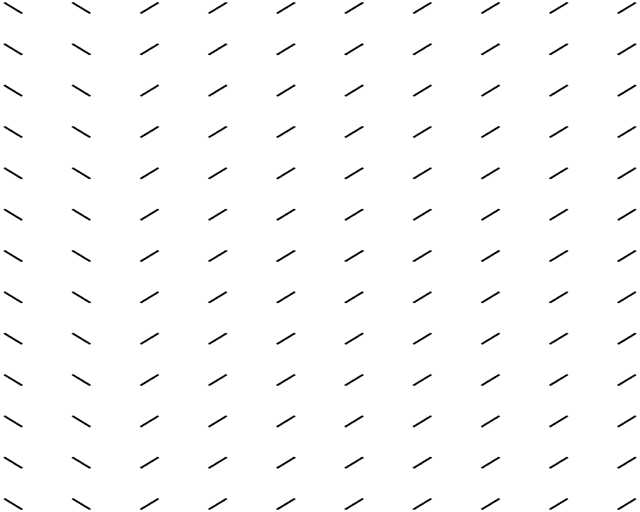}}
\fbox{\includegraphics[width=.12\linewidth]{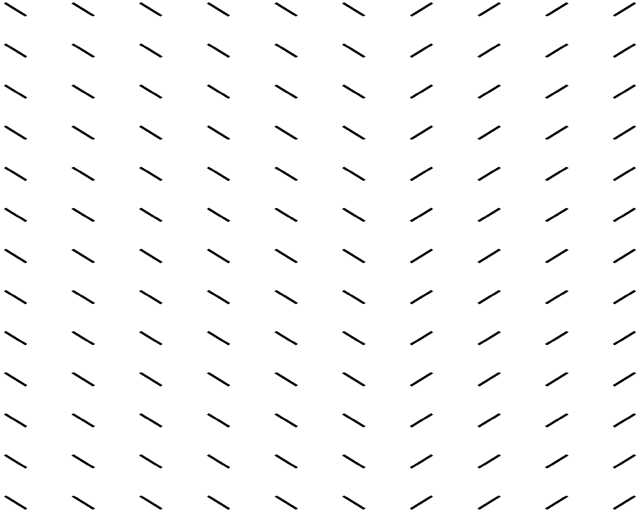}}
\fbox{\includegraphics[width=.12\linewidth]{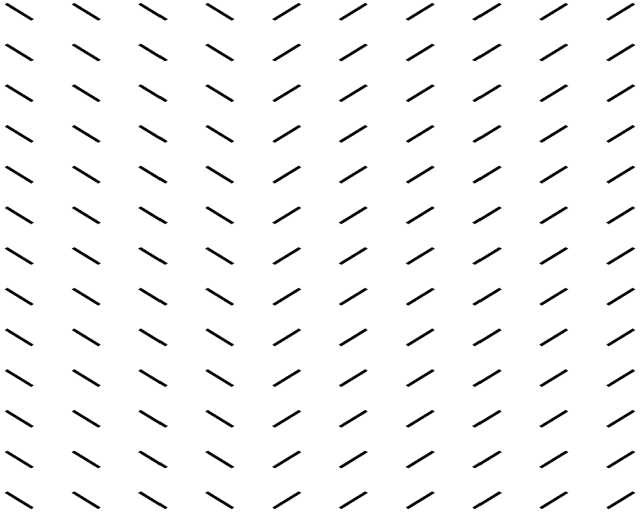}}
\fbox{\includegraphics[width=.12\linewidth]{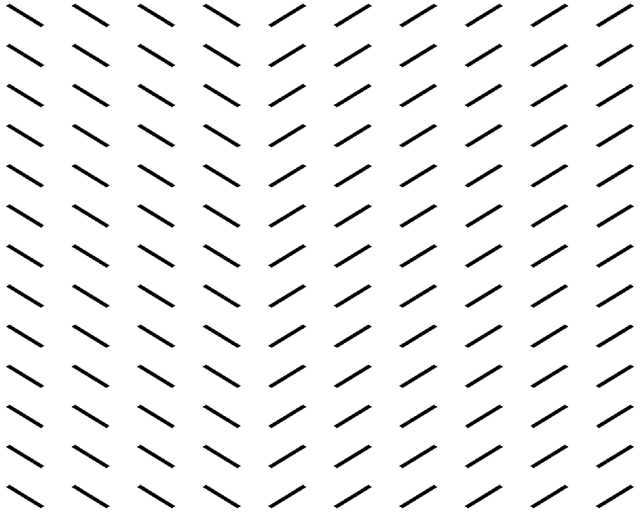}}
\fbox{\includegraphics[width=.12\linewidth]{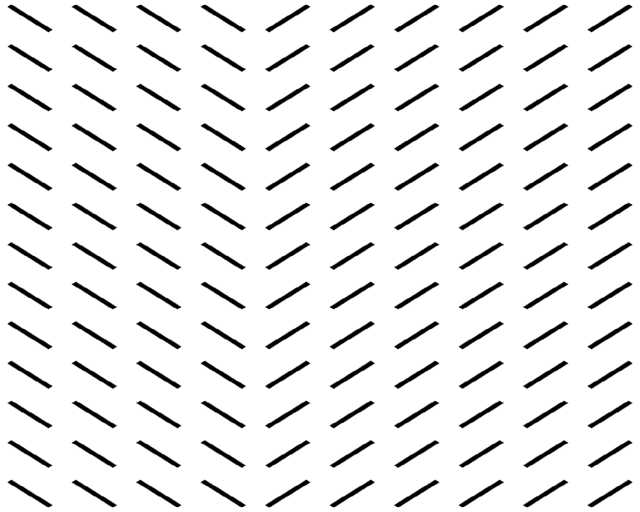}}
\fbox{\includegraphics[width=.12\linewidth]{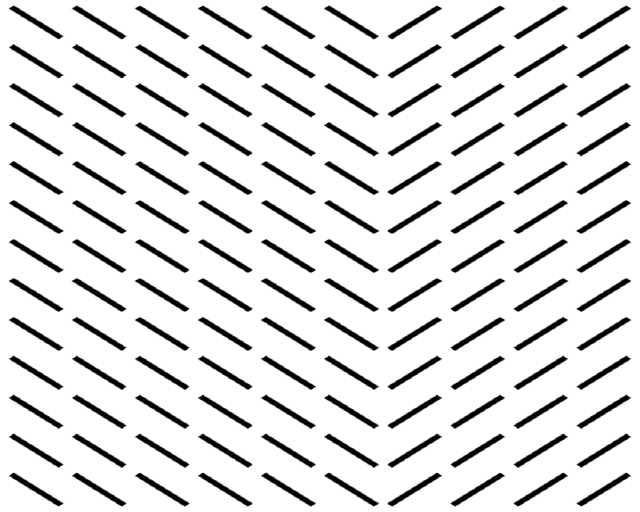}}
\fbox{\includegraphics[width=.12\linewidth]{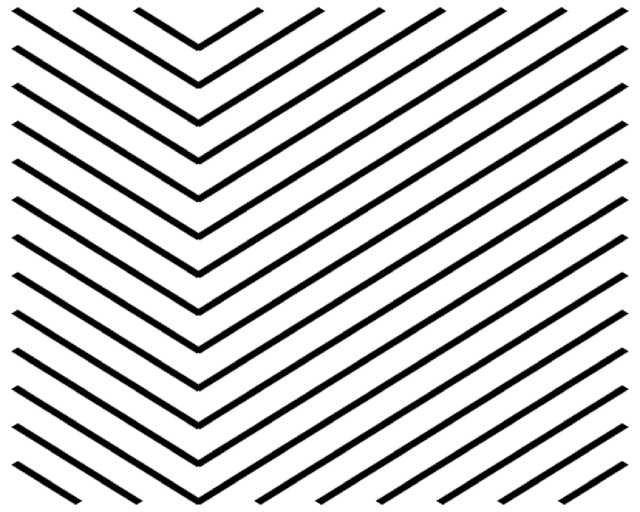}}
\\\makebox[1em]{4)}
\fbox{\includegraphics[width=.12\linewidth]{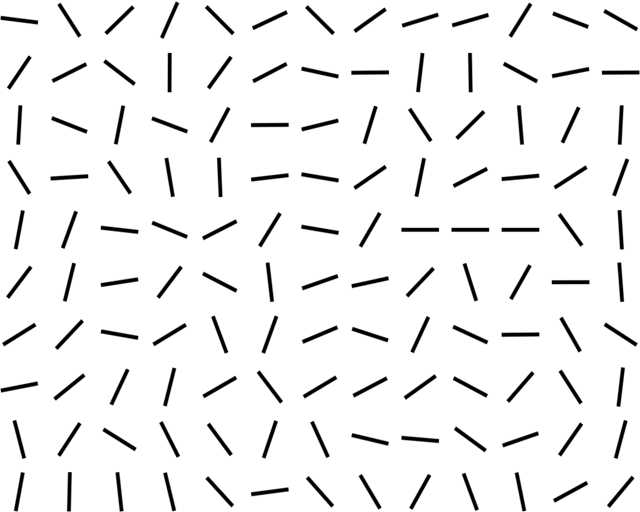}}
\fbox{\includegraphics[width=.12\linewidth]{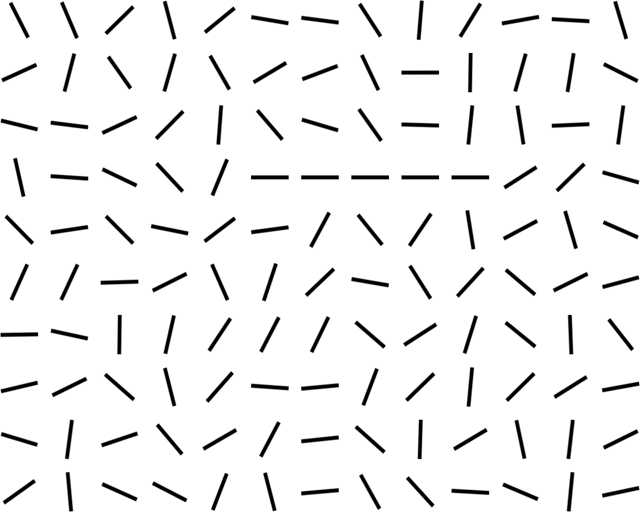}}
\fbox{\includegraphics[width=.12\linewidth]{images/d2Bfv4BdefaultB5.jpg}}
\fbox{\includegraphics[width=.12\linewidth]{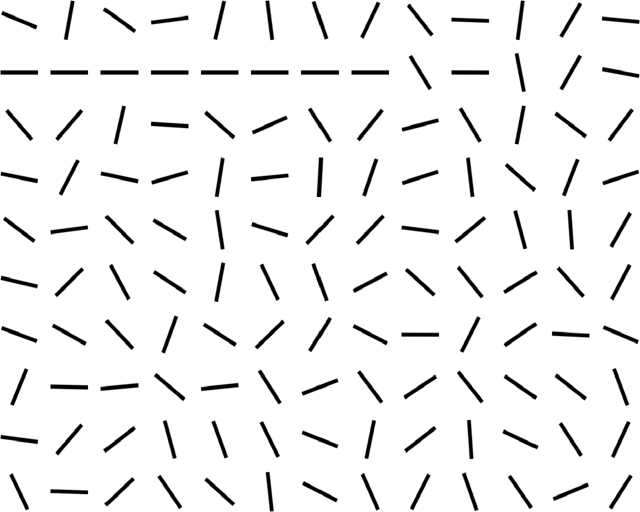}}
\fbox{\includegraphics[width=.12\linewidth]{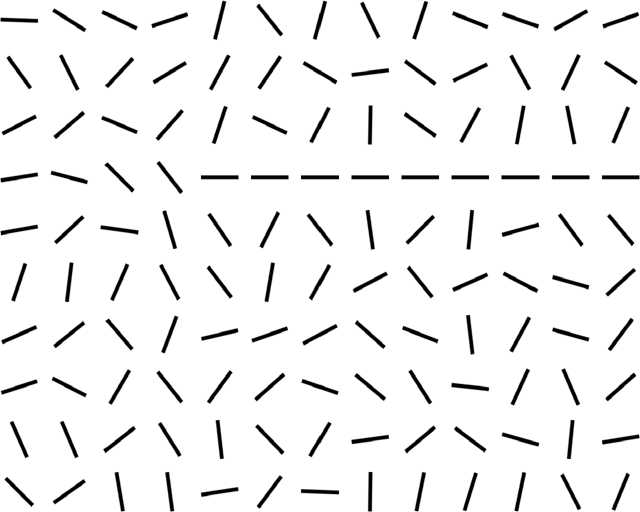}}
\fbox{\includegraphics[width=.12\linewidth]{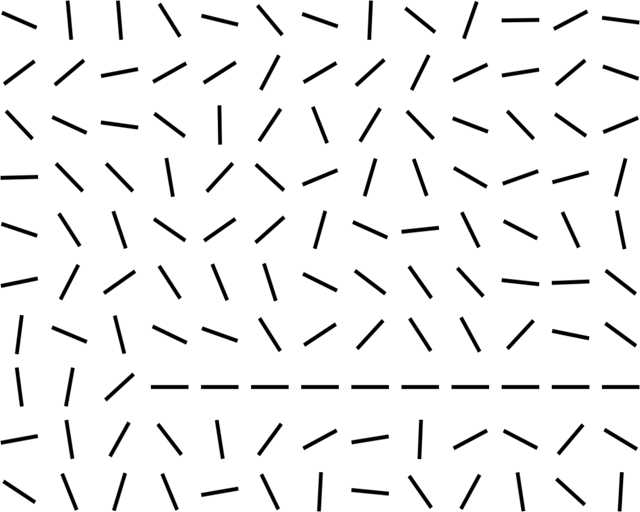}}
\\\makebox[1em]{5)}
\fbox{\includegraphics[width=.12\linewidth]{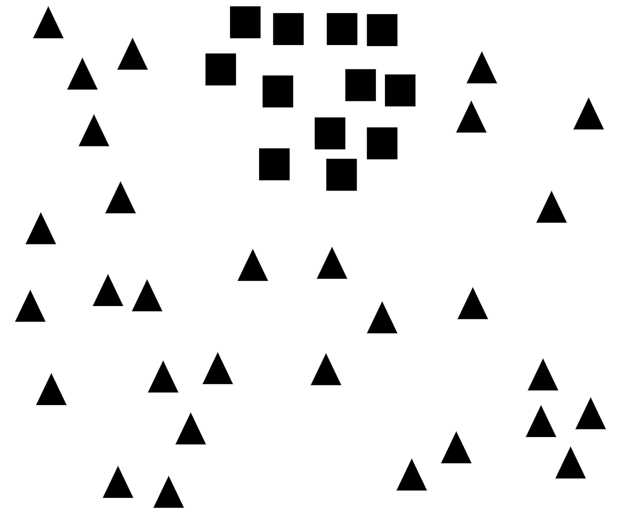}}
\fbox{\includegraphics[width=.12\linewidth]{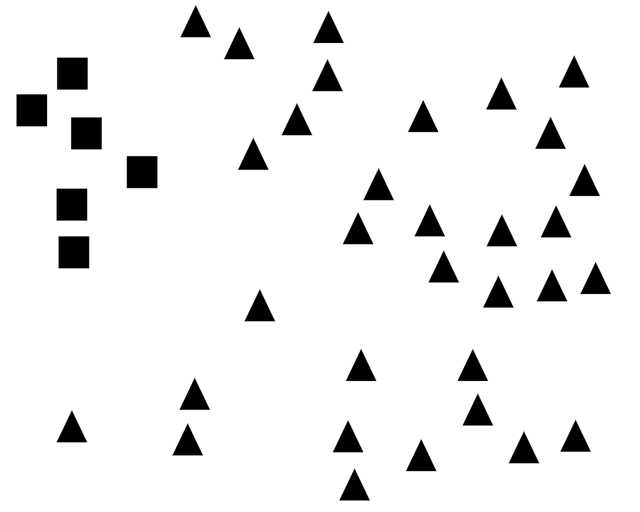}}
\fbox{\includegraphics[width=.12\linewidth]{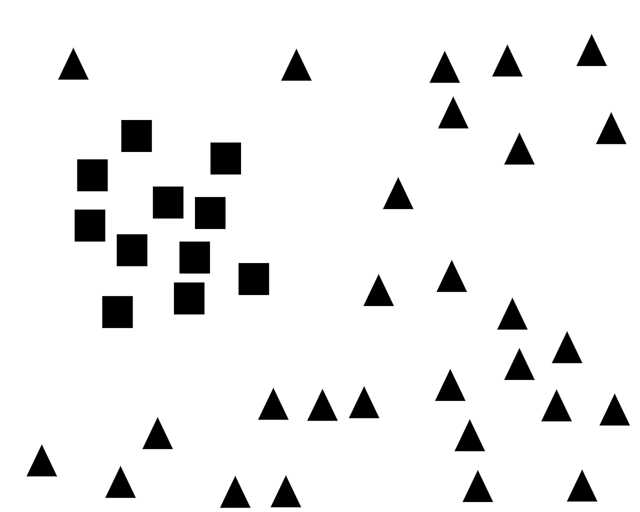}}
\fbox{\includegraphics[width=.12\linewidth]{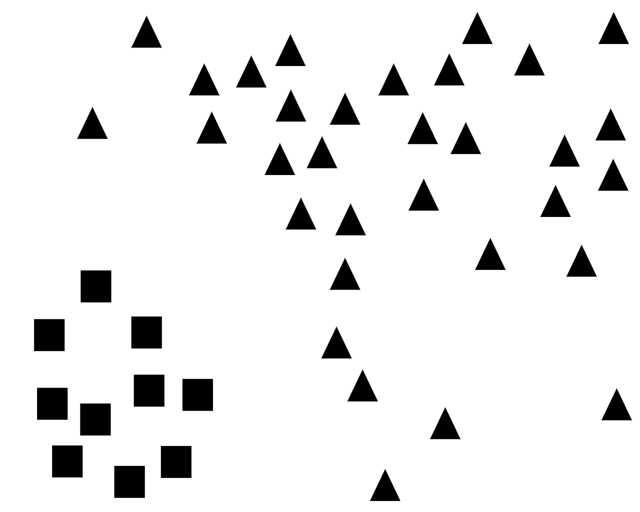}}
\fbox{\includegraphics[width=.12\linewidth]{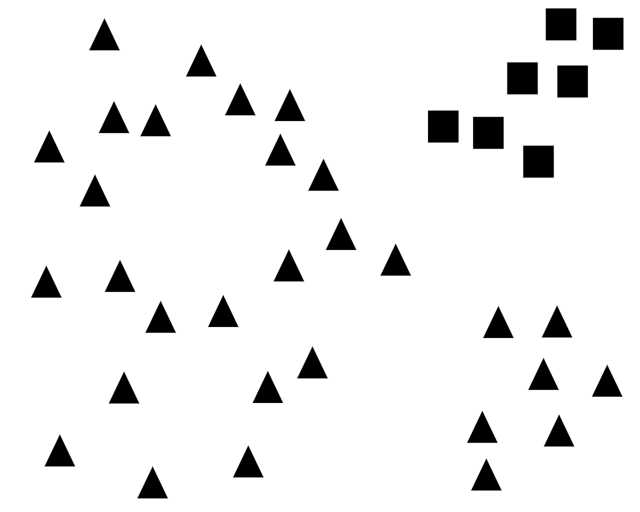}}
\fbox{\includegraphics[width=.12\linewidth]{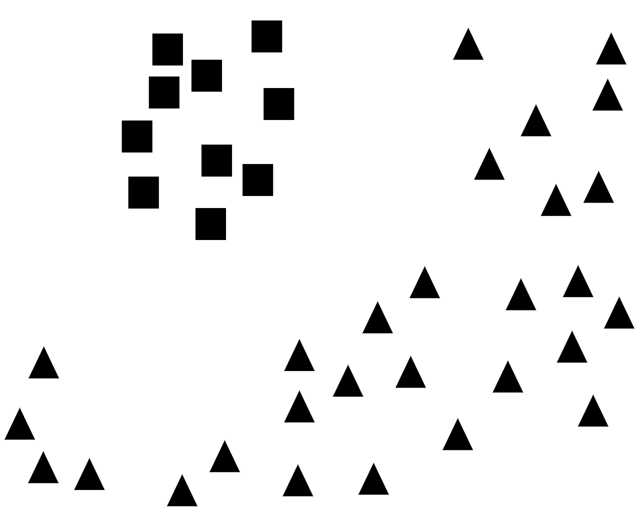}}
\fbox{\includegraphics[width=.12\linewidth]{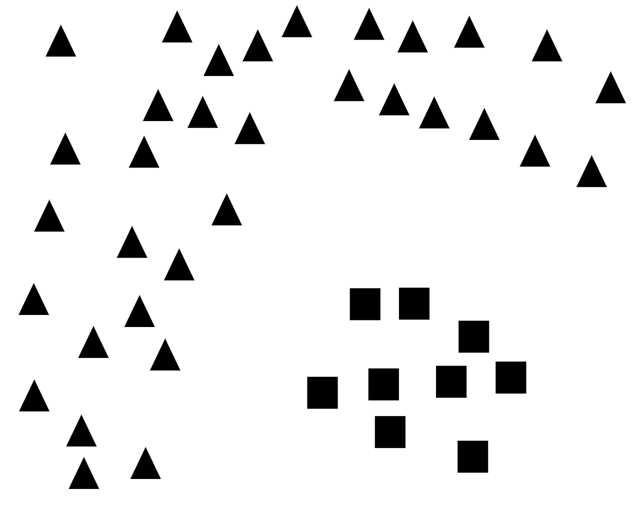}}
\\\hspace{1em}
\fbox{\includegraphics[width=.12\linewidth]{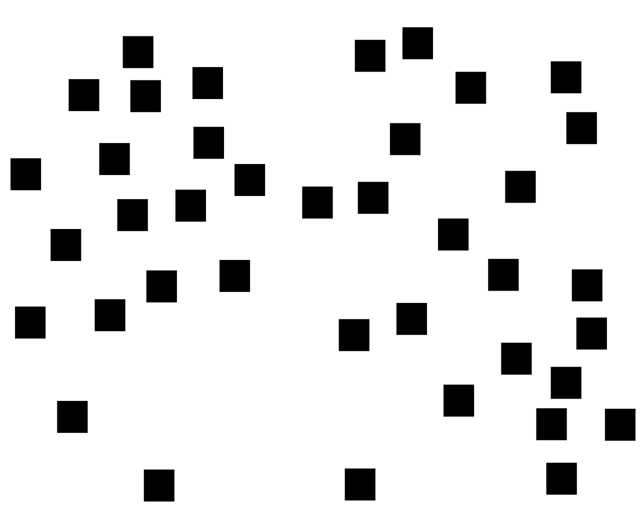}}
\fbox{\includegraphics[width=.12\linewidth]{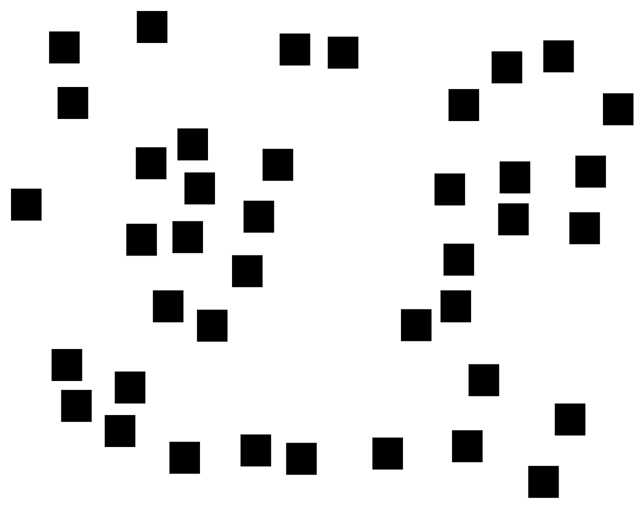}}
\fbox{\includegraphics[width=.12\linewidth]{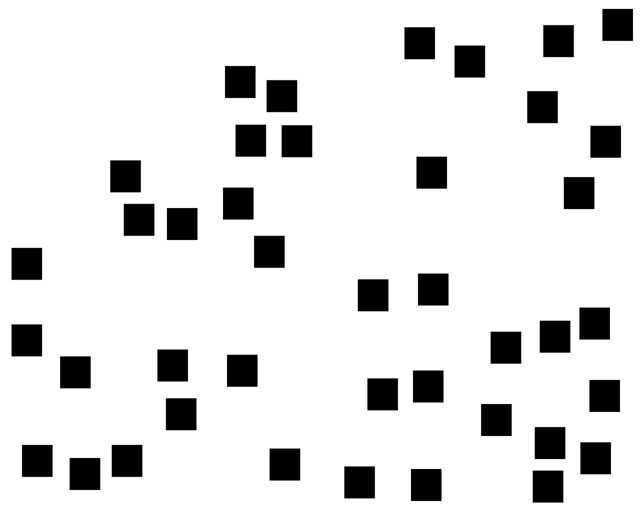}}
\fbox{\includegraphics[width=.12\linewidth]{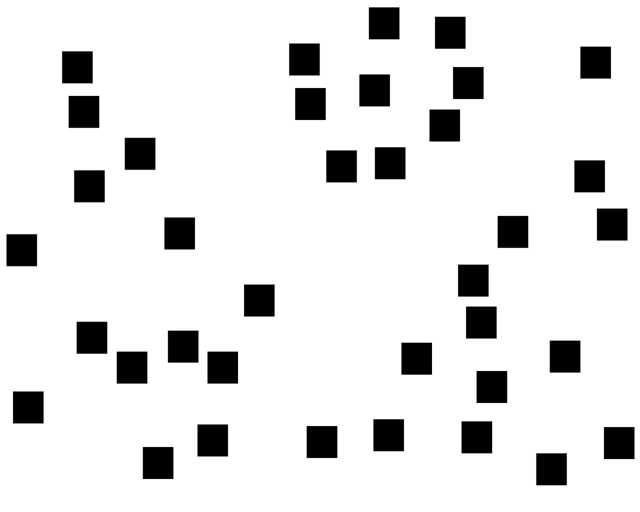}}
\fbox{\includegraphics[width=.12\linewidth]{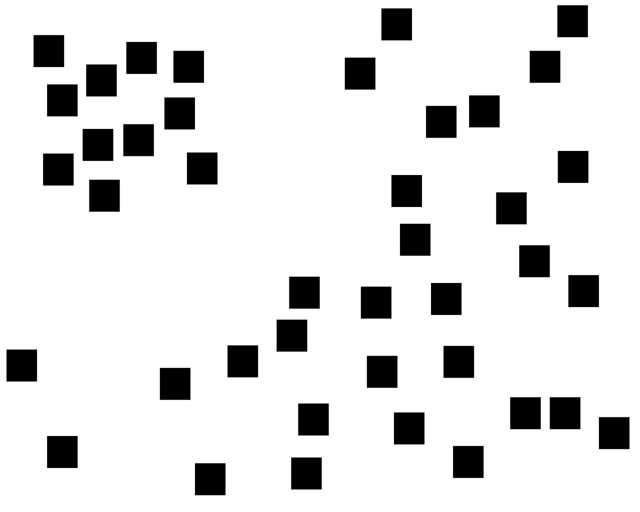}}
\fbox{\includegraphics[width=.12\linewidth]{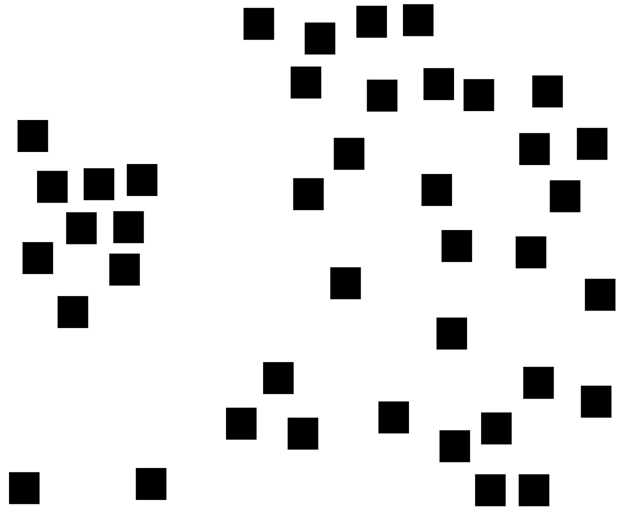}}
\fbox{\includegraphics[width=.12\linewidth]{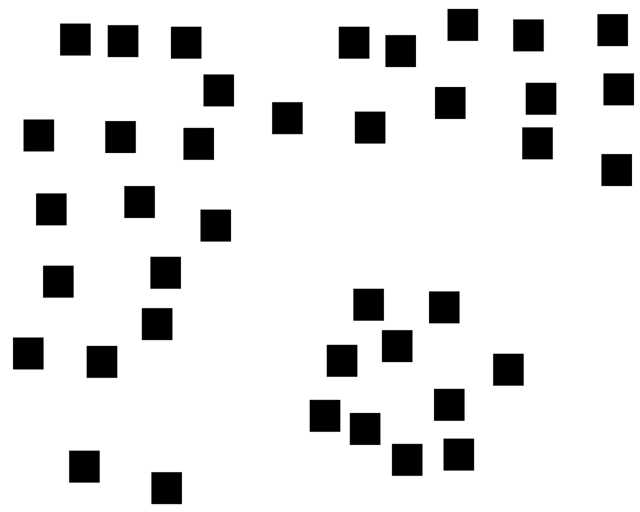}}
\\
\makebox[0.05\linewidth]{ }\makebox[0.13\linewidth]{1}\makebox[0.13\linewidth]{2}\makebox[0.13\linewidth]{3}\makebox[0.13\linewidth]{4}\makebox[0.13\linewidth]{5}\makebox[0.13\linewidth]{6}\makebox[0.13\linewidth]{7}
hard  $\longleftarrow$ $\Psi$ $\longrightarrow$ easy
\caption{Free-Viewing stimuli}
\label{fig:fv}
\end{figure}

\begin{figure}[!]
\centering
\makebox[1em]{6)}
\fbox{\includegraphics[width=.12\linewidth]{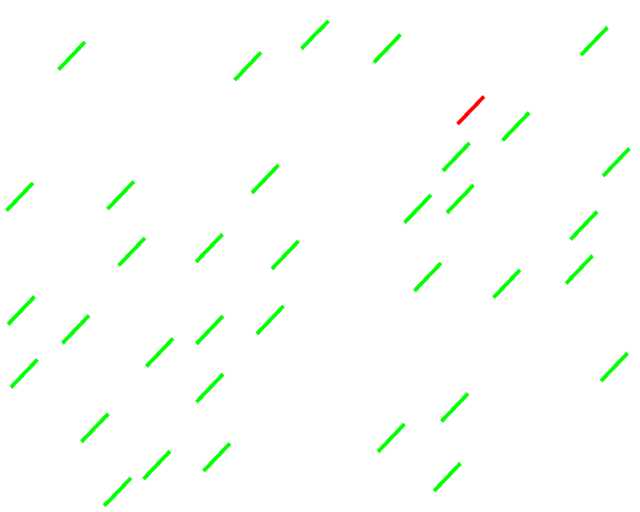}}
\fbox{\includegraphics[width=.12\linewidth]{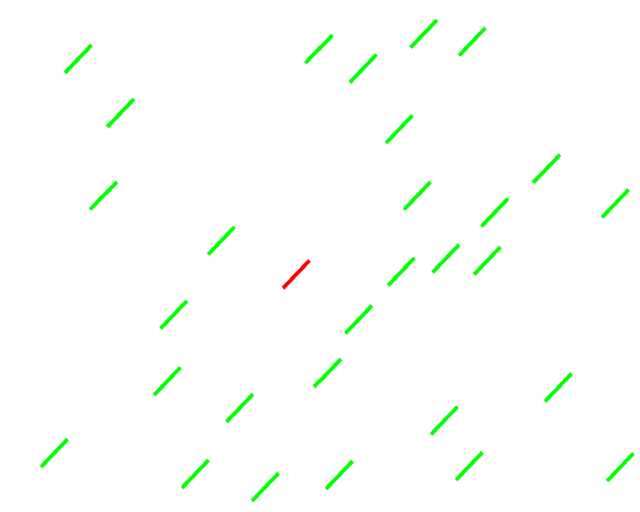}}
\fbox{\includegraphics[width=.12\linewidth]{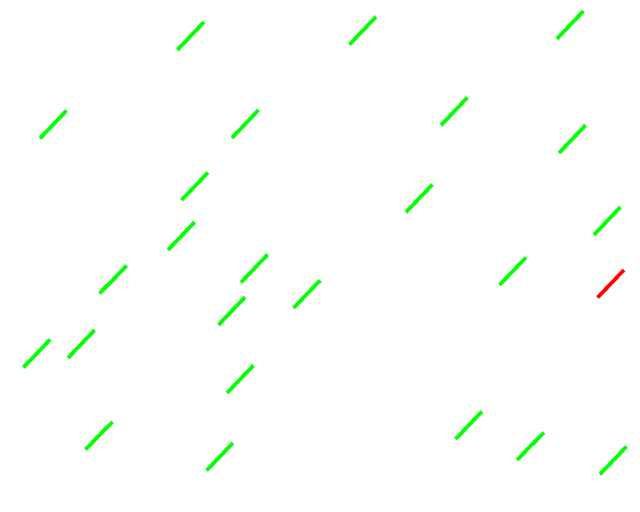}}
\fbox{\includegraphics[width=.12\linewidth]{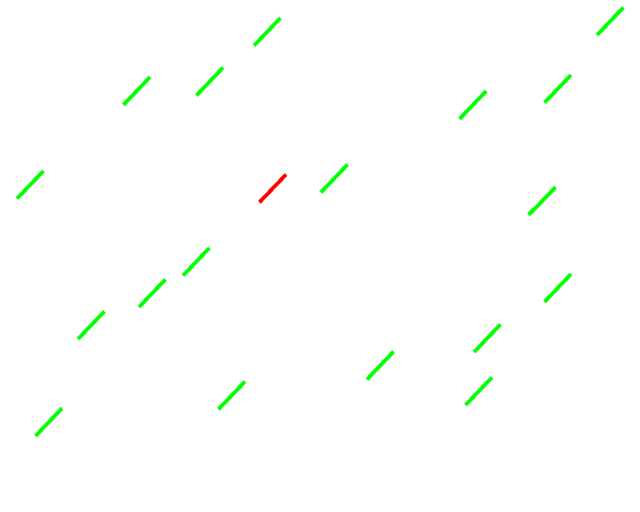}}
\fbox{\includegraphics[width=.12\linewidth]{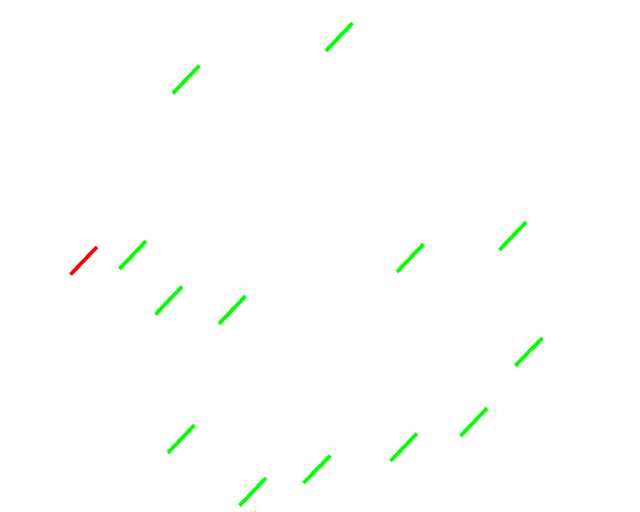}}
\fbox{\includegraphics[width=.12\linewidth]{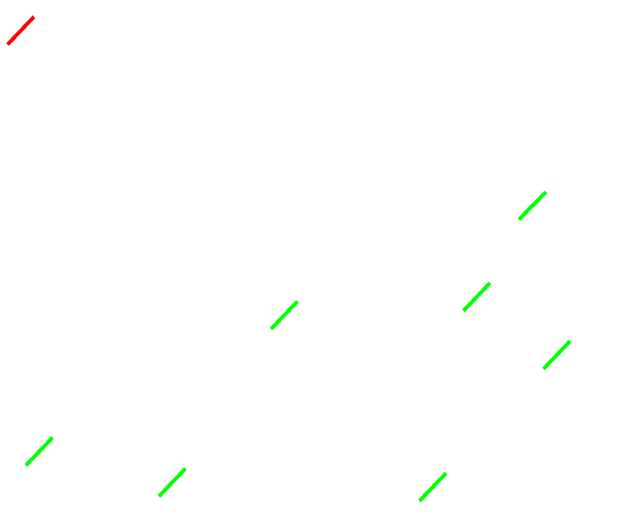}}
\fbox{\includegraphics[width=.12\linewidth]{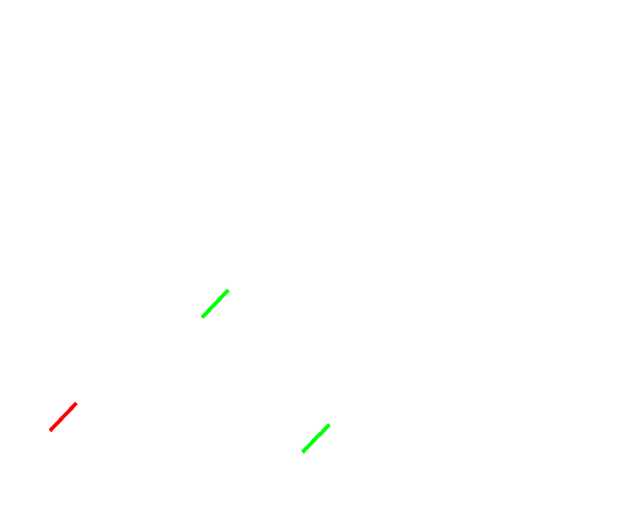}}
\\\hspace{1em}
\fbox{\includegraphics[width=.12\linewidth]{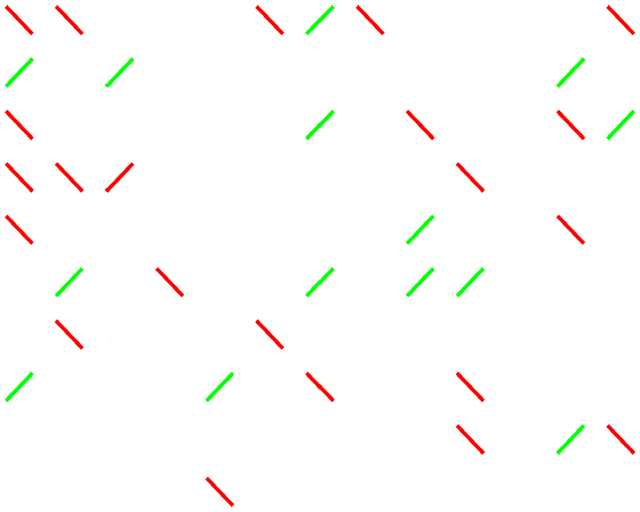}}
\fbox{\includegraphics[width=.12\linewidth]{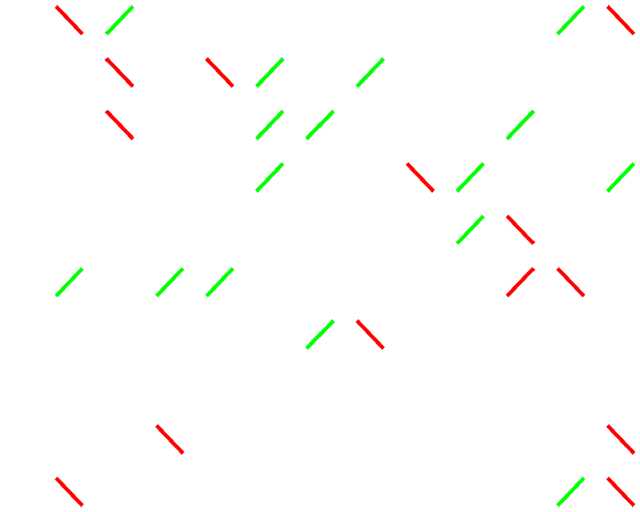}}
\fbox{\includegraphics[width=.12\linewidth]{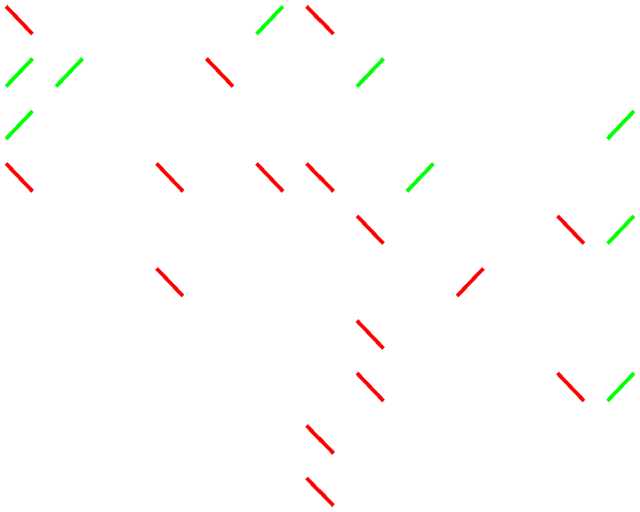}}
\fbox{\includegraphics[width=.12\linewidth]{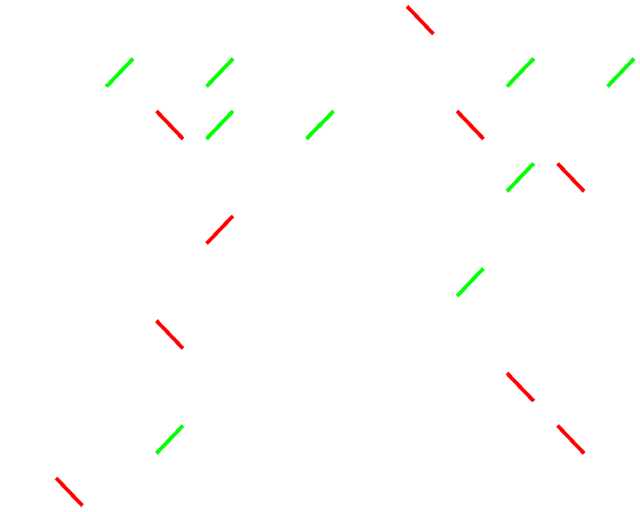}}
\fbox{\includegraphics[width=.12\linewidth]{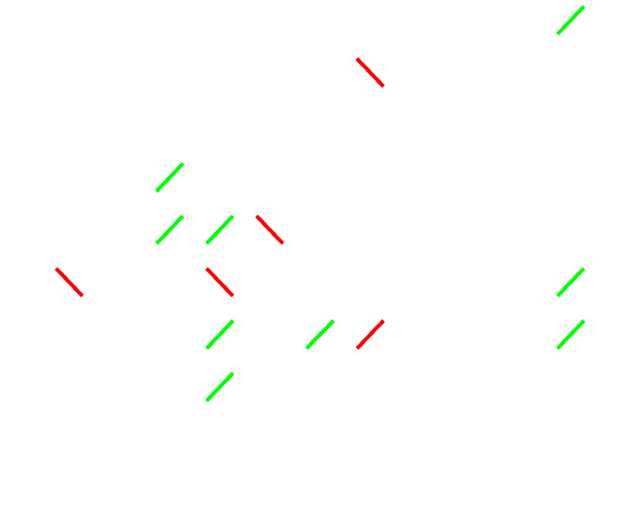}}
\fbox{\includegraphics[width=.12\linewidth]{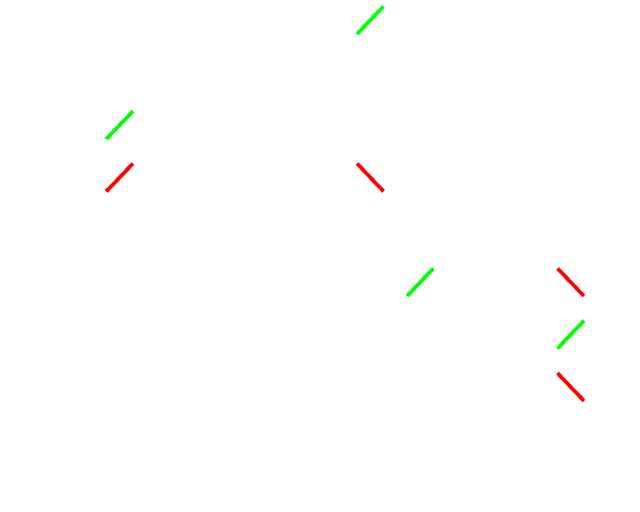}}
\fbox{\includegraphics[width=.12\linewidth]{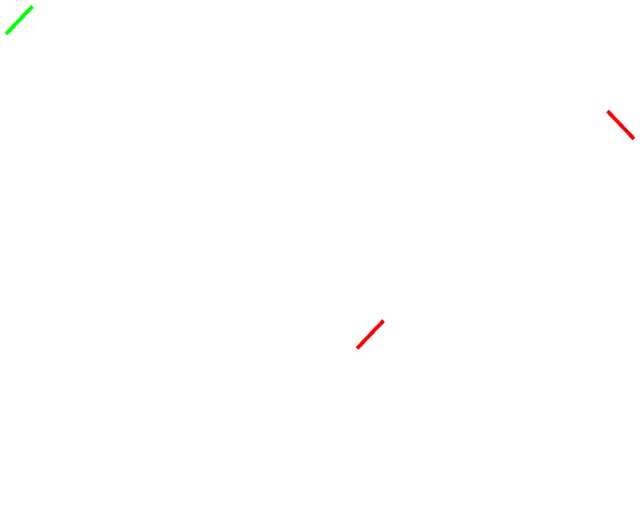}}
\\\hspace{1em}
\fbox{\includegraphics[width=.12\linewidth]{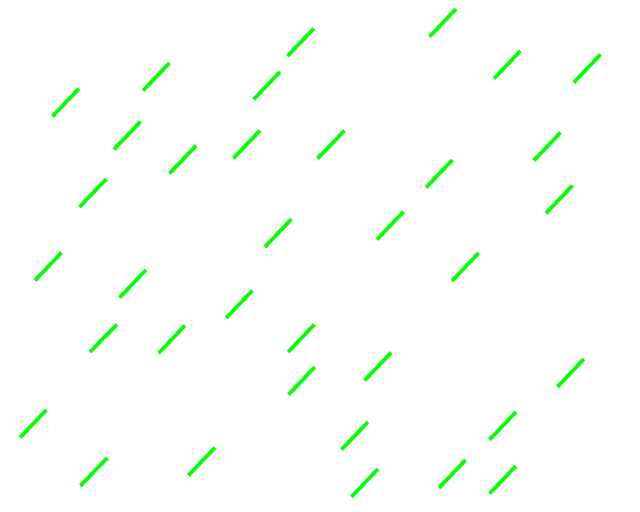}}
\fbox{\includegraphics[width=.12\linewidth]{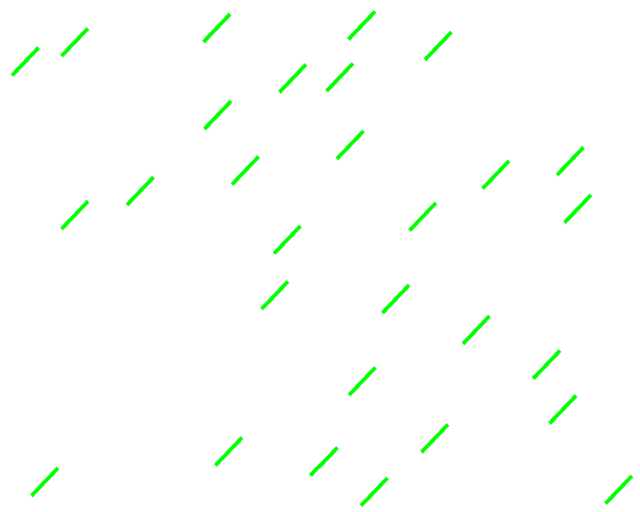}}
\fbox{\includegraphics[width=.12\linewidth]{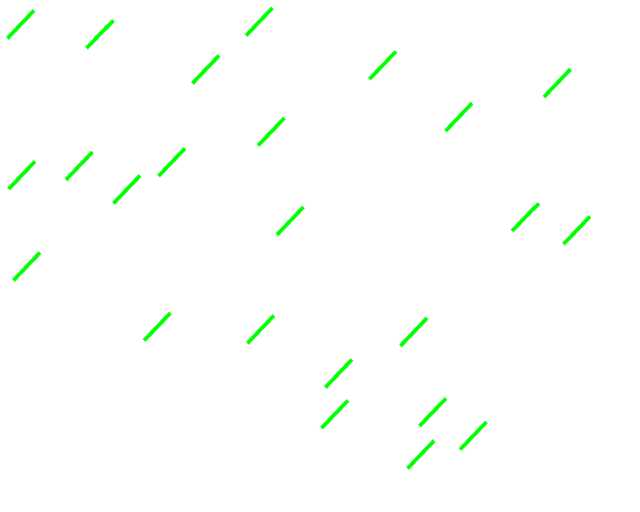}}
\fbox{\includegraphics[width=.12\linewidth]{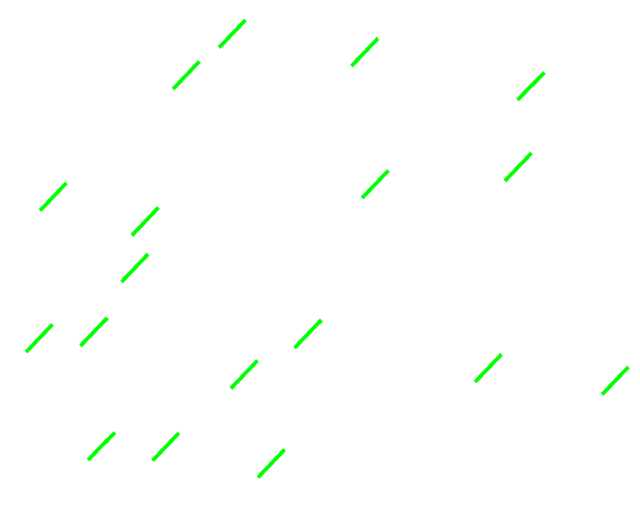}}
\fbox{\includegraphics[width=.12\linewidth]{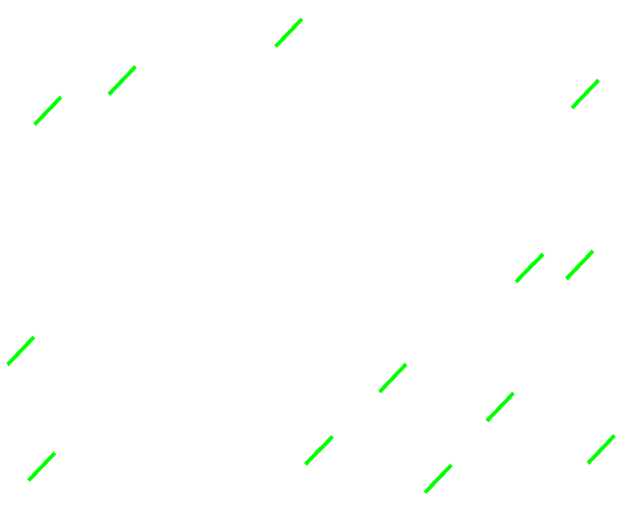}}
\fbox{\includegraphics[width=.12\linewidth]{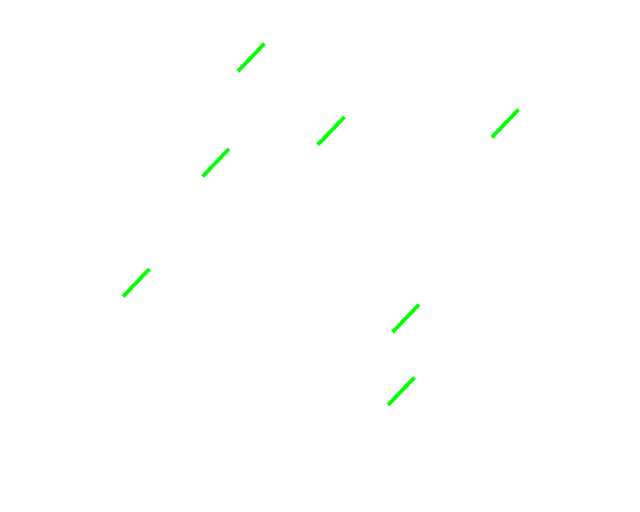}}
\fbox{\includegraphics[width=.12\linewidth]{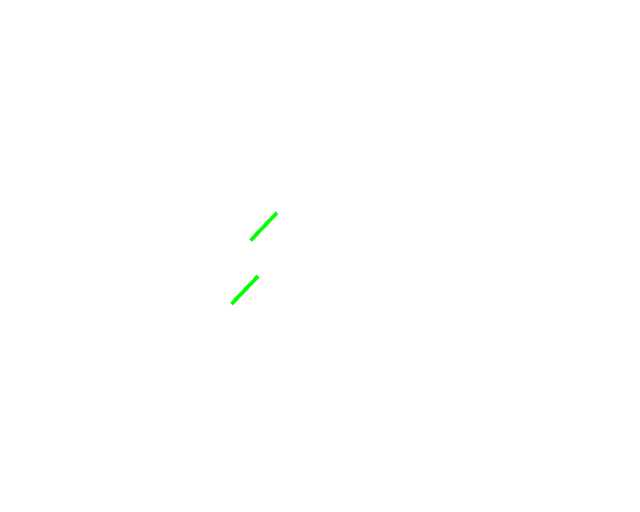}}
\\\hspace{1em}
\fbox{\includegraphics[width=.12\linewidth]{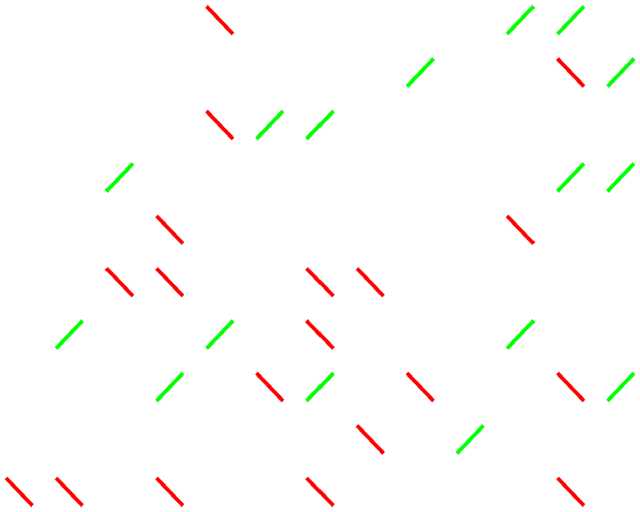}}
\fbox{\includegraphics[width=.12\linewidth]{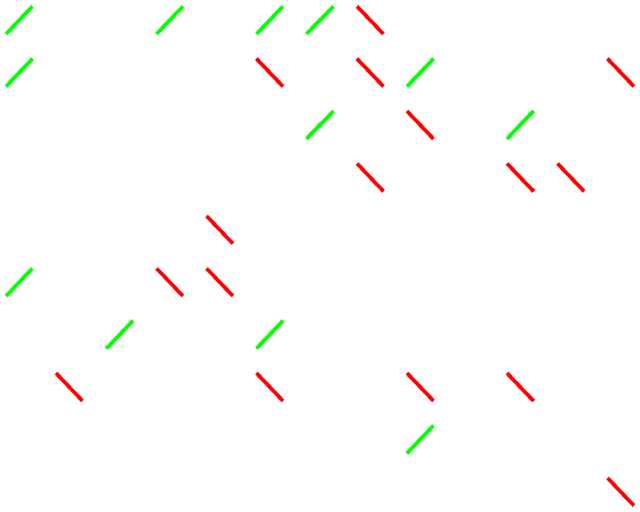}}
\fbox{\includegraphics[width=.12\linewidth]{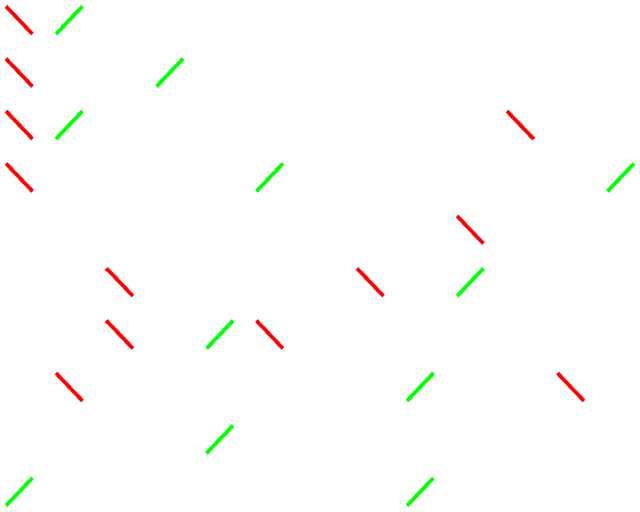}}
\fbox{\includegraphics[width=.12\linewidth]{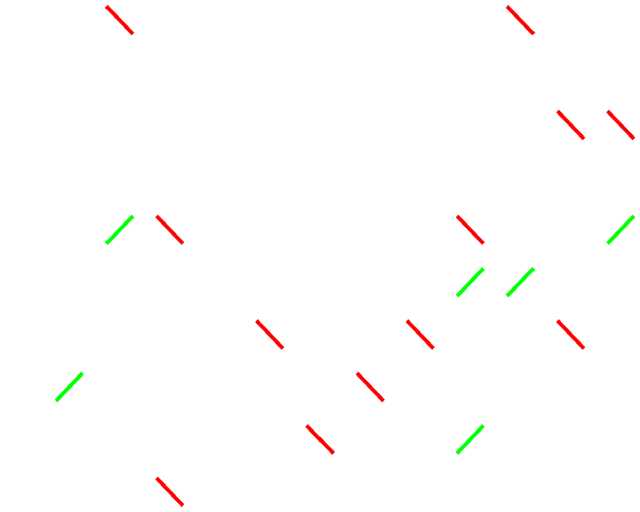}}
\fbox{\includegraphics[width=.12\linewidth]{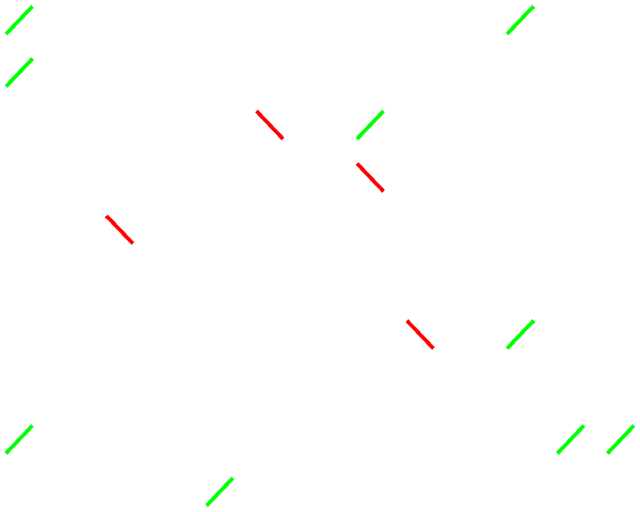}}
\fbox{\includegraphics[width=.12\linewidth]{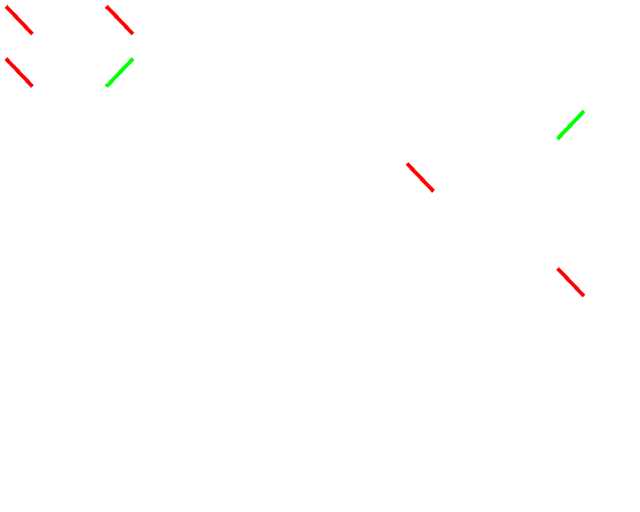}}
\fbox{\includegraphics[width=.12\linewidth]{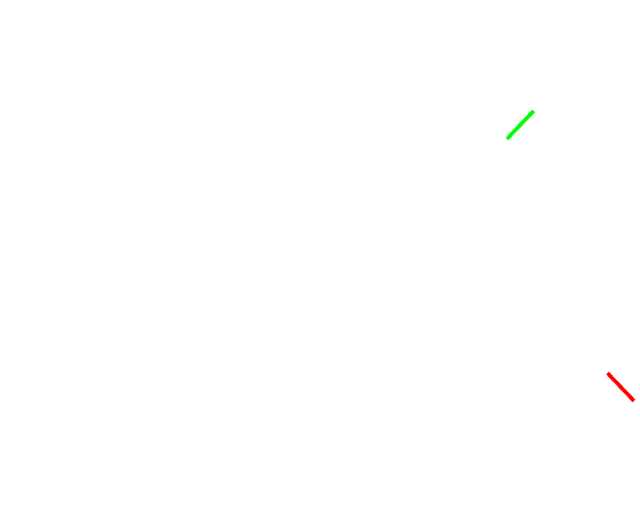}}
\\\makebox[1em]{7)}
\fbox{\includegraphics[width=.12\linewidth]{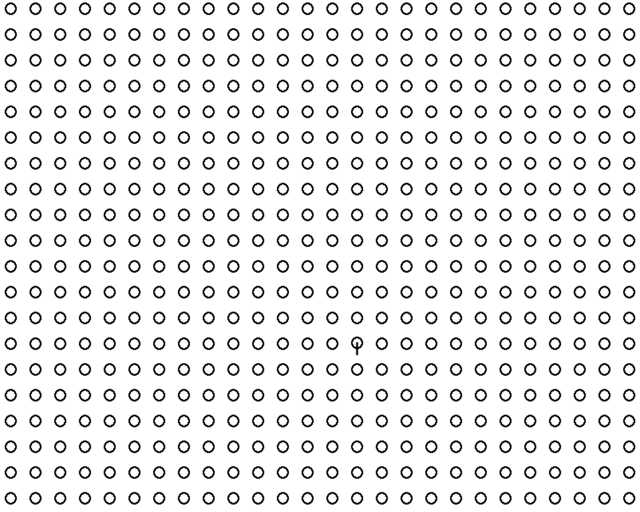}}
\fbox{\includegraphics[width=.12\linewidth]{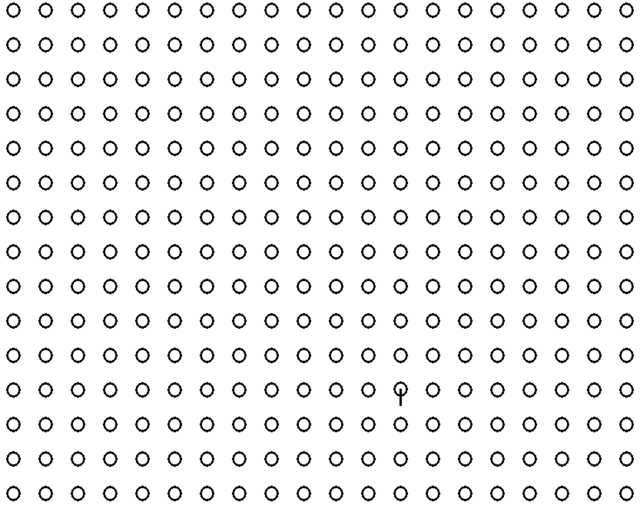}}
\fbox{\includegraphics[width=.12\linewidth]{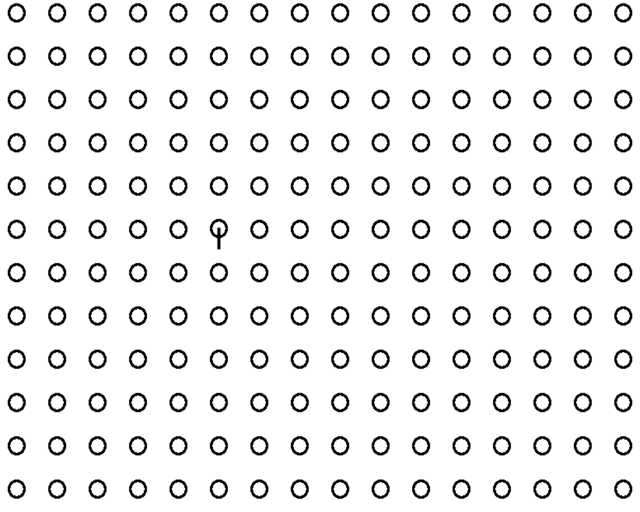}}
\fbox{\includegraphics[width=.12\linewidth]{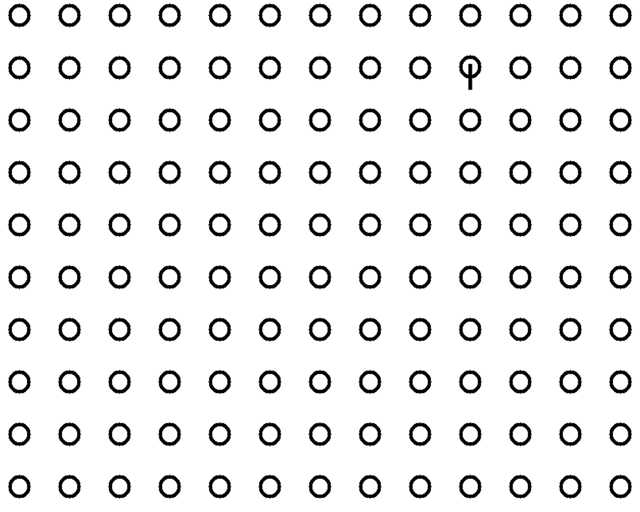}}
\fbox{\includegraphics[width=.12\linewidth]{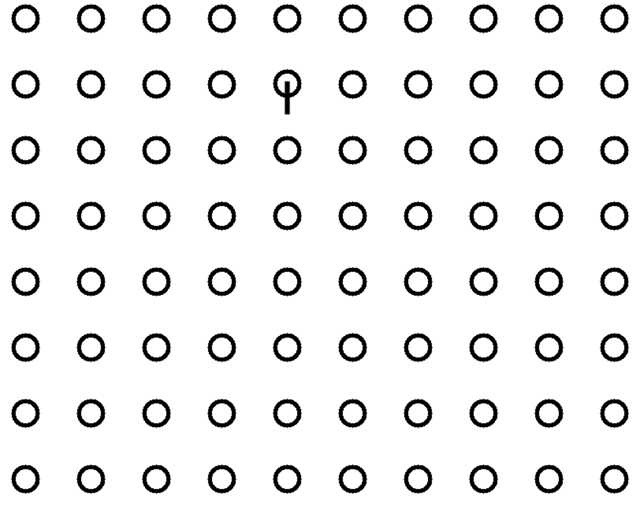}}
\fbox{\includegraphics[width=.12\linewidth]{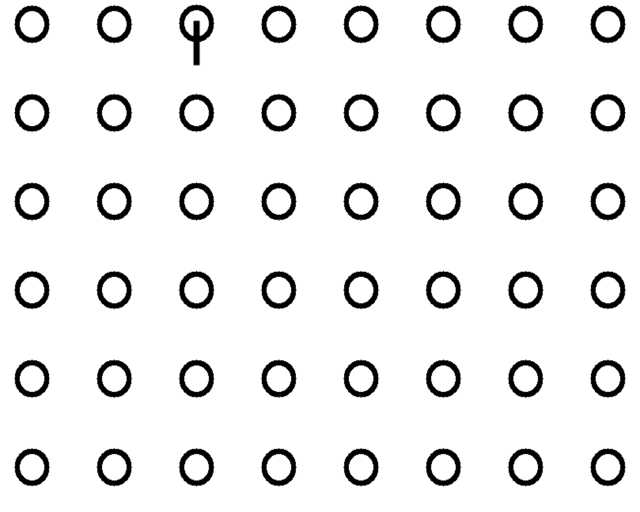}}
\fbox{\includegraphics[width=.12\linewidth]{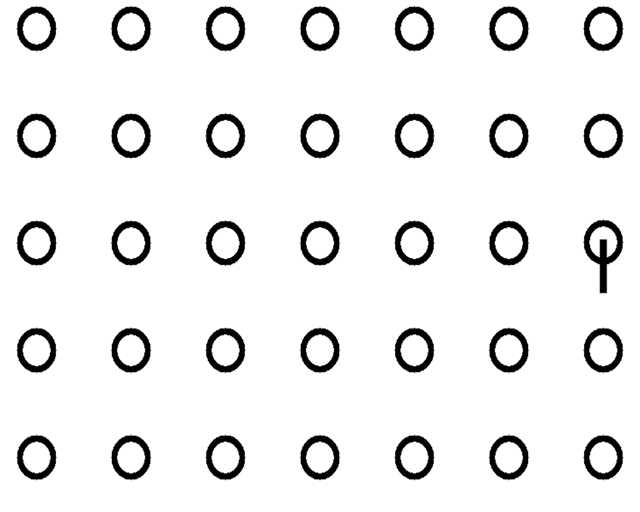}}
\\\hspace{1em}
\fbox{\includegraphics[width=.12\linewidth]{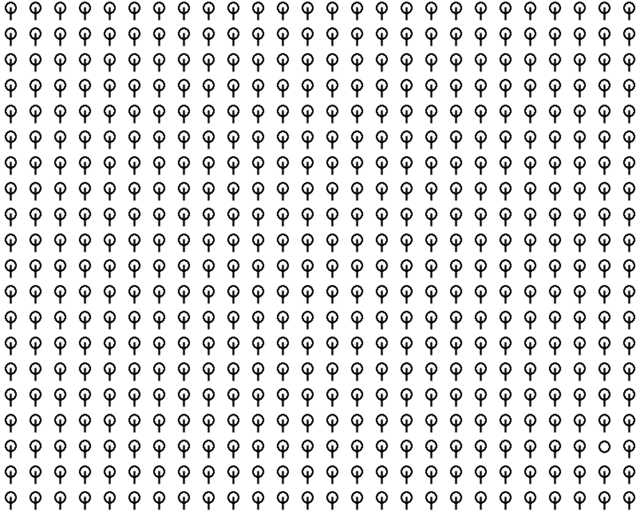}}
\fbox{\includegraphics[width=.12\linewidth]{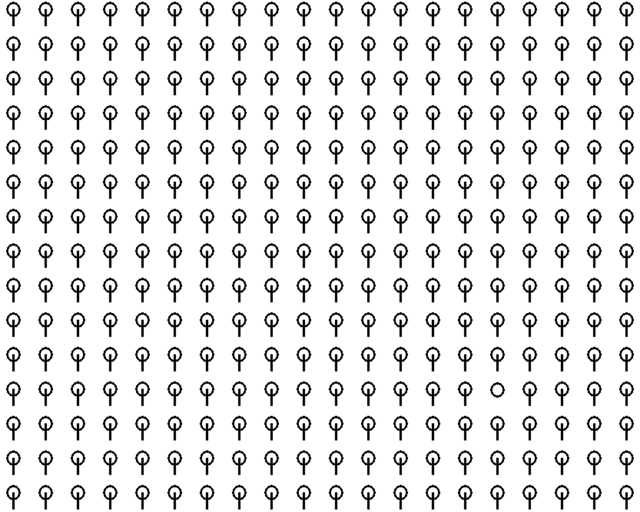}}
\fbox{\includegraphics[width=.12\linewidth]{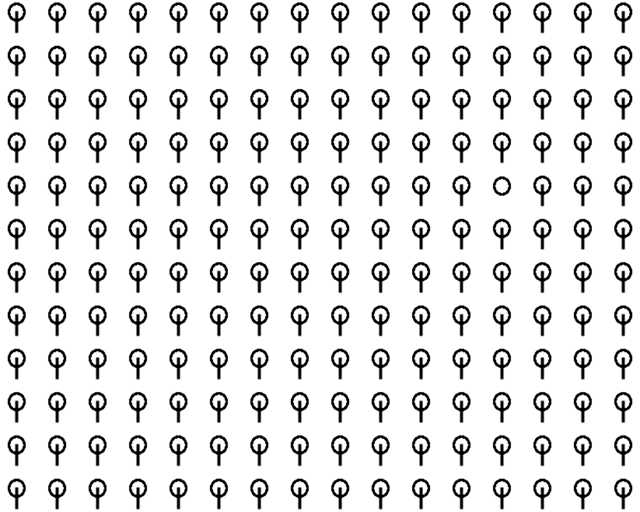}}
\fbox{\includegraphics[width=.12\linewidth]{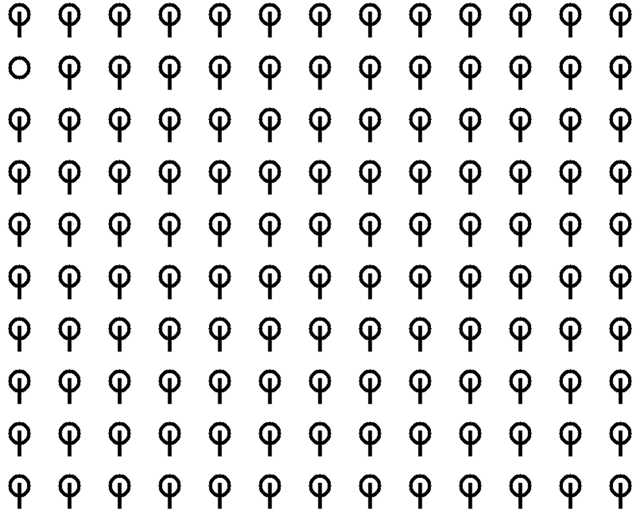}}
\fbox{\includegraphics[width=.12\linewidth]{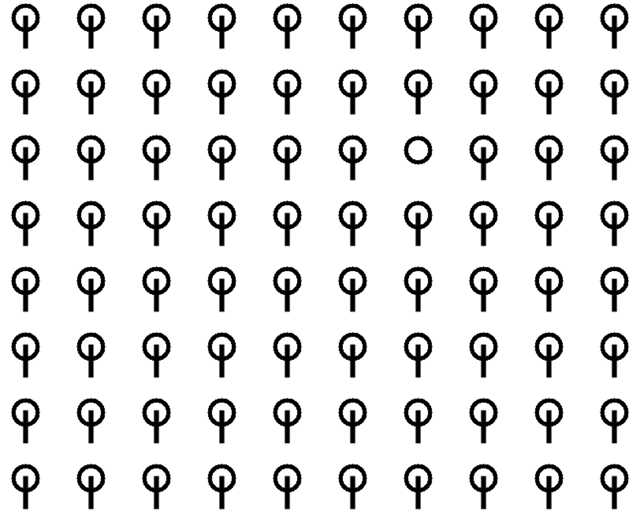}}
\fbox{\includegraphics[width=.12\linewidth]{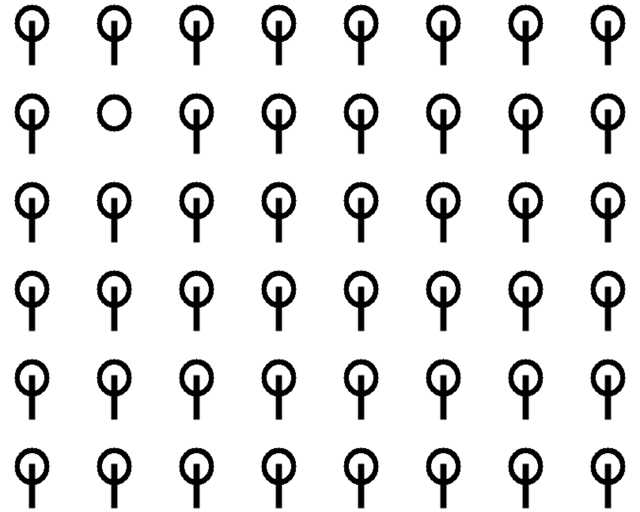}}
\fbox{\includegraphics[width=.12\linewidth]{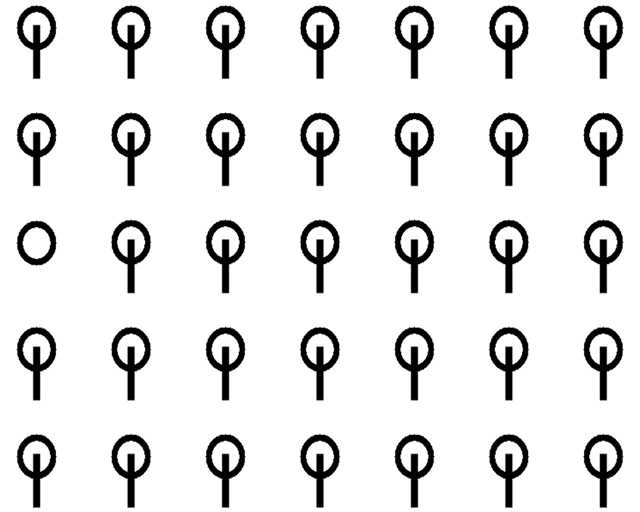}}
\\\makebox[1em]{8)}
\fbox{\includegraphics[width=.12\linewidth]{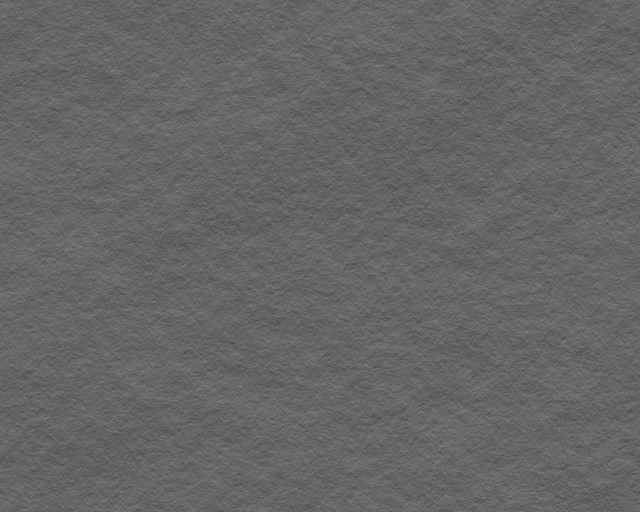}}
\fbox{\includegraphics[width=.12\linewidth]{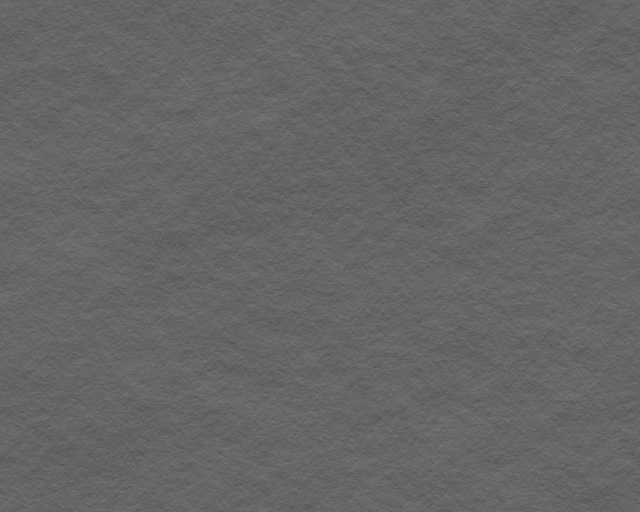}}
\fbox{\includegraphics[width=.12\linewidth]{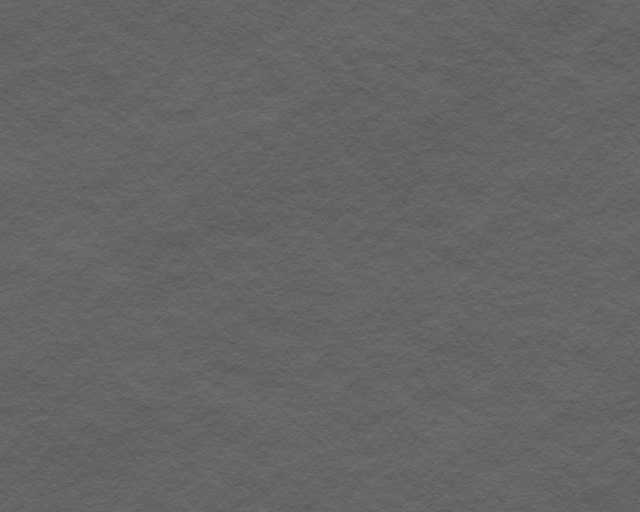}}
\fbox{\includegraphics[width=.12\linewidth]{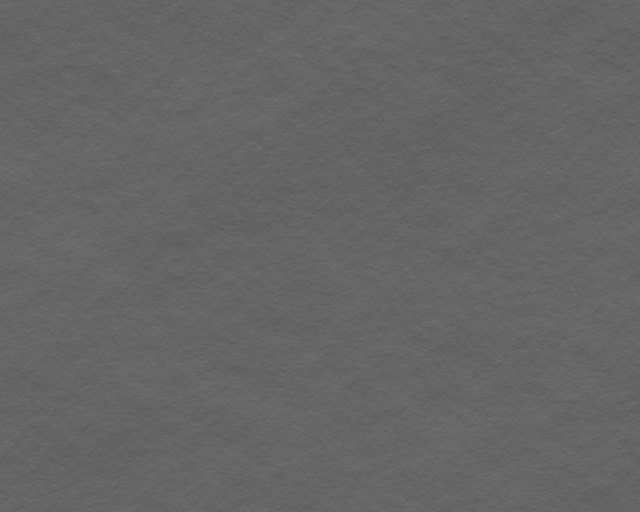}}
\fbox{\includegraphics[width=.12\linewidth]{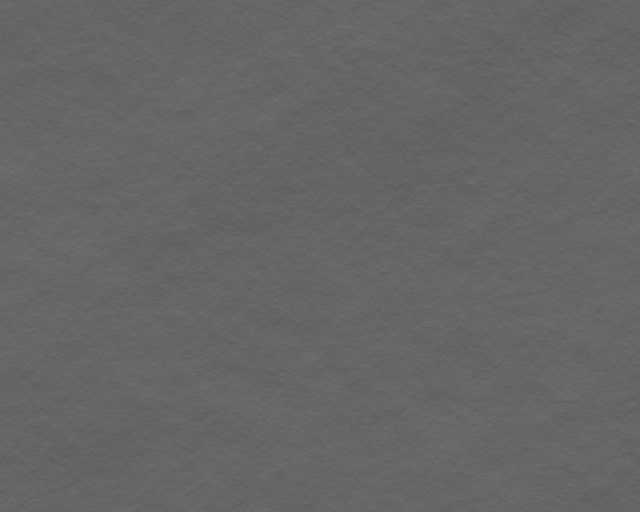}}
\fbox{\includegraphics[width=.12\linewidth]{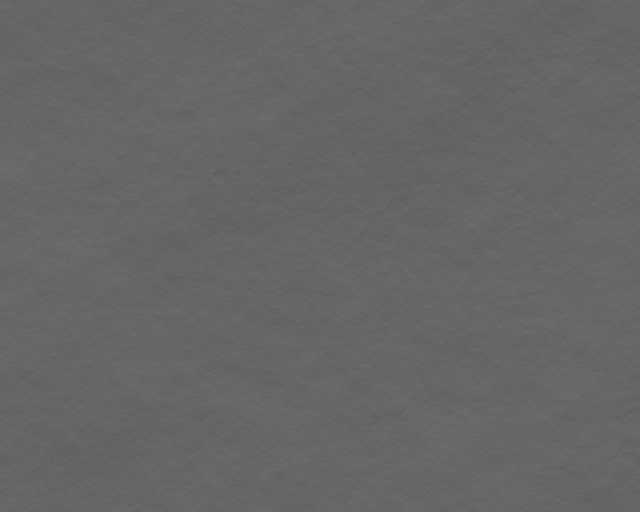}}
\fbox{\includegraphics[width=.12\linewidth]{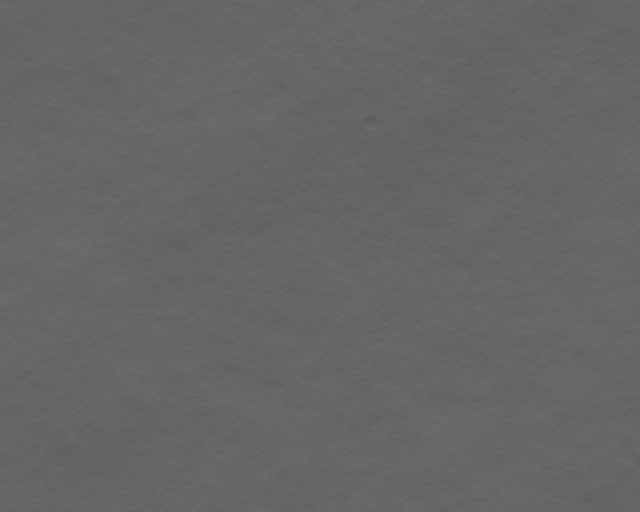}}
\\\hspace{1em}
\fbox{\includegraphics[width=.12\linewidth]{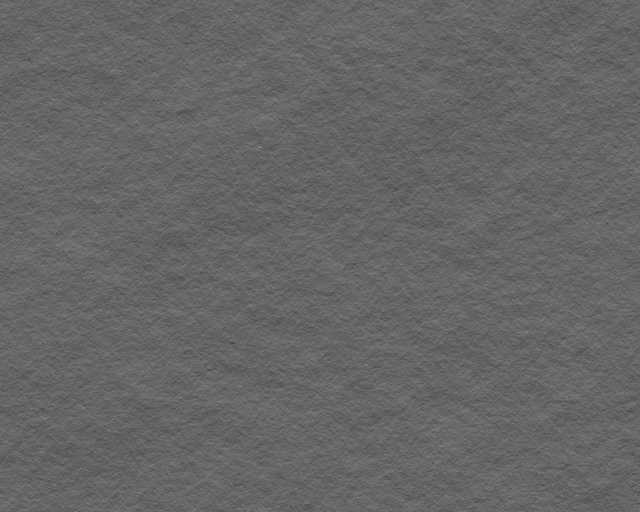}}
\fbox{\includegraphics[width=.12\linewidth]{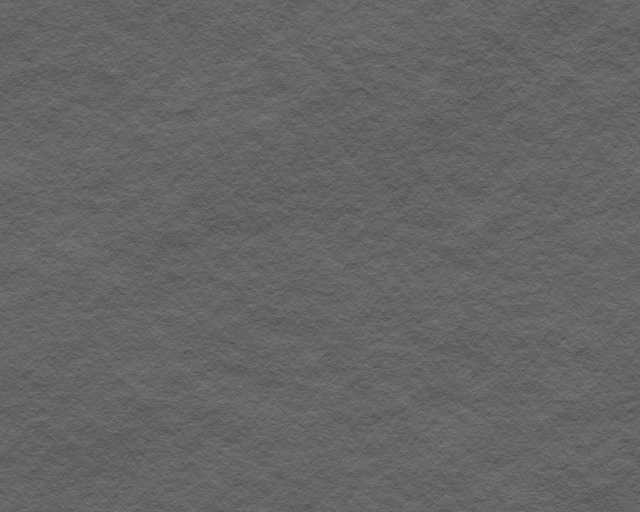}}
\fbox{\includegraphics[width=.12\linewidth]{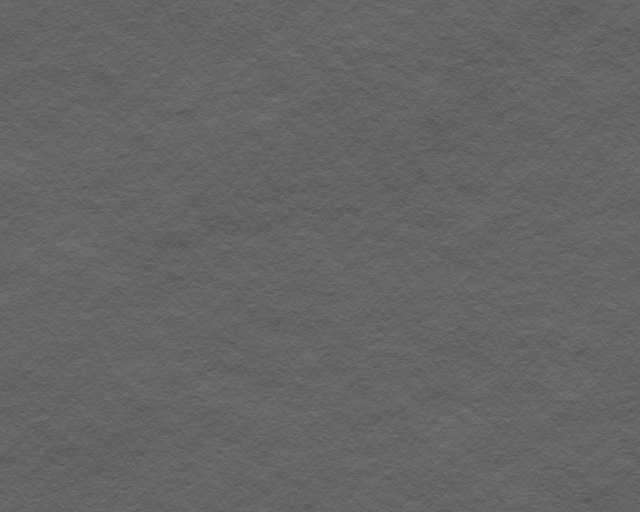}}
\fbox{\includegraphics[width=.12\linewidth]{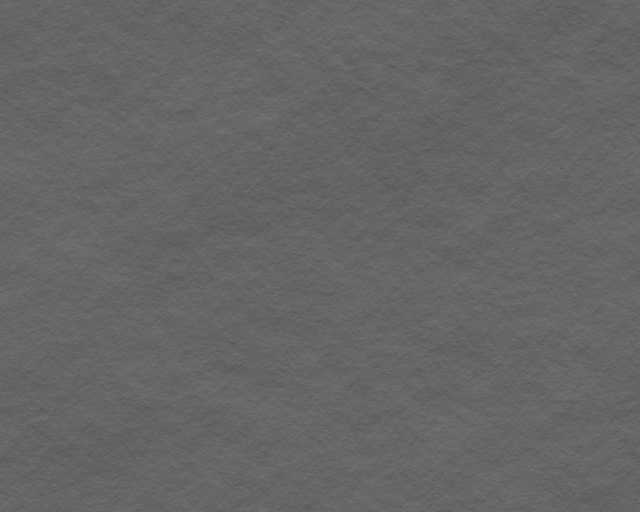}}
\fbox{\includegraphics[width=.12\linewidth]{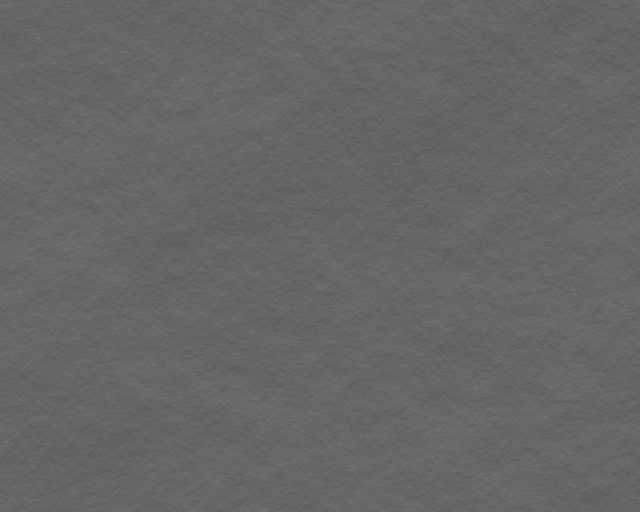}}
\fbox{\includegraphics[width=.12\linewidth]{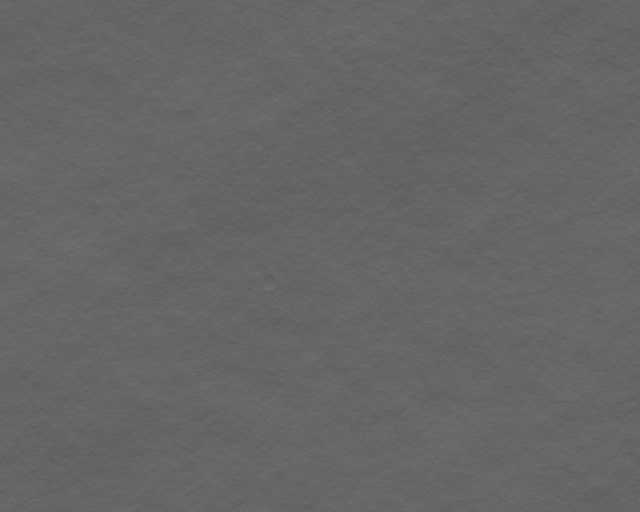}}
\fbox{\includegraphics[width=.12\linewidth]{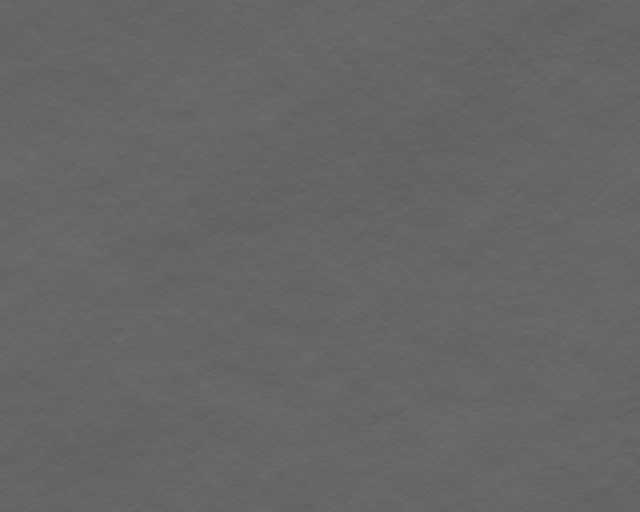}}
\\\makebox[1em]{9)}
\fbox{\includegraphics[width=.12\linewidth]{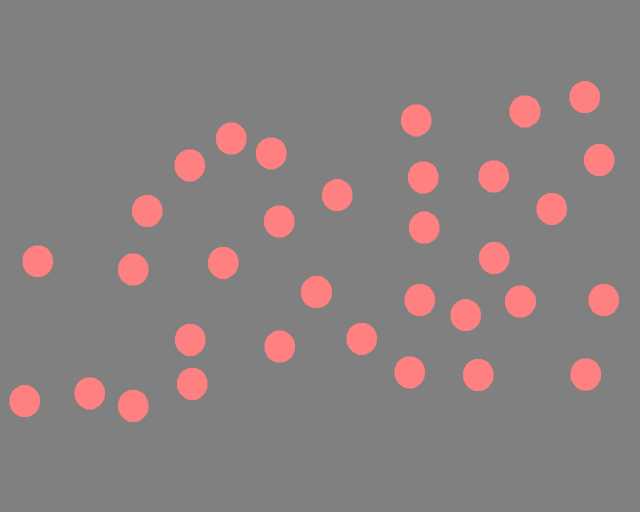}}
\fbox{\includegraphics[width=.12\linewidth]{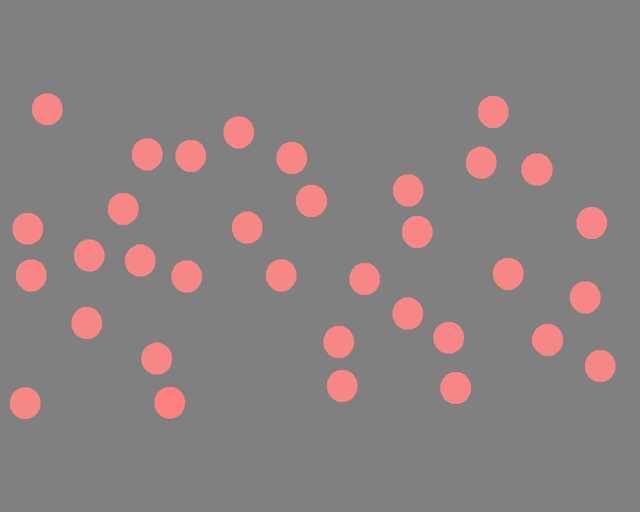}}
\fbox{\includegraphics[width=.12\linewidth]{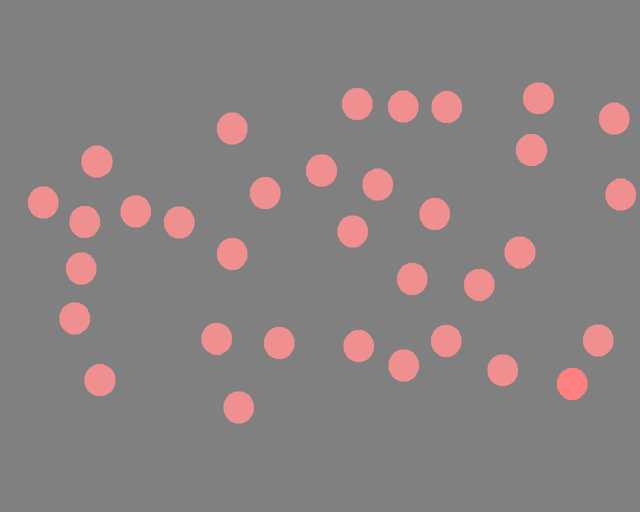}}
\fbox{\includegraphics[width=.12\linewidth]{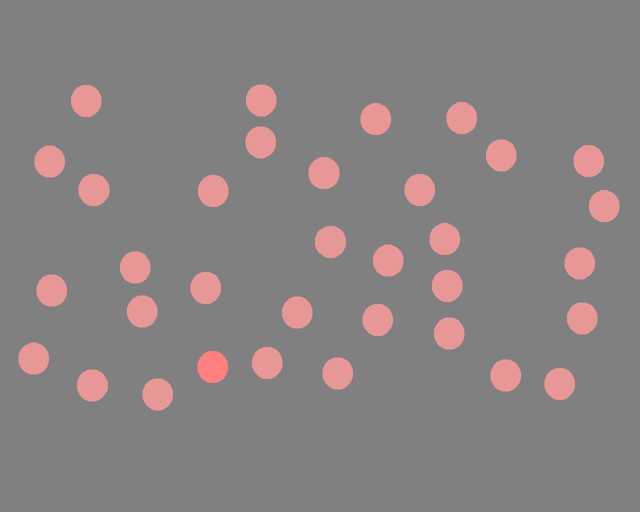}}
\fbox{\includegraphics[width=.12\linewidth]{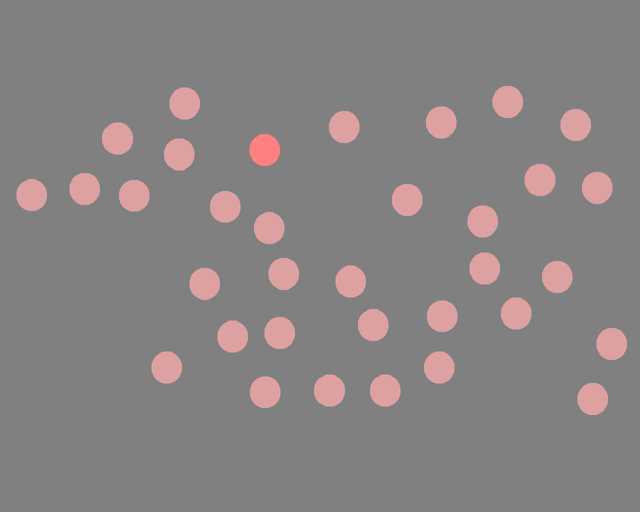}}
\fbox{\includegraphics[width=.12\linewidth]{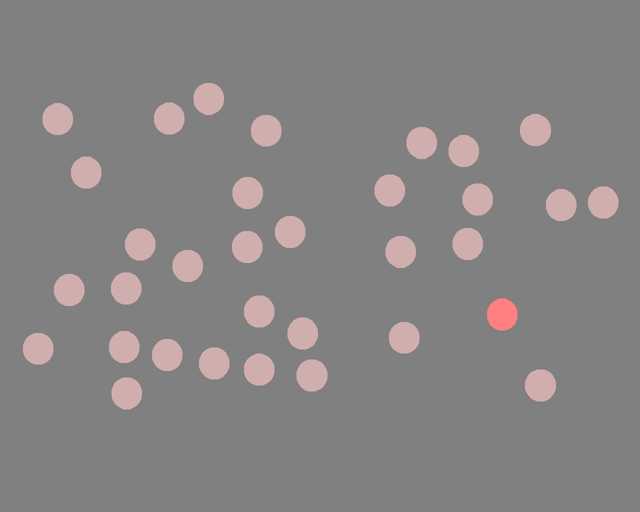}}
\fbox{\includegraphics[width=.12\linewidth]{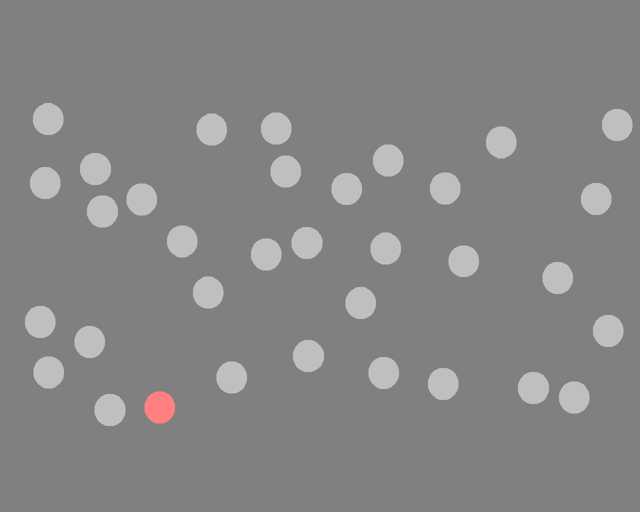}}
\\\hspace{1em}
\fbox{\includegraphics[width=.12\linewidth]{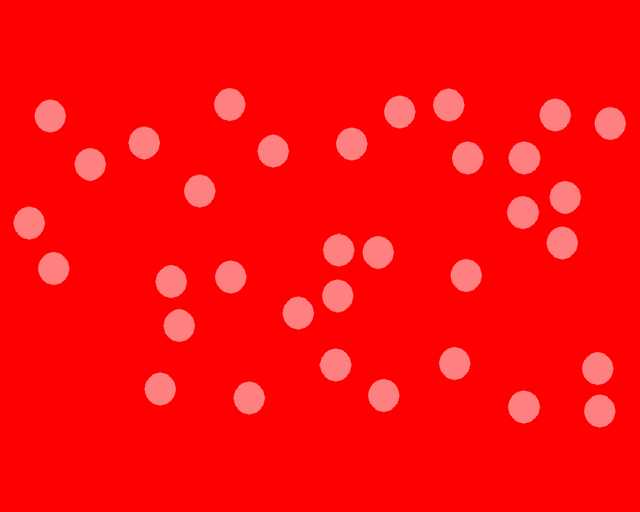}}
\fbox{\includegraphics[width=.12\linewidth]{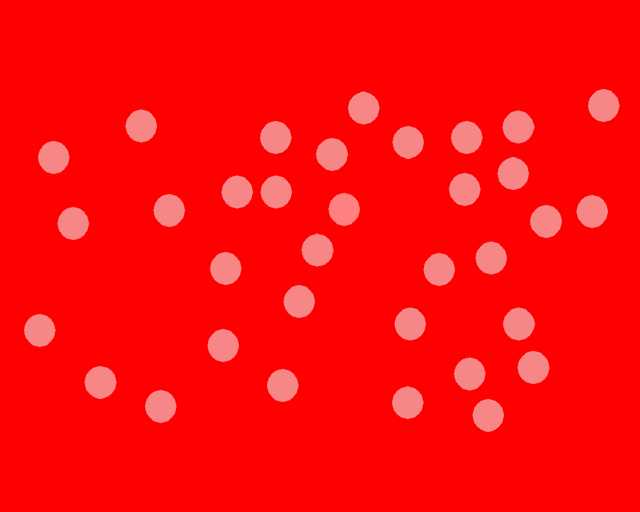}}
\fbox{\includegraphics[width=.12\linewidth]{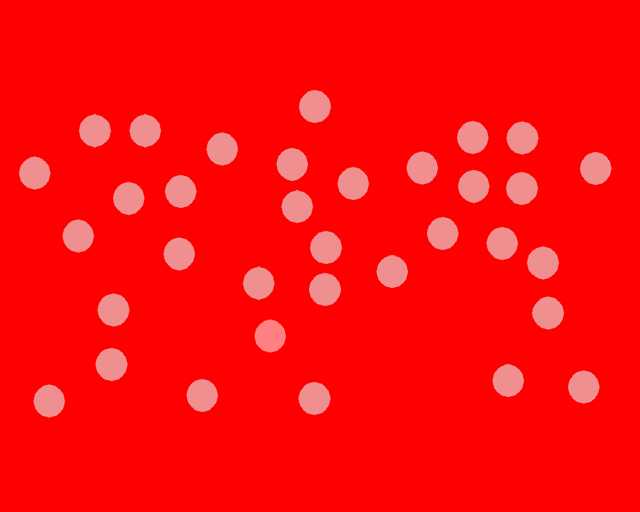}}
\fbox{\includegraphics[width=.12\linewidth]{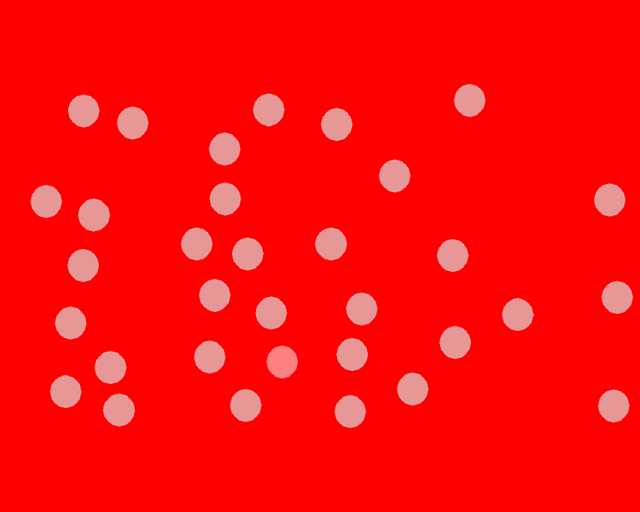}}
\fbox{\includegraphics[width=.12\linewidth]{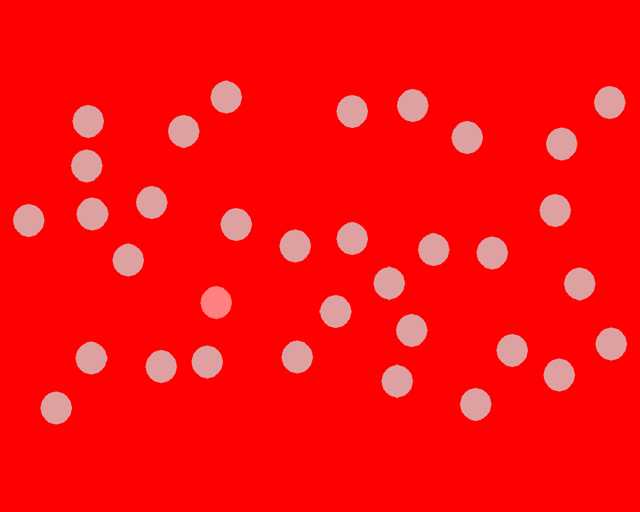}}
\fbox{\includegraphics[width=.12\linewidth]{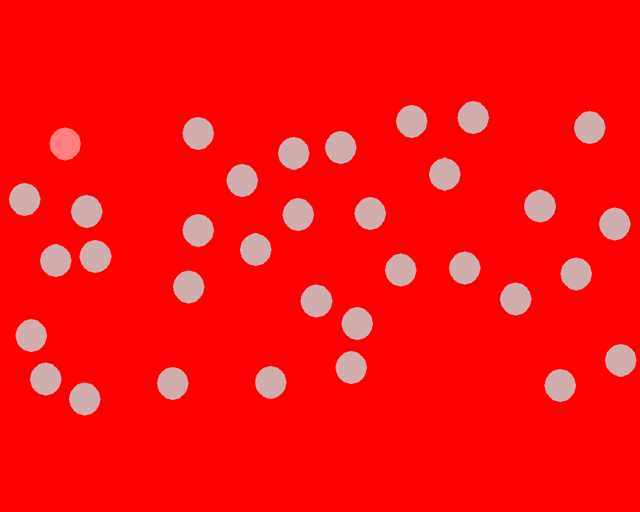}}
\fbox{\includegraphics[width=.12\linewidth]{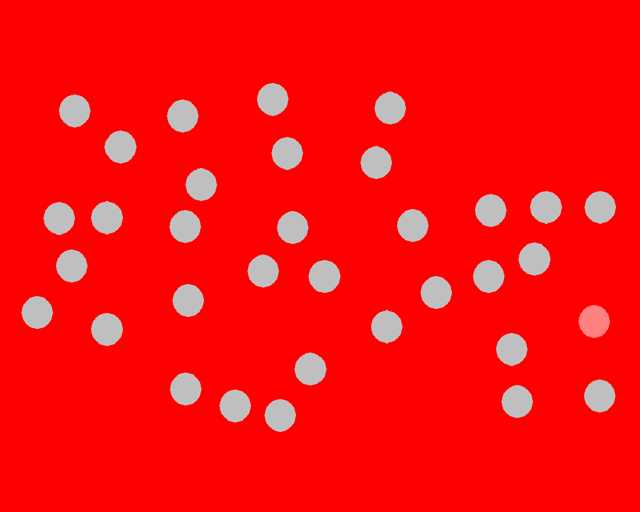}}
\\\hspace{1em}
\fbox{\includegraphics[width=.12\linewidth]{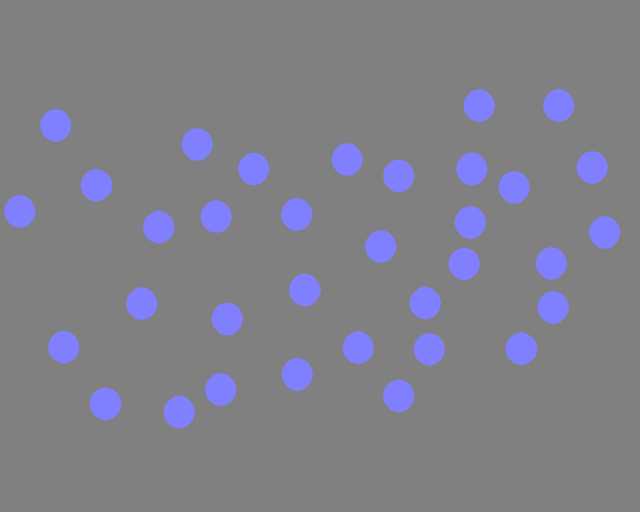}}
\fbox{\includegraphics[width=.12\linewidth]{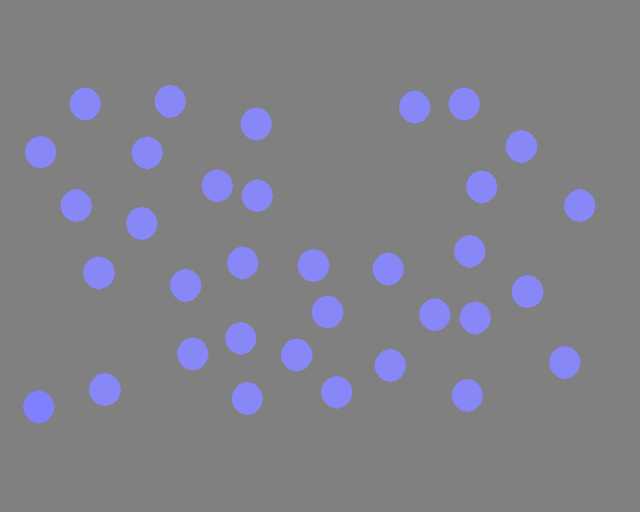}}
\fbox{\includegraphics[width=.12\linewidth]{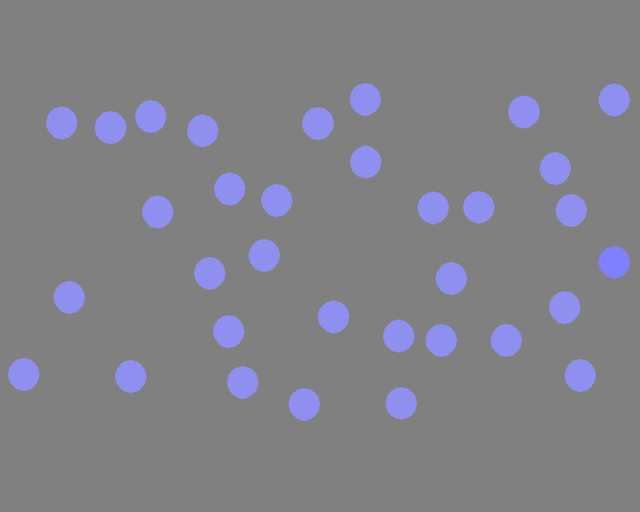}}
\fbox{\includegraphics[width=.12\linewidth]{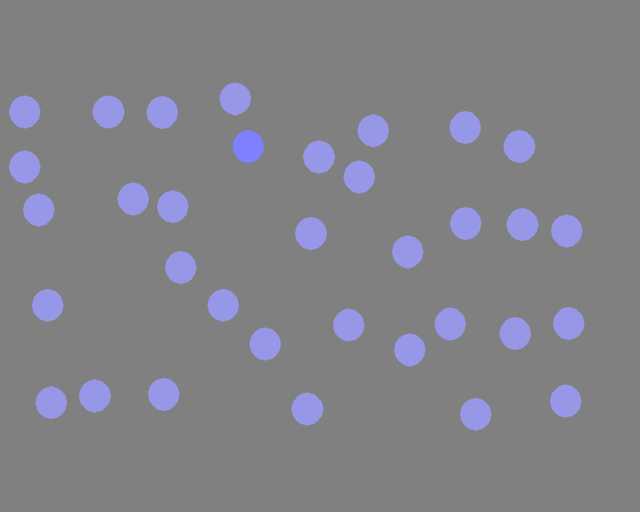}}
\fbox{\includegraphics[width=.12\linewidth]{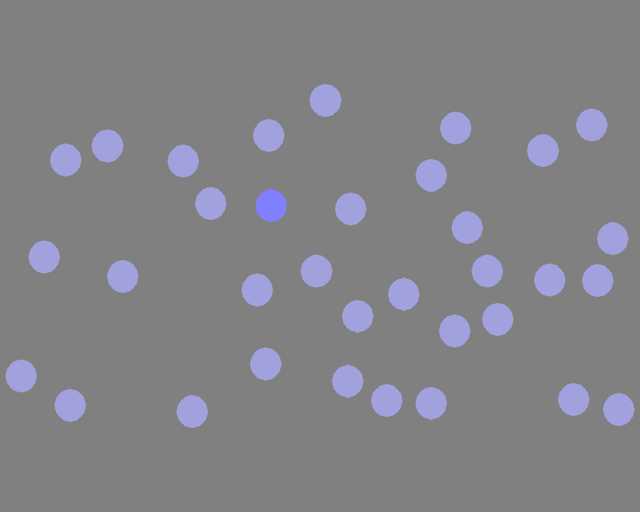}}
\fbox{\includegraphics[width=.12\linewidth]{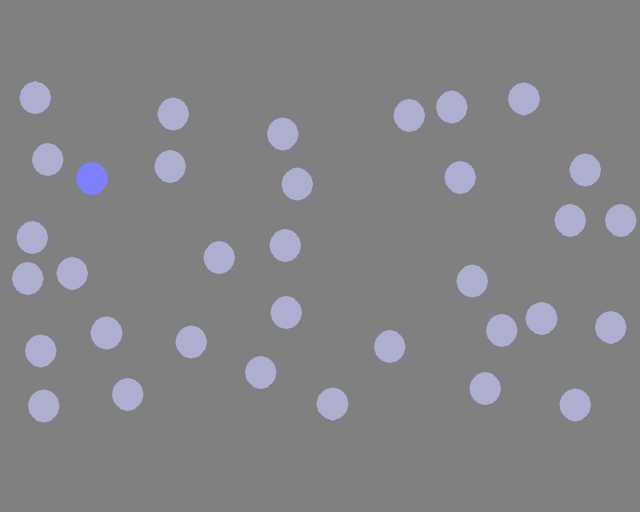}}
\fbox{\includegraphics[width=.12\linewidth]{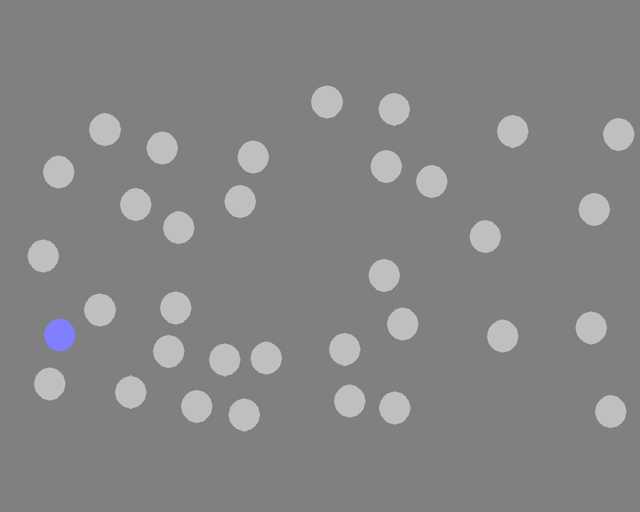}} 
\\\hspace{1em}
\fbox{\includegraphics[width=.12\linewidth]{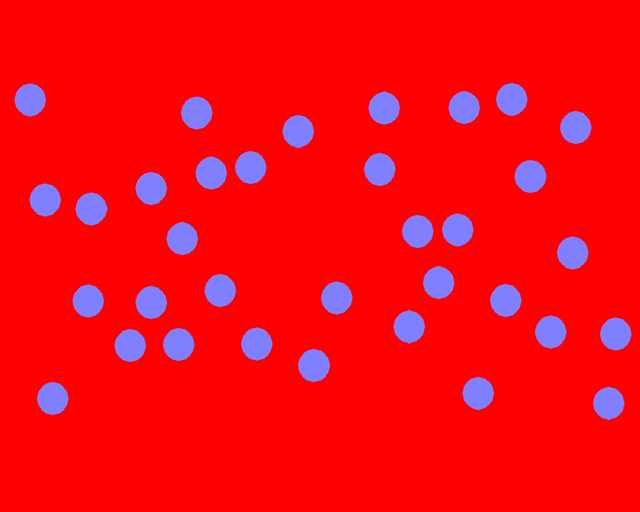}}
\fbox{\includegraphics[width=.12\linewidth]{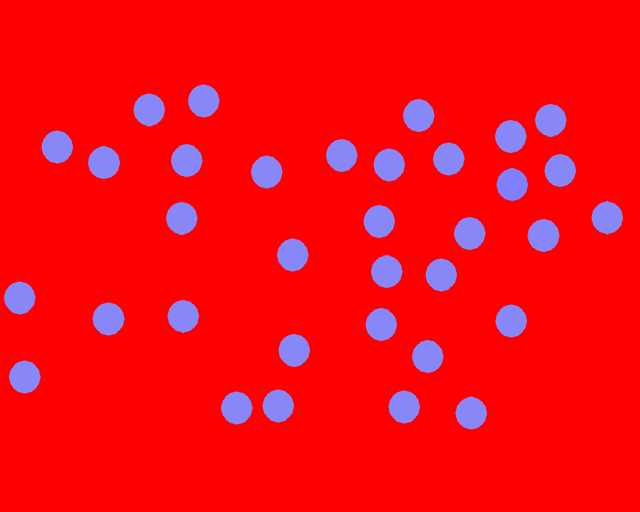}}
\fbox{\includegraphics[width=.12\linewidth]{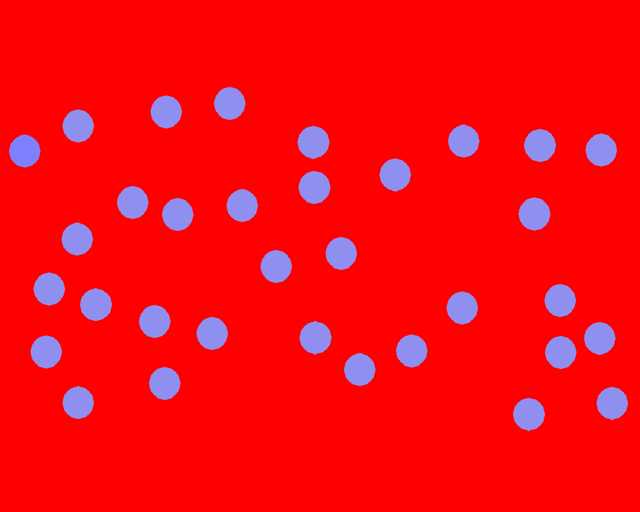}}
\fbox{\includegraphics[width=.12\linewidth]{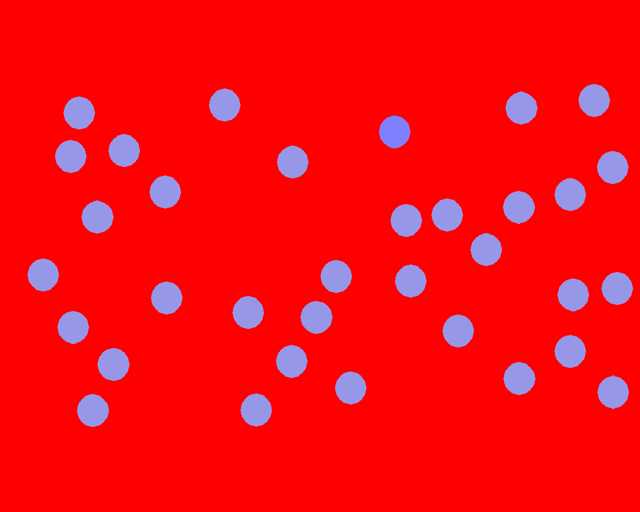}}
\fbox{\includegraphics[width=.12\linewidth]{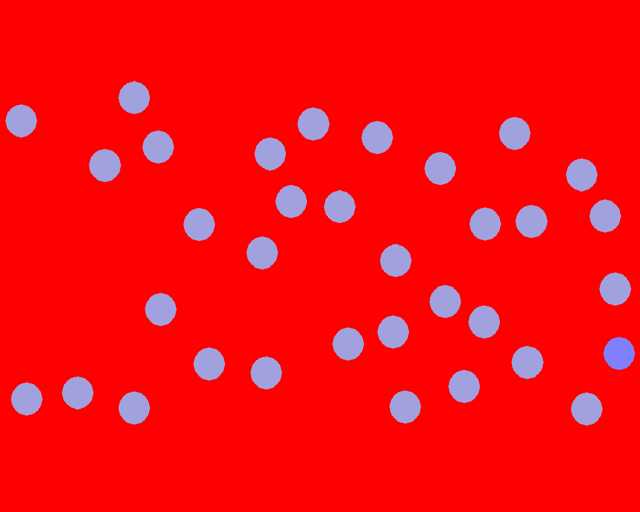}}
\fbox{\includegraphics[width=.12\linewidth]{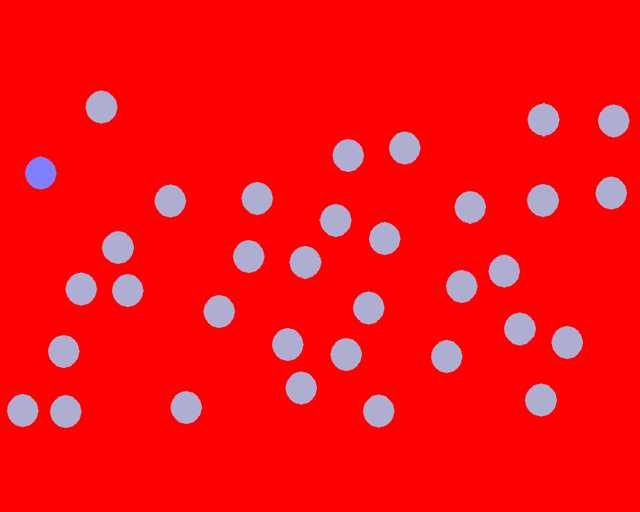}}
\fbox{\includegraphics[width=.12\linewidth]{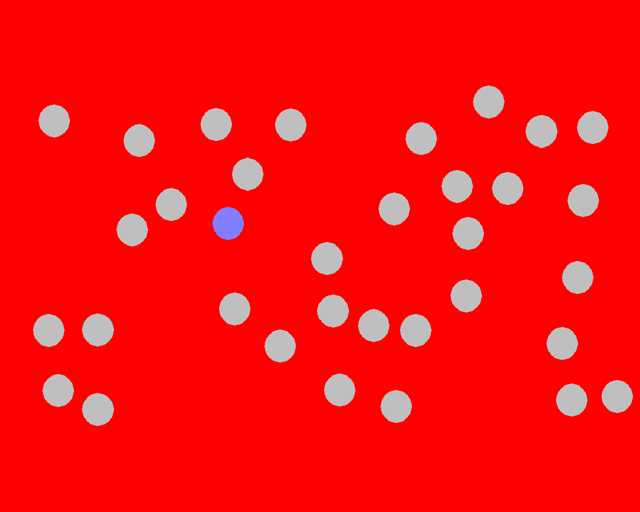}}
\\\makebox[1em]{10)}
\fbox{\includegraphics[width=.12\linewidth]{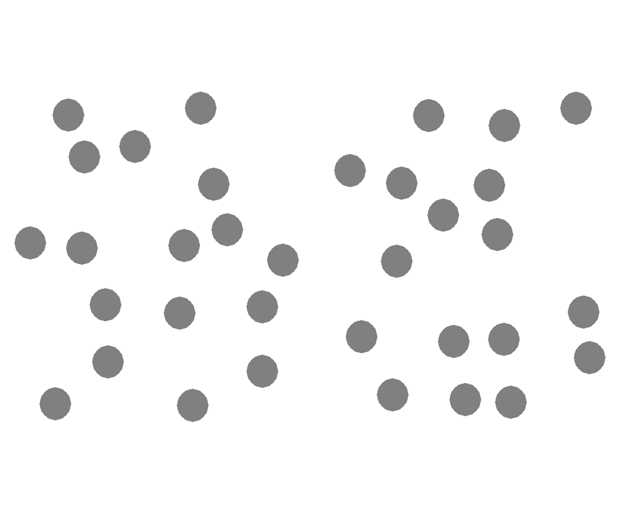}}
\fbox{\includegraphics[width=.12\linewidth]{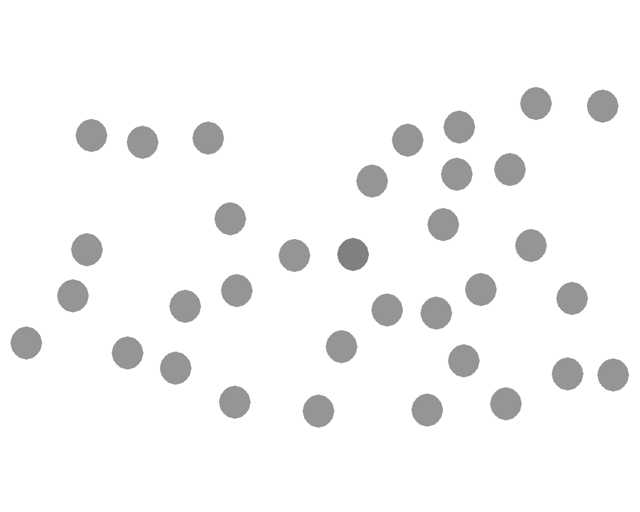}}
\fbox{\includegraphics[width=.12\linewidth]{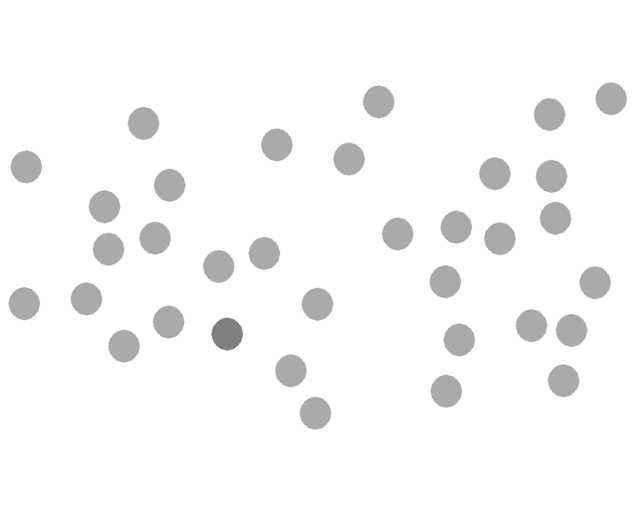}}
\fbox{\includegraphics[width=.12\linewidth]{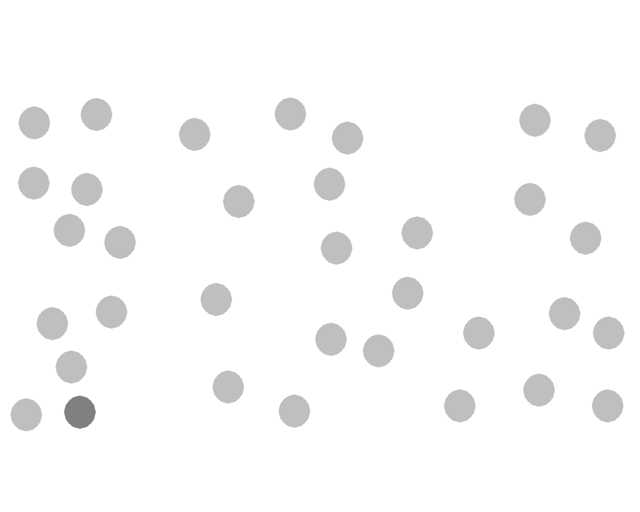}}
\fbox{\includegraphics[width=.12\linewidth]{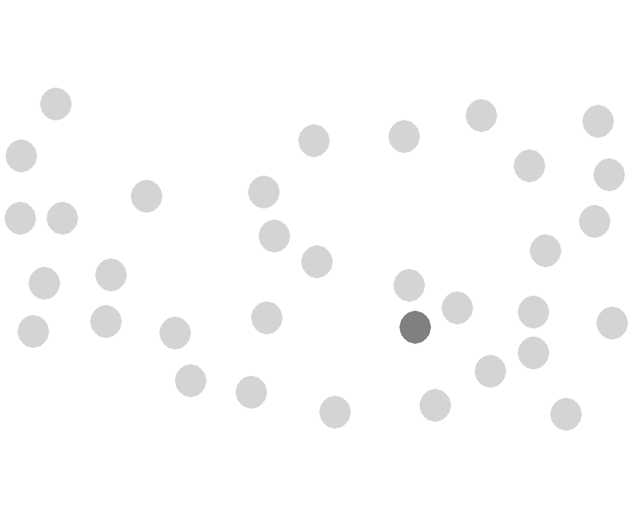}}
\fbox{\includegraphics[width=.12\linewidth]{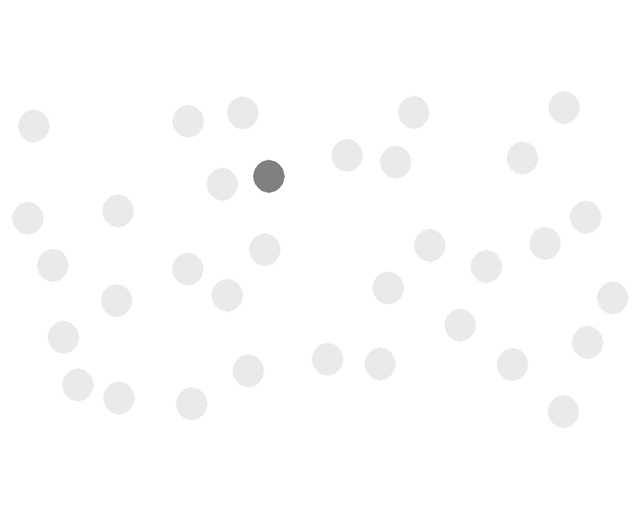}}
\fbox{\includegraphics[width=.12\linewidth]{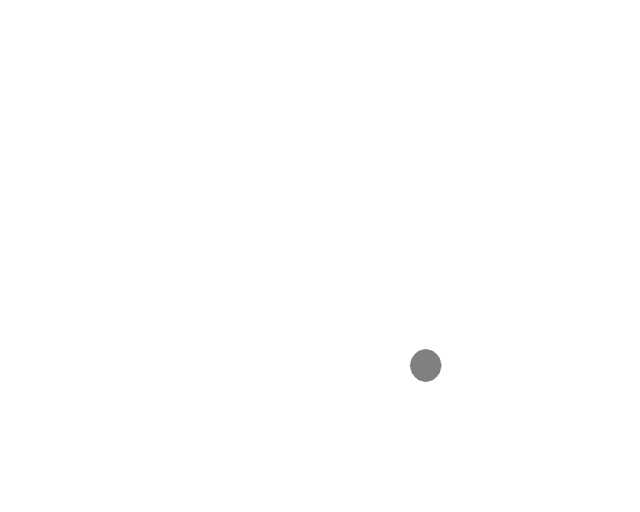}}
\\\hspace{1em}
\fbox{\includegraphics[width=.12\linewidth]{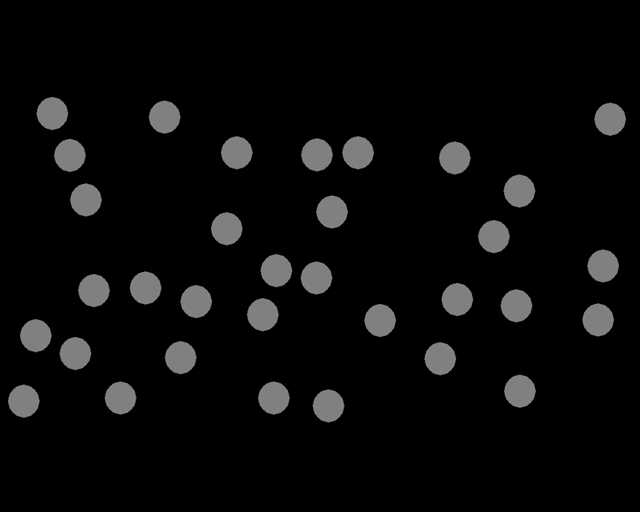}}
\fbox{\includegraphics[width=.12\linewidth]{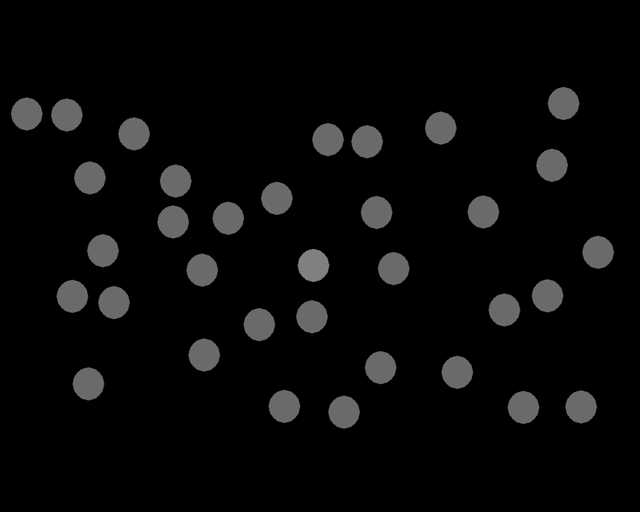}}
\fbox{\includegraphics[width=.12\linewidth]{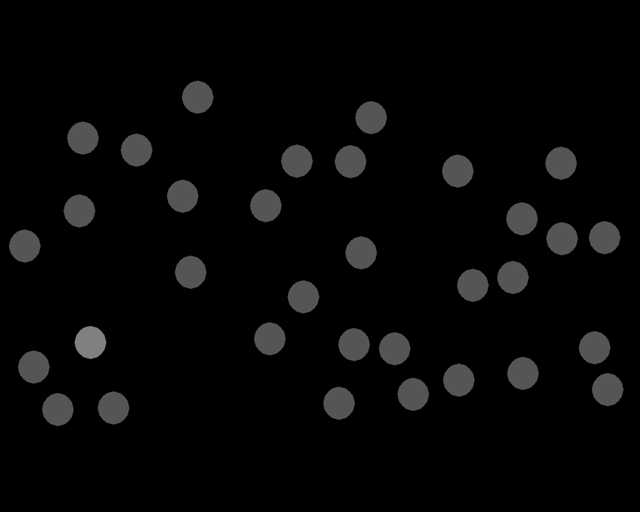}}
\fbox{\includegraphics[width=.12\linewidth]{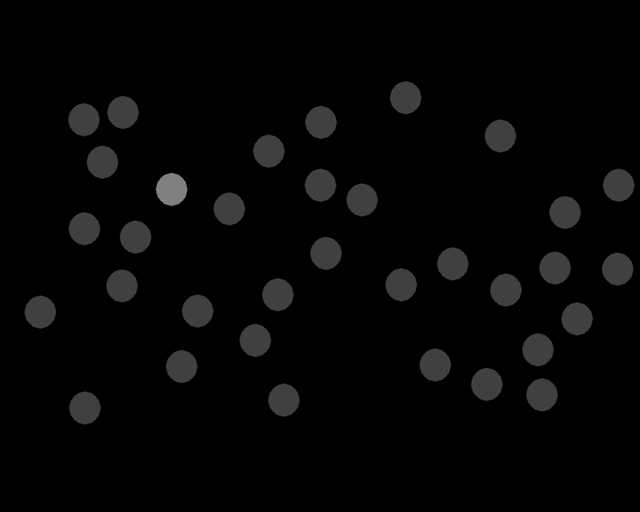}}
\fbox{\includegraphics[width=.12\linewidth]{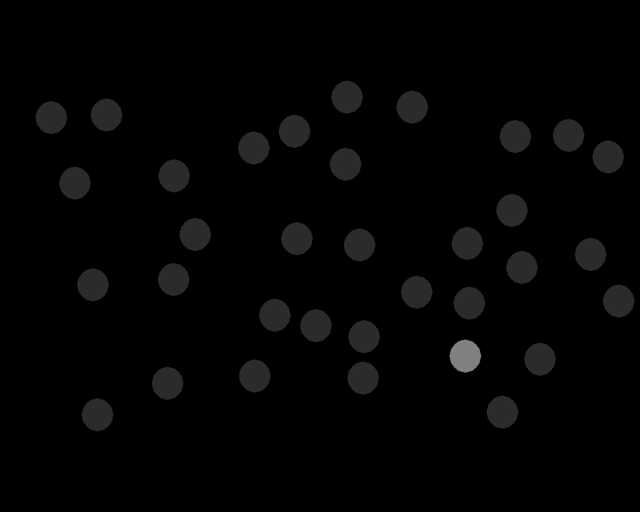}}
\fbox{\includegraphics[width=.12\linewidth]{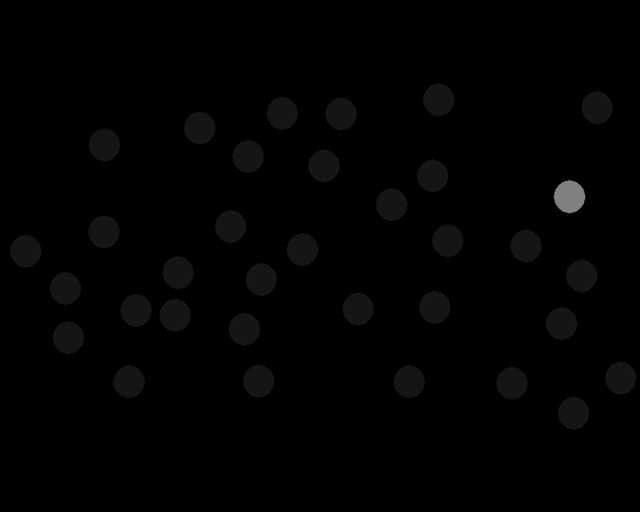}}
\fbox{\includegraphics[width=.12\linewidth]{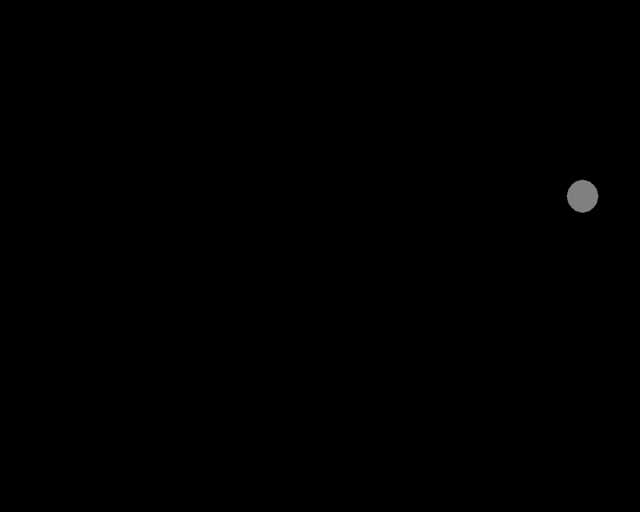}}
\\\makebox[1em]{11)}
\fbox{\includegraphics[width=.12\linewidth]{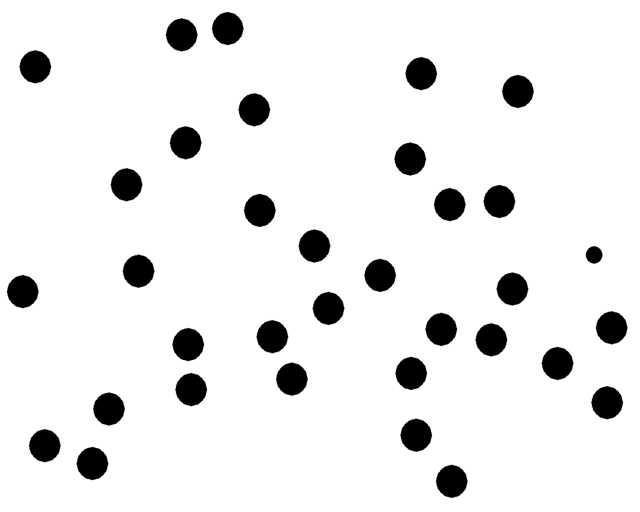}}
\fbox{\includegraphics[width=.12\linewidth]{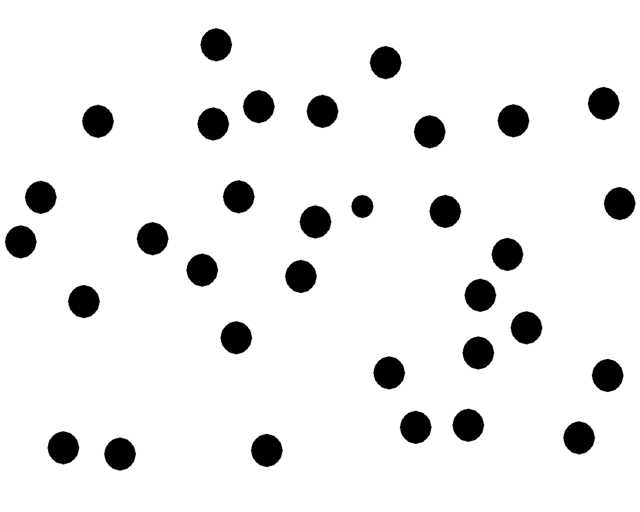}}
\fbox{\includegraphics[width=.12\linewidth]{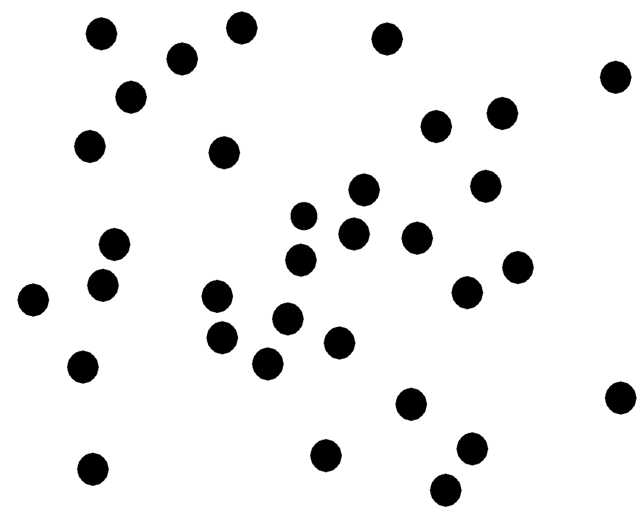}}
\fbox{\includegraphics[width=.12\linewidth]{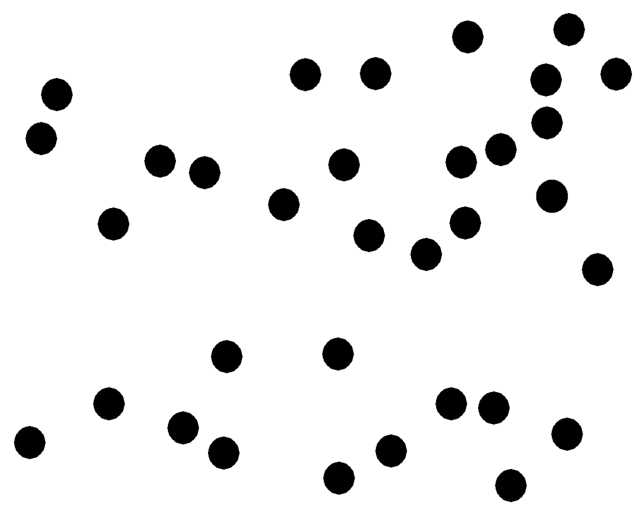}}
\fbox{\includegraphics[width=.12\linewidth]{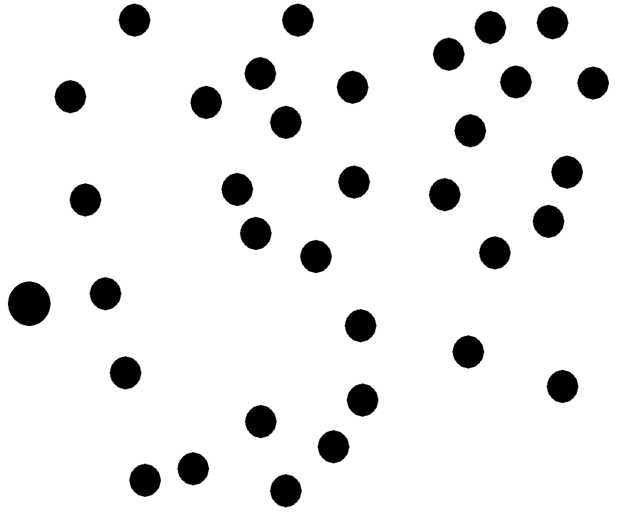}}
\fbox{\includegraphics[width=.12\linewidth]{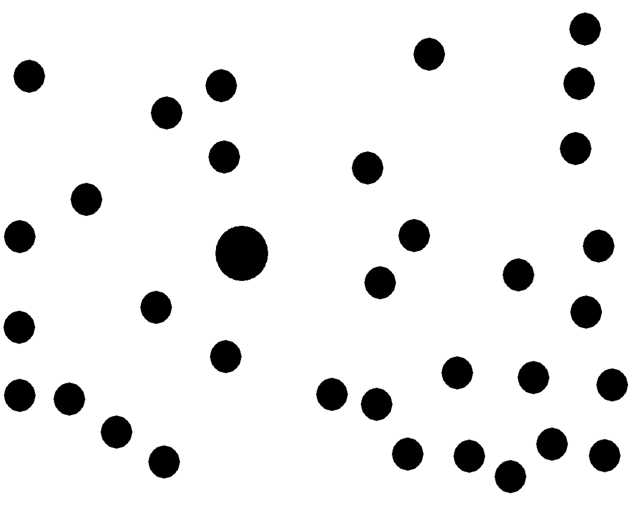}}
\fbox{\includegraphics[width=.12\linewidth]{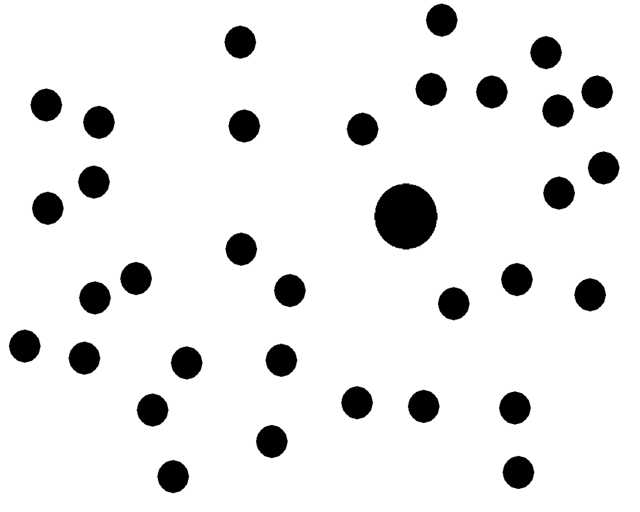}}
\\\makebox[1em]{12)}
\fbox{\includegraphics[width=.12\linewidth]{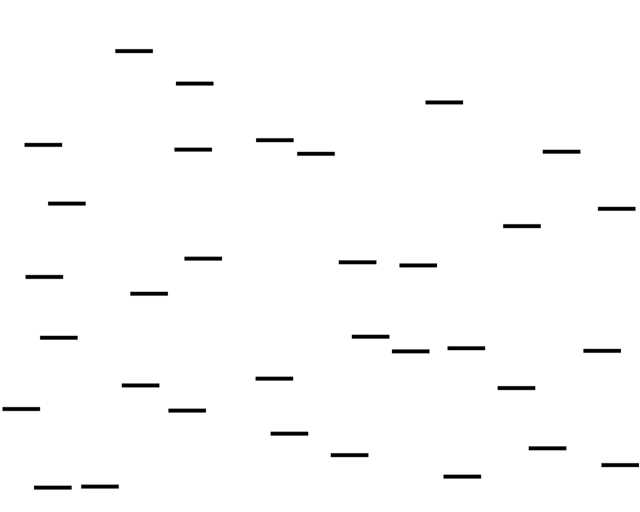}}
\fbox{\includegraphics[width=.12\linewidth]{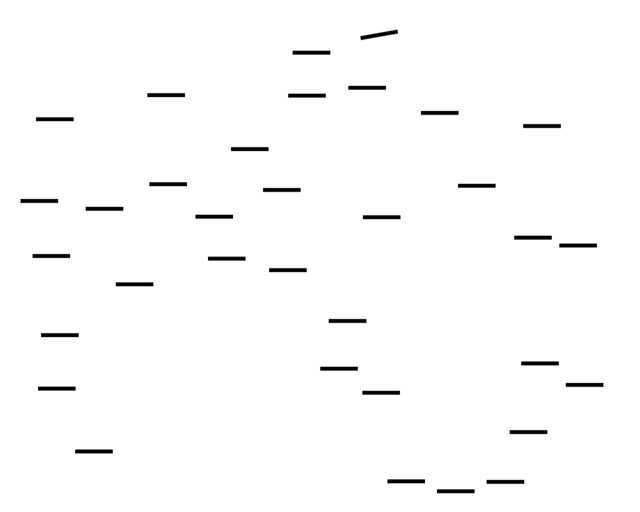}}
\fbox{\includegraphics[width=.12\linewidth]{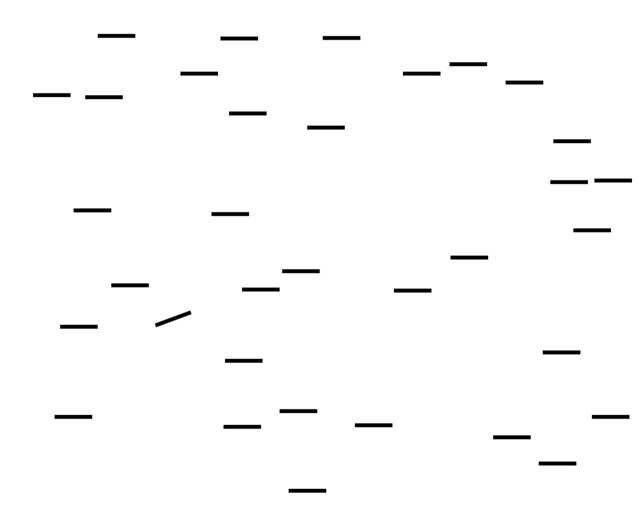}}
\fbox{\includegraphics[width=.12\linewidth]{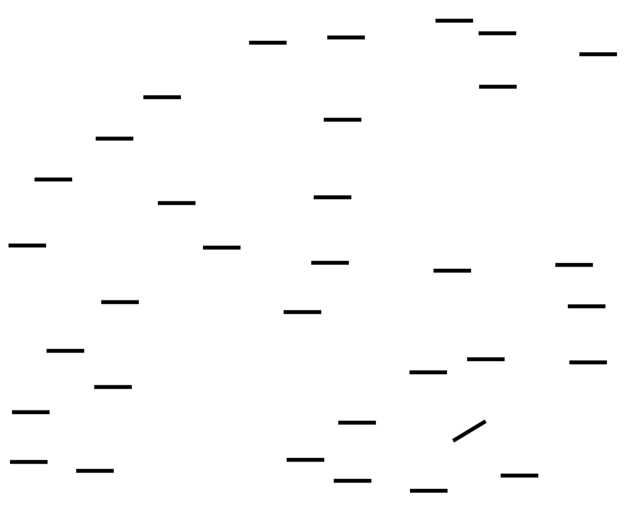}}
\fbox{\includegraphics[width=.12\linewidth]{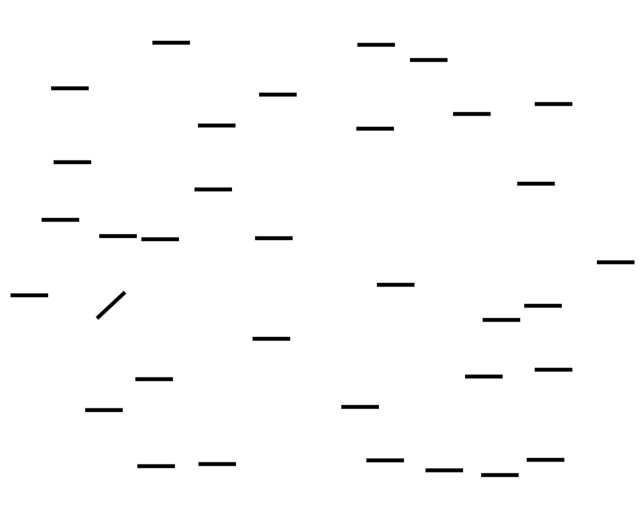}}
\fbox{\includegraphics[width=.12\linewidth]{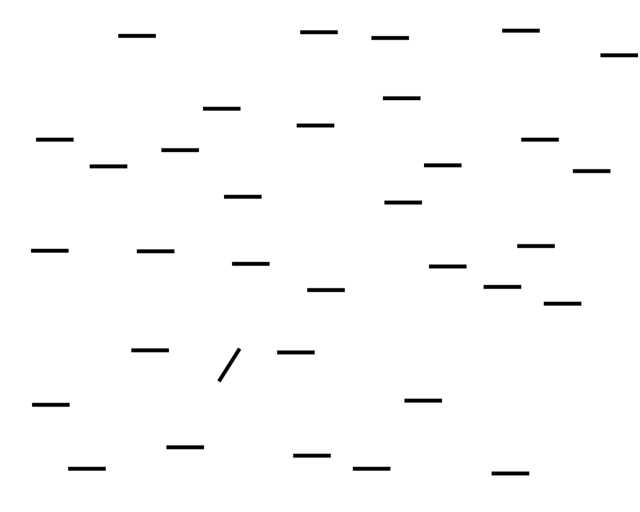}}
\fbox{\includegraphics[width=.12\linewidth]{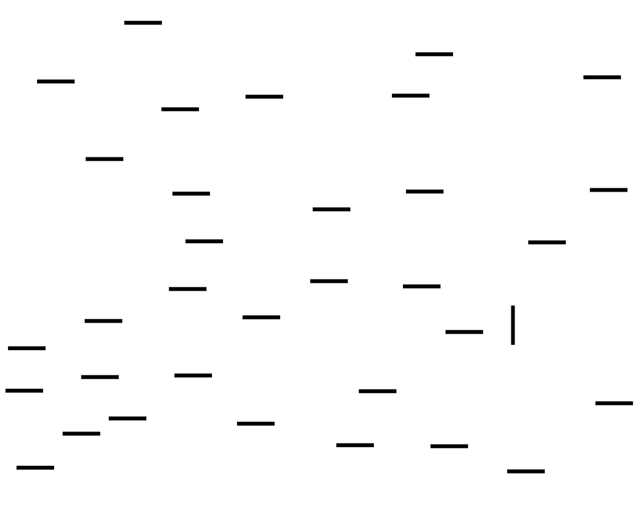}}
\\\makebox[1em]{13)}
\fbox{\includegraphics[width=.12\linewidth]{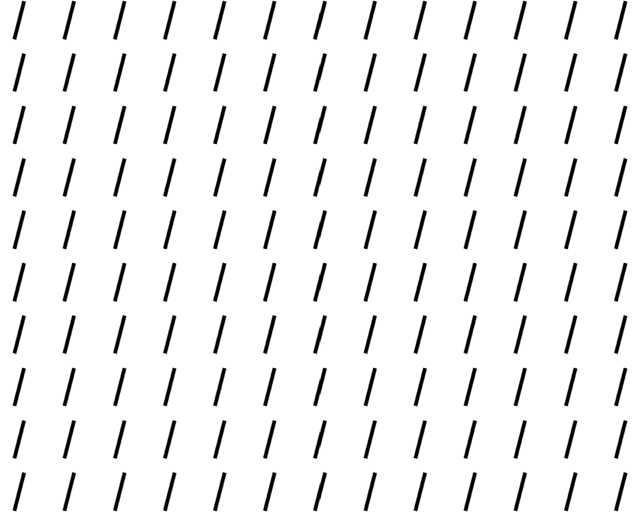}}
\fbox{\includegraphics[width=.12\linewidth]{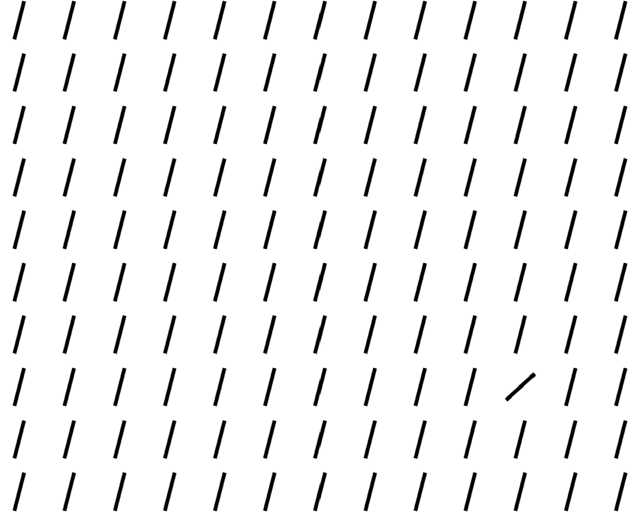}}
\fbox{\includegraphics[width=.12\linewidth]{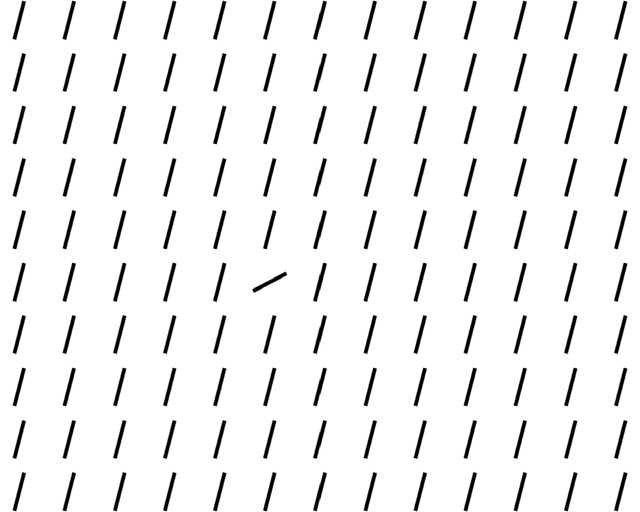}}
\fbox{\includegraphics[width=.12\linewidth]{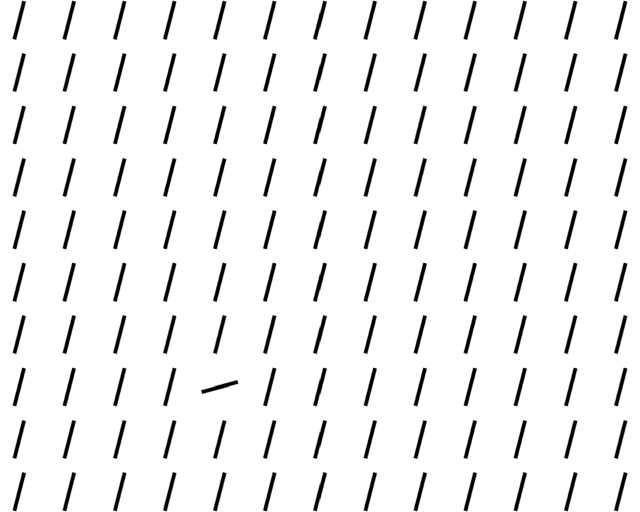}}
\fbox{\includegraphics[width=.12\linewidth]{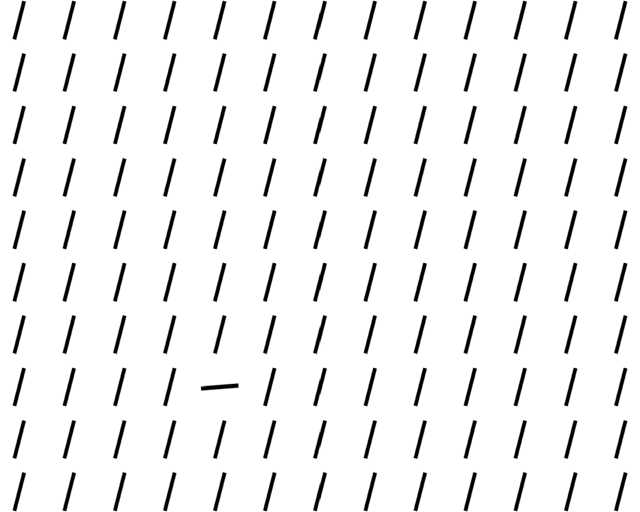}}
\fbox{\includegraphics[width=.12\linewidth]{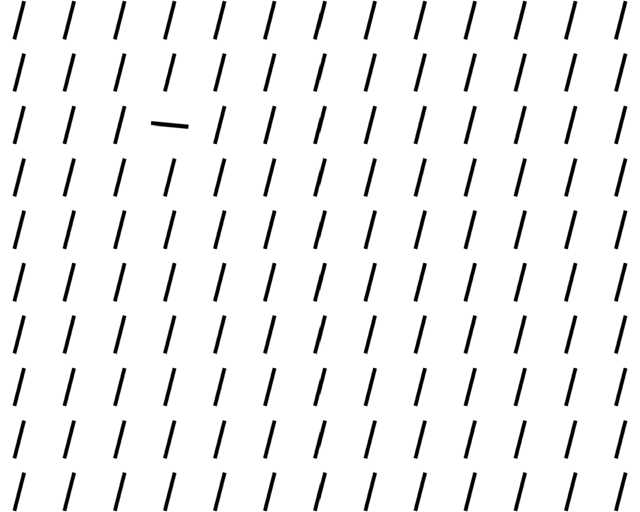}}
\fbox{\includegraphics[width=.12\linewidth]{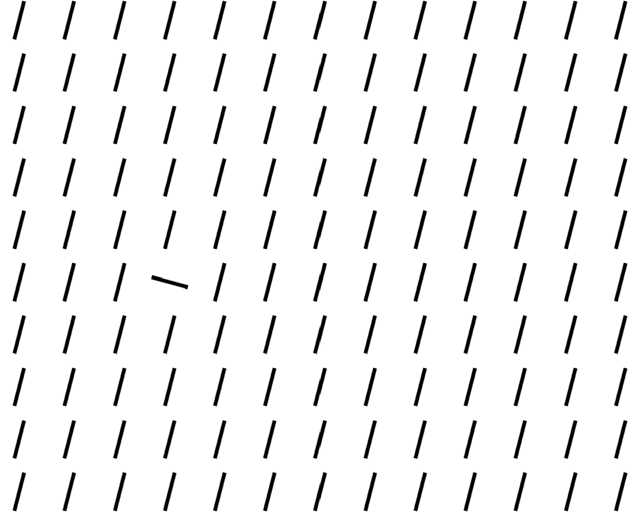}}
\\\hspace{1em}
\fbox{\includegraphics[width=.12\linewidth]{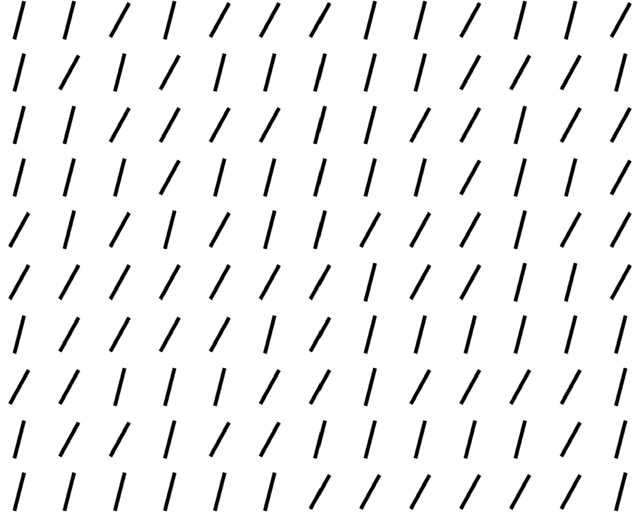}}
\fbox{\includegraphics[width=.12\linewidth]{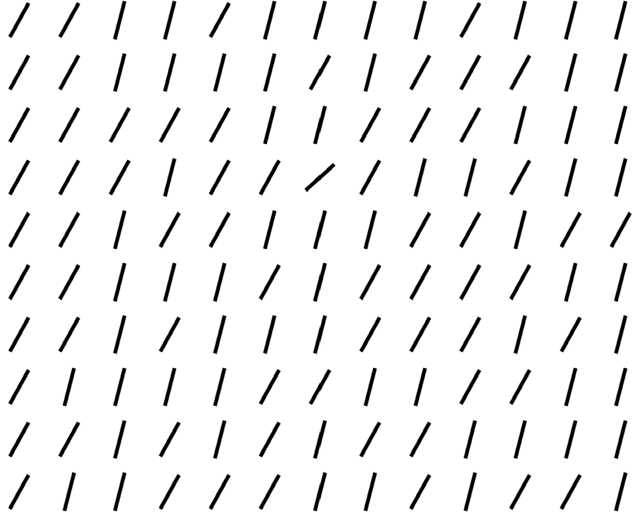}}
\fbox{\includegraphics[width=.12\linewidth]{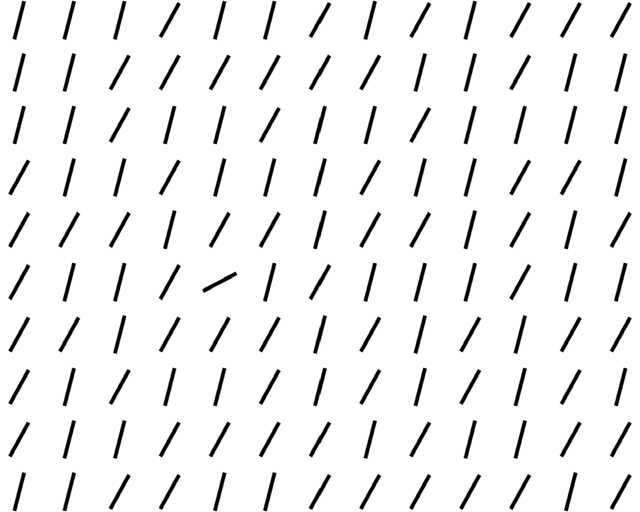}}
\fbox{\includegraphics[width=.12\linewidth]{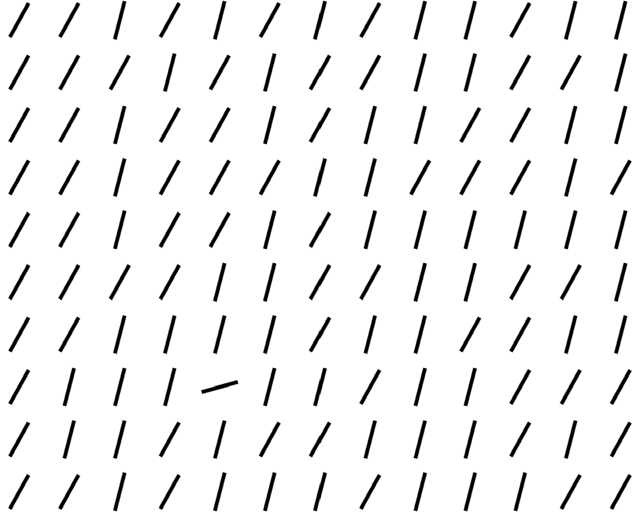}}
\fbox{\includegraphics[width=.12\linewidth]{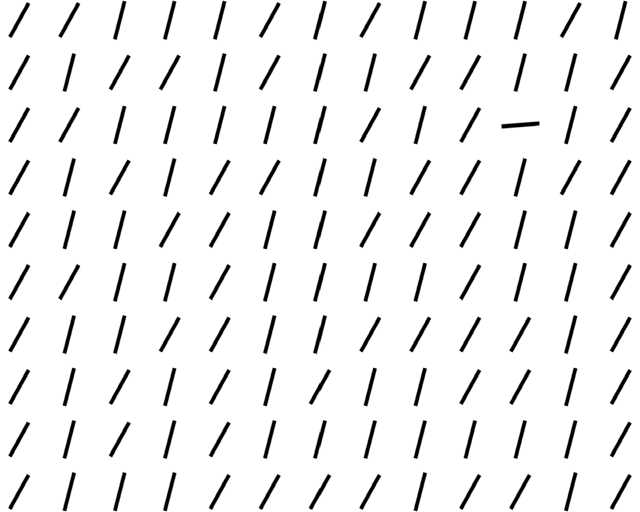}}
\fbox{\includegraphics[width=.12\linewidth]{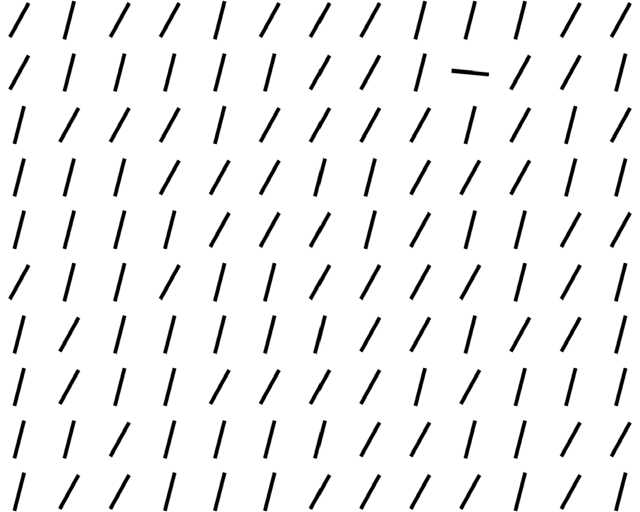}}
\fbox{\includegraphics[width=.12\linewidth]{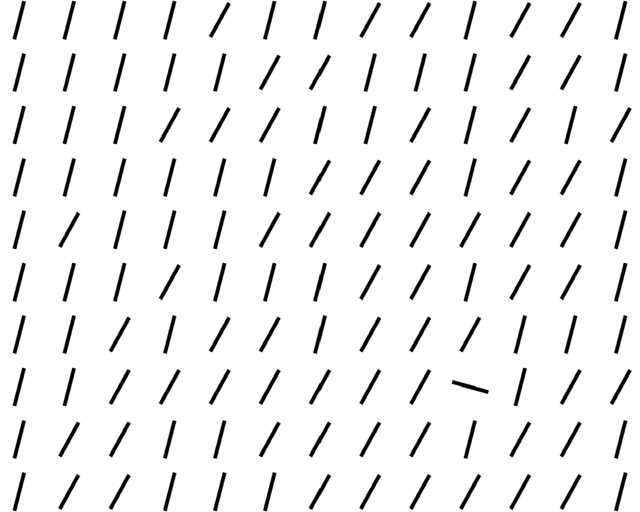}}
\\\hspace{1em}
\fbox{\includegraphics[width=.12\linewidth]{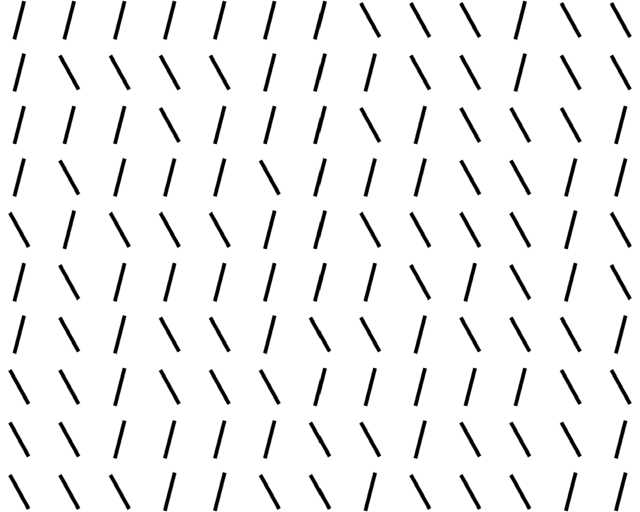}}
\fbox{\includegraphics[width=.12\linewidth]{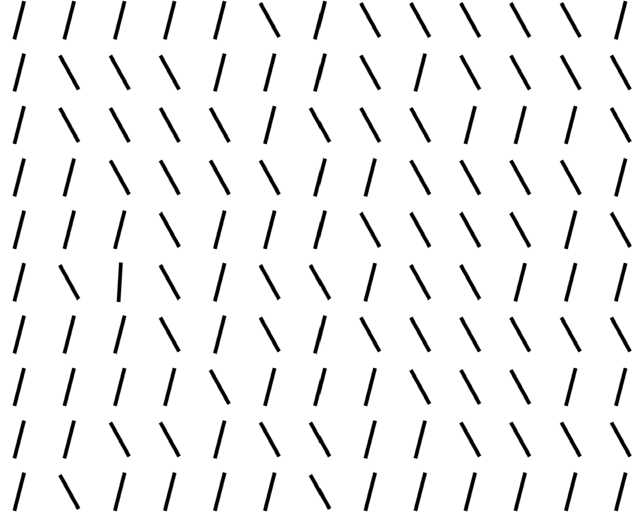}}
\fbox{\includegraphics[width=.12\linewidth]{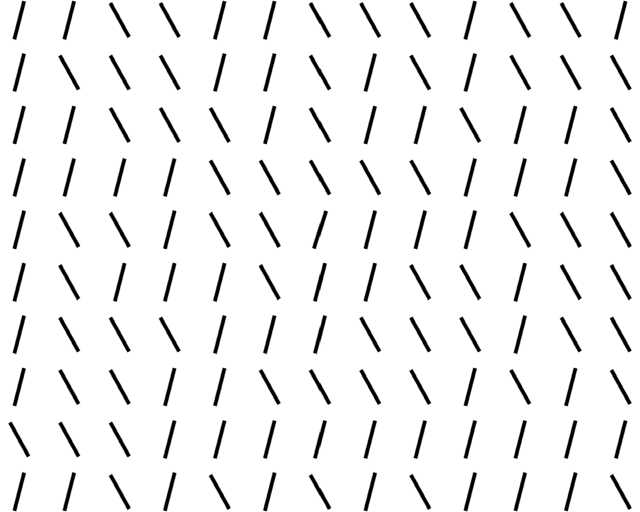}}
\fbox{\includegraphics[width=.12\linewidth]{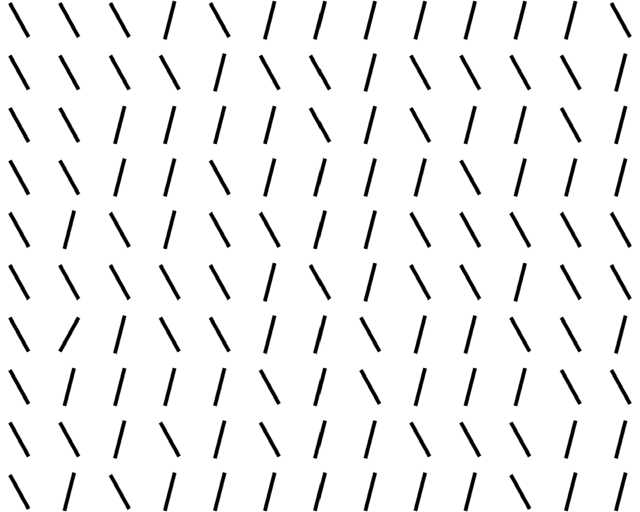}}
\fbox{\includegraphics[width=.12\linewidth]{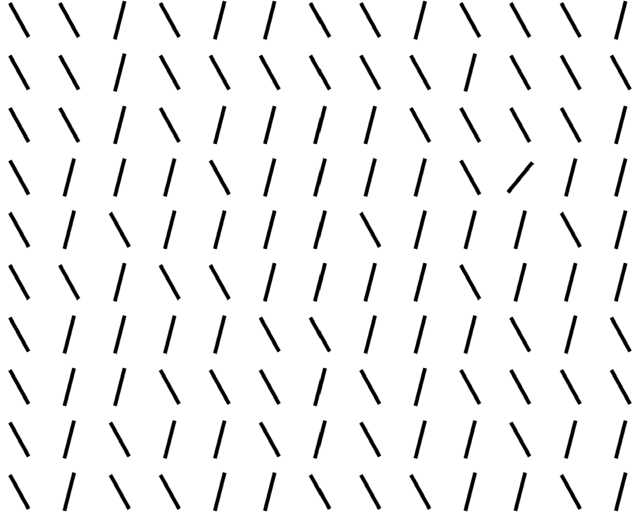}}
\fbox{\includegraphics[width=.12\linewidth]{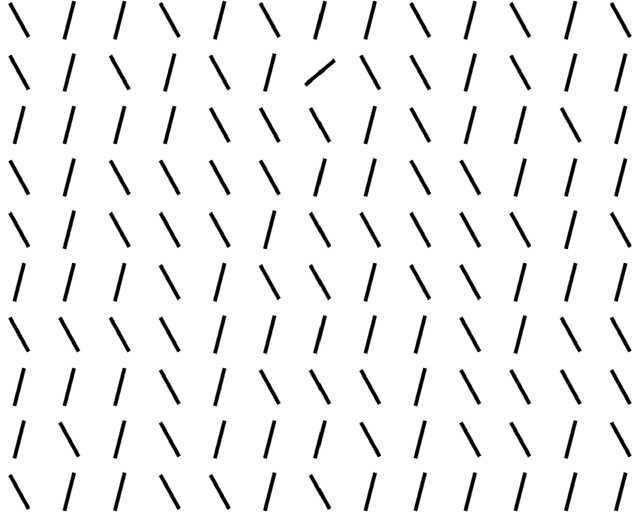}}
\fbox{\includegraphics[width=.12\linewidth]{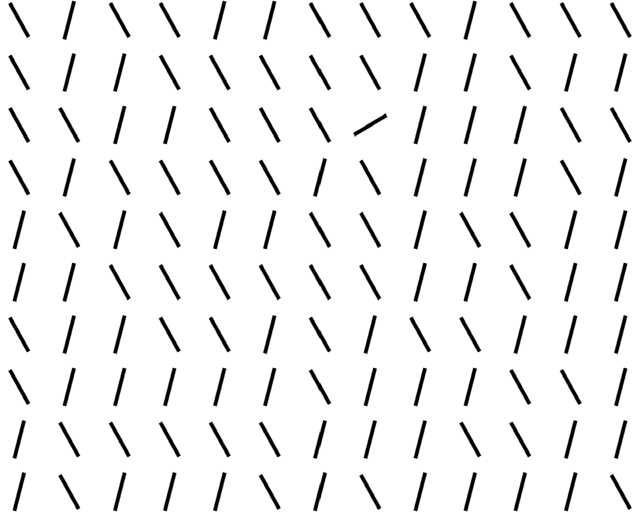}}
\\\makebox[1em]{14)}
\fbox{\includegraphics[width=.12\linewidth]{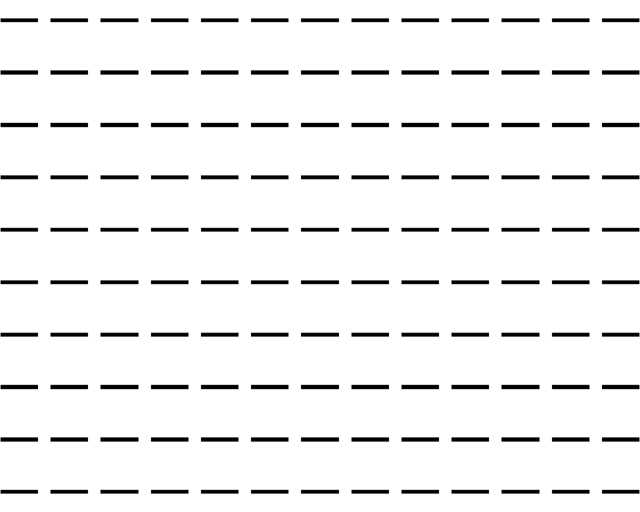}}
\fbox{\includegraphics[width=.12\linewidth]{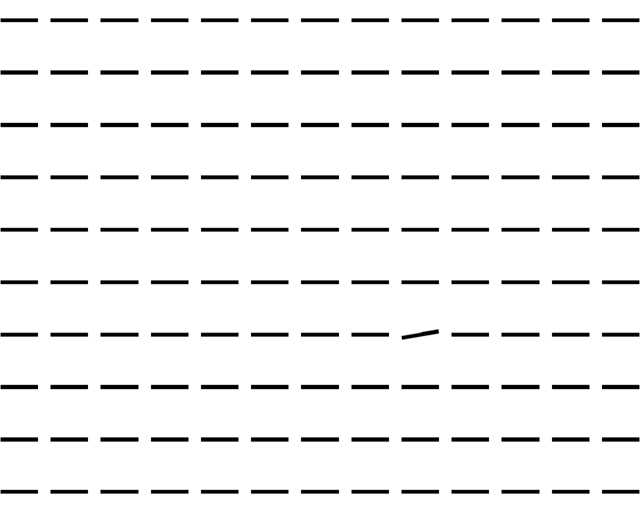}}
\fbox{\includegraphics[width=.12\linewidth]{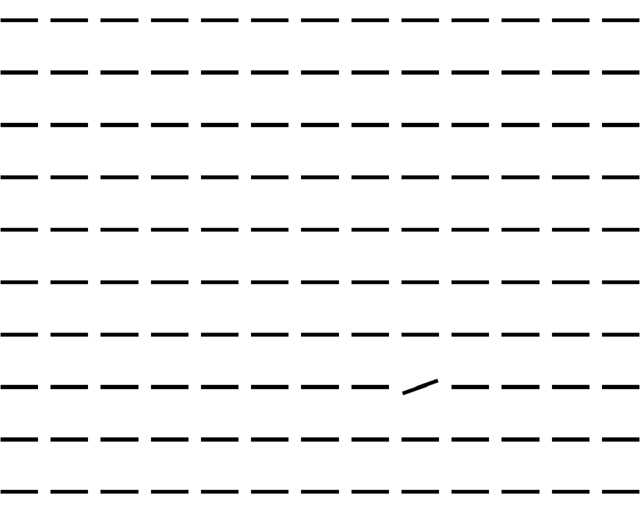}}
\fbox{\includegraphics[width=.12\linewidth]{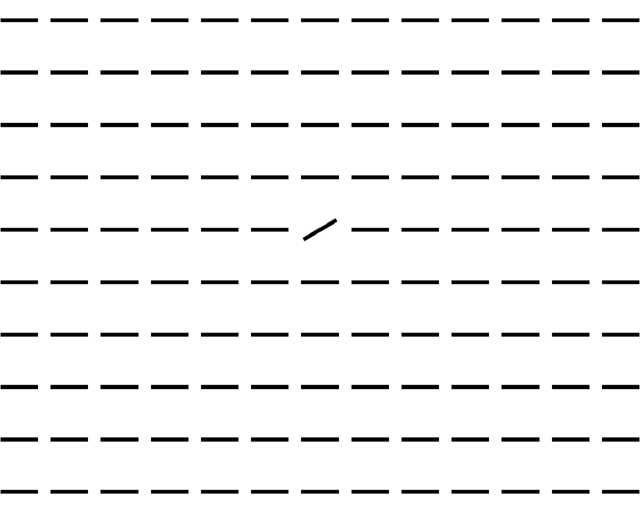}}
\fbox{\includegraphics[width=.12\linewidth]{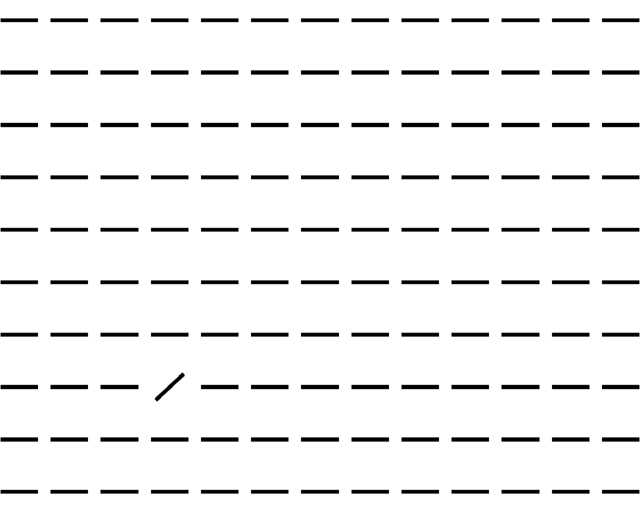}}
\fbox{\includegraphics[width=.12\linewidth]{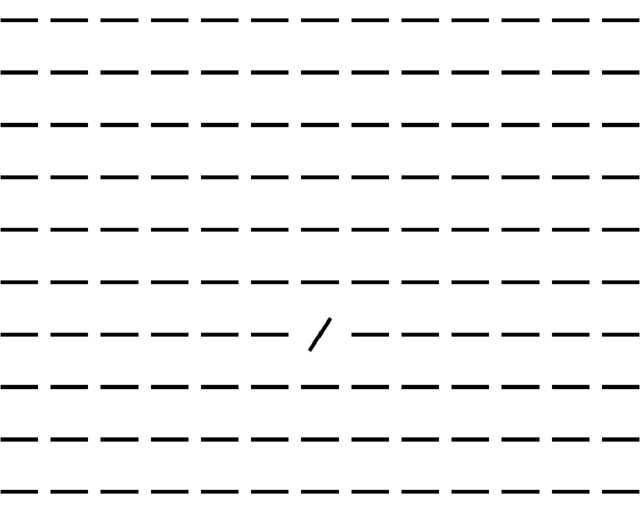}}
\fbox{\includegraphics[width=.12\linewidth]{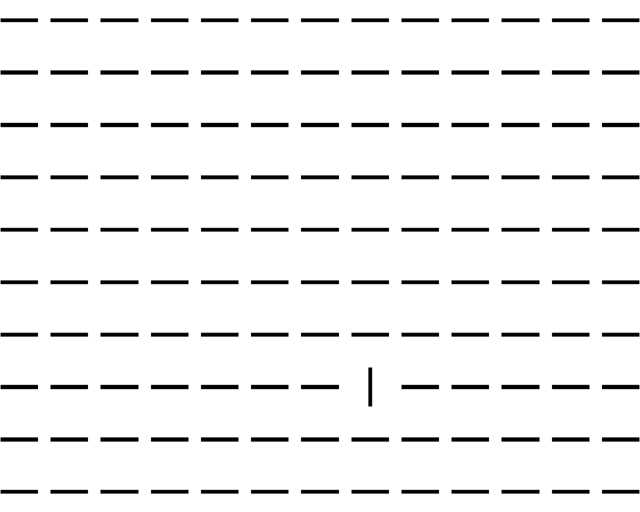}}
\\\hspace{1em}
\fbox{\includegraphics[width=.12\linewidth]{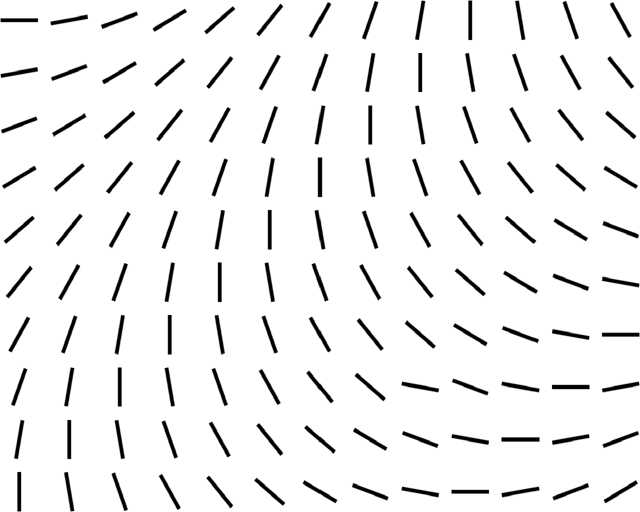}}
\fbox{\includegraphics[width=.12\linewidth]{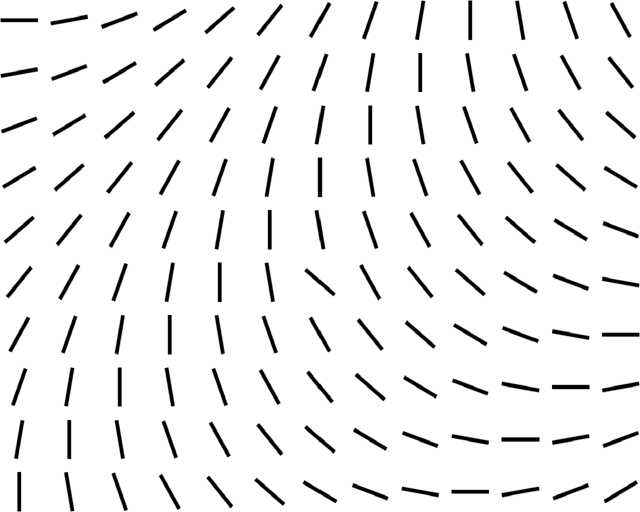}}
\fbox{\includegraphics[width=.12\linewidth]{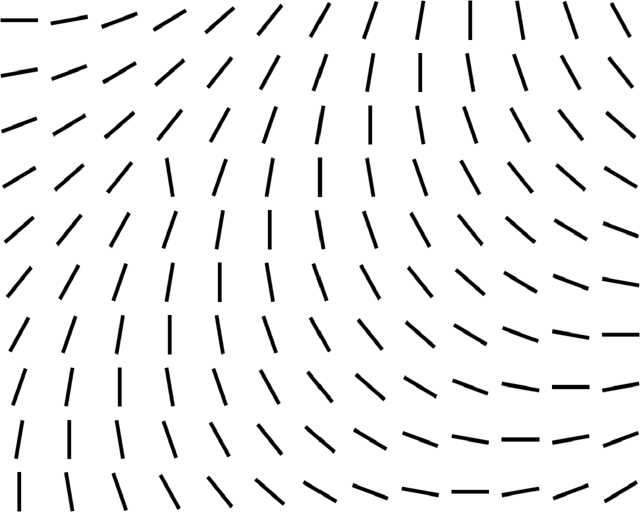}}
\fbox{\includegraphics[width=.12\linewidth]{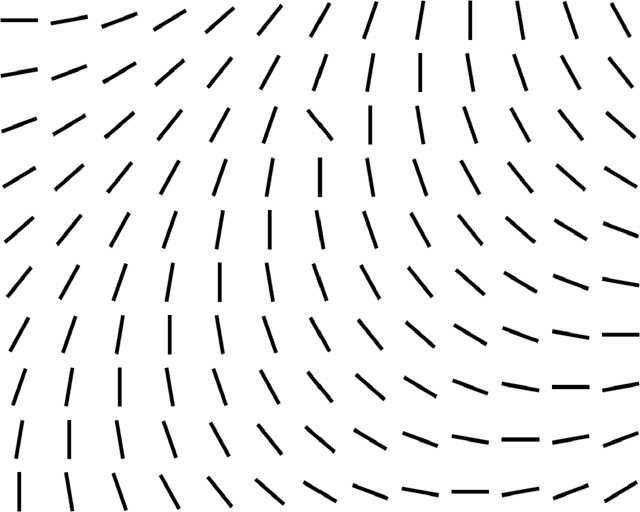}}
\fbox{\includegraphics[width=.12\linewidth]{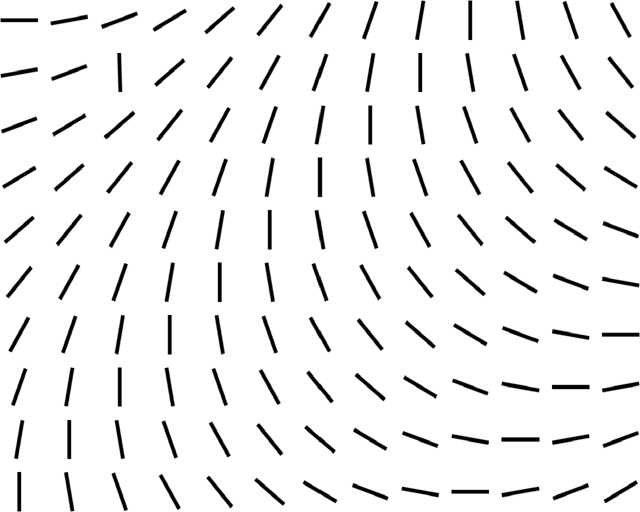}}
\fbox{\includegraphics[width=.12\linewidth]{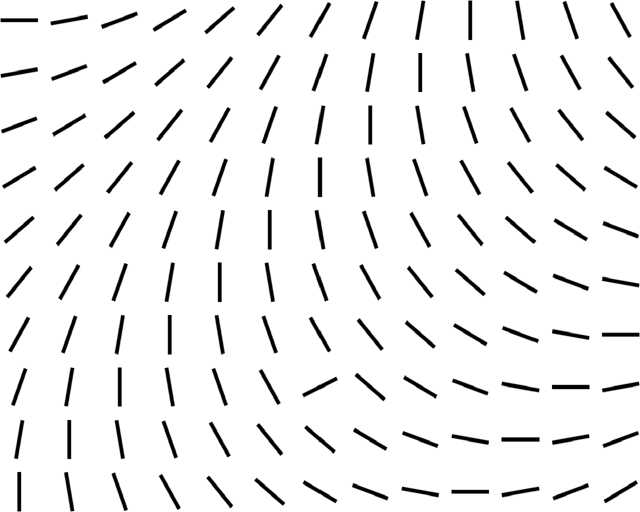}}
\fbox{\includegraphics[width=.12\linewidth]{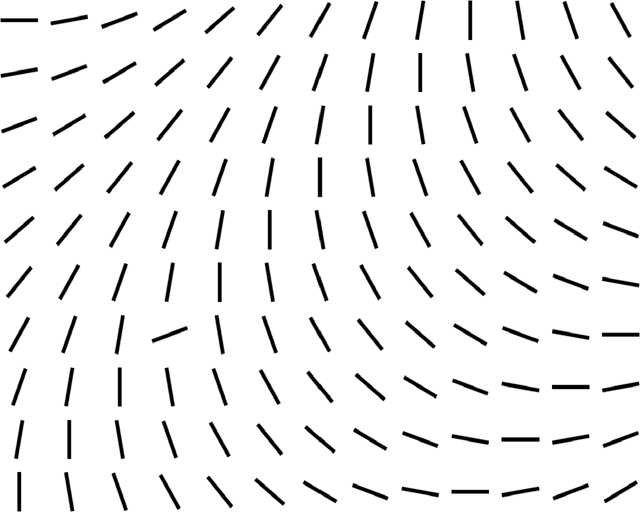}}
\\\hspace{1em}
\fbox{\includegraphics[width=.12\linewidth]{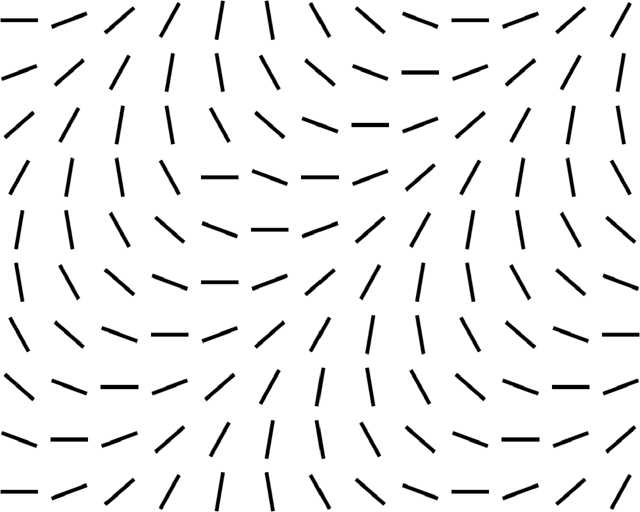}}
\fbox{\includegraphics[width=.12\linewidth]{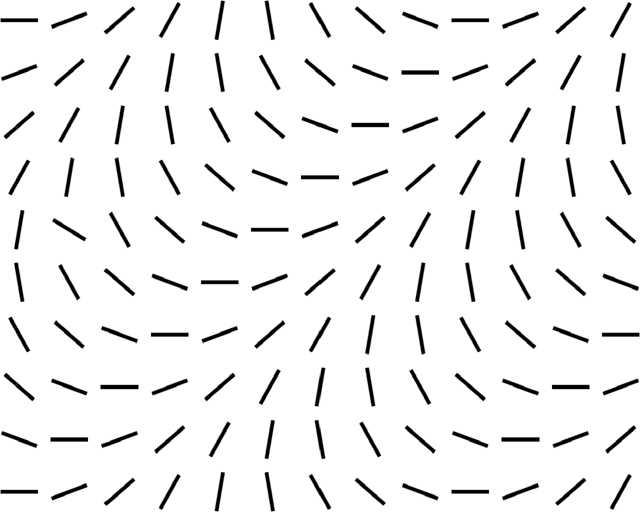}}
\fbox{\includegraphics[width=.12\linewidth]{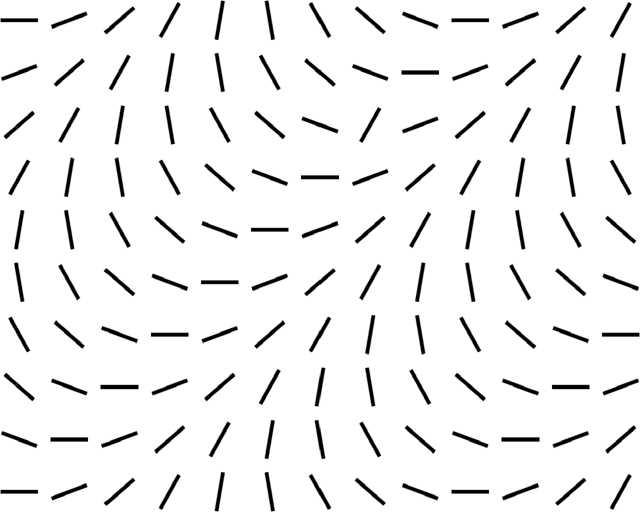}}
\fbox{\includegraphics[width=.12\linewidth]{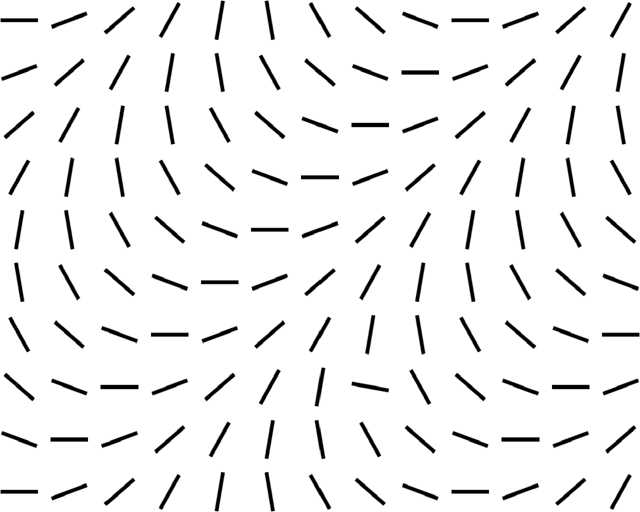}}
\fbox{\includegraphics[width=.12\linewidth]{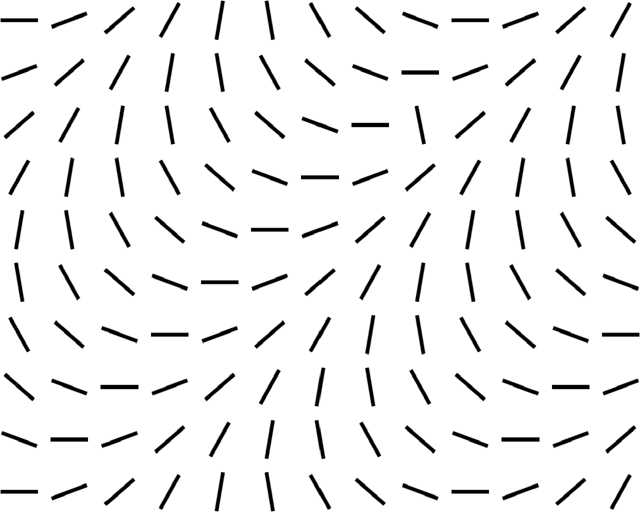}}
\fbox{\includegraphics[width=.12\linewidth]{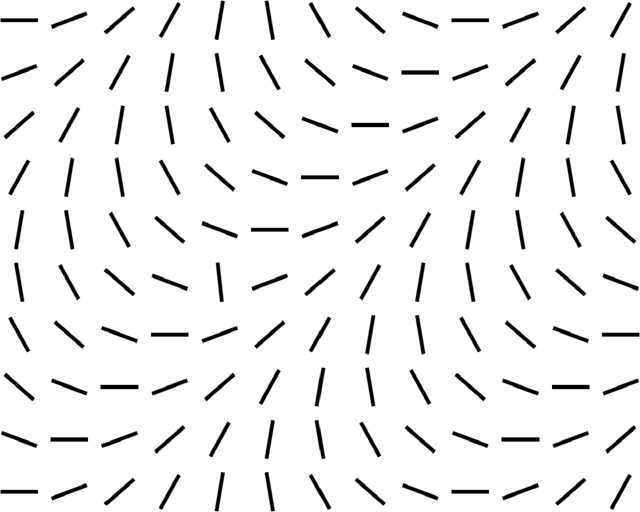}}
\fbox{\includegraphics[width=.12\linewidth]{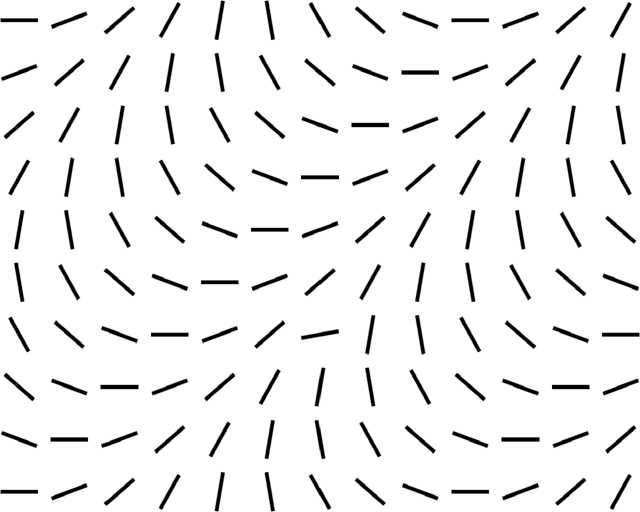}}
\\\hspace{1em}
\fbox{\includegraphics[width=.12\linewidth]{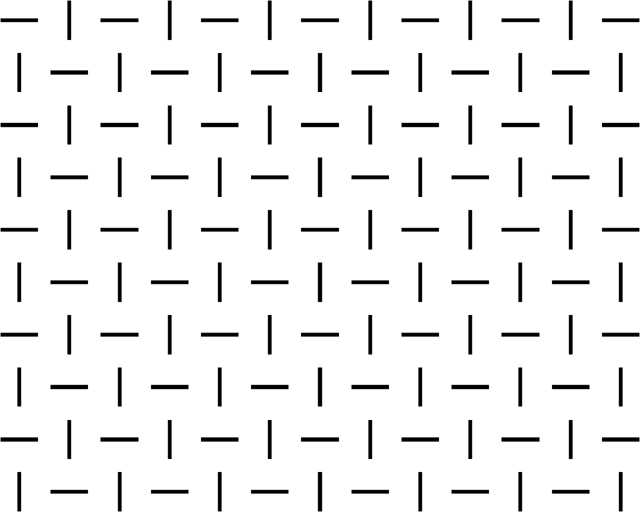}}
\fbox{\includegraphics[width=.12\linewidth]{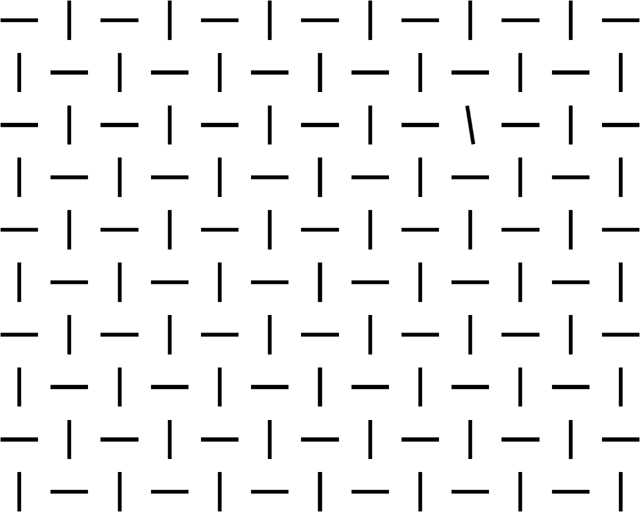}}
\fbox{\includegraphics[width=.12\linewidth]{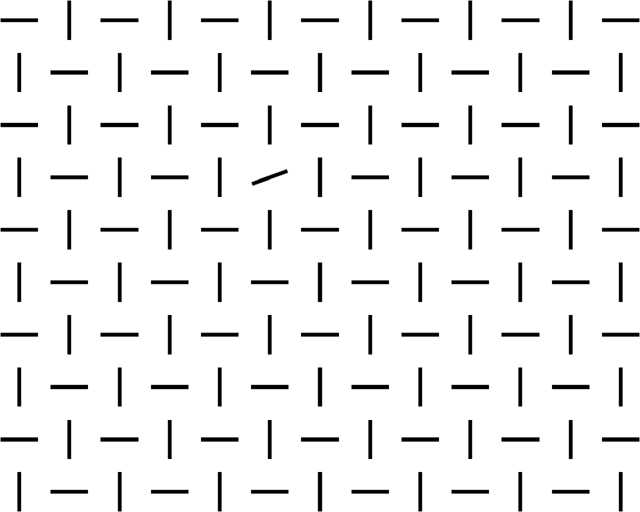}}
\fbox{\includegraphics[width=.12\linewidth]{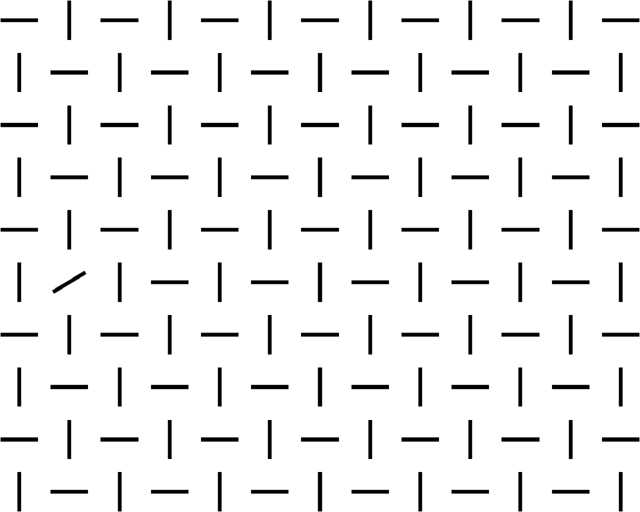}}
\fbox{\includegraphics[width=.12\linewidth]{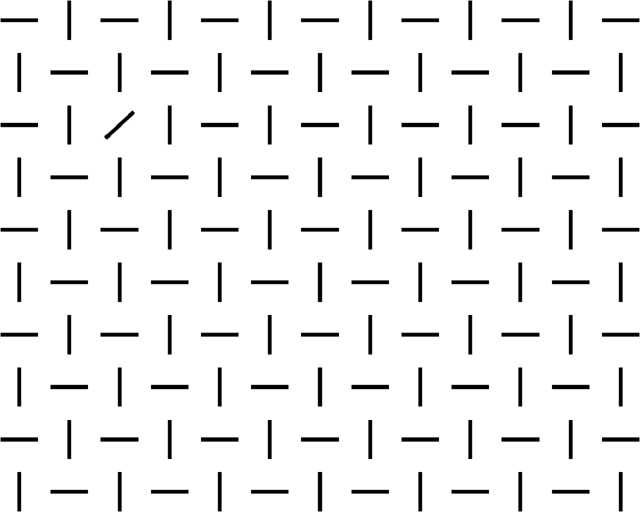}}
\fbox{\includegraphics[width=.12\linewidth]{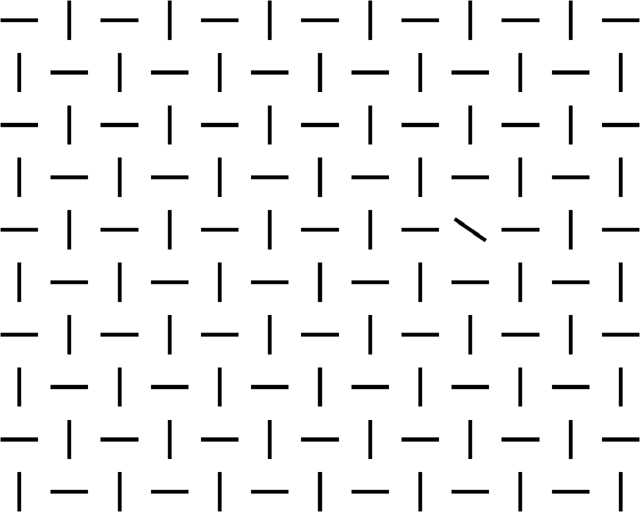}}
\fbox{\includegraphics[width=.12\linewidth]{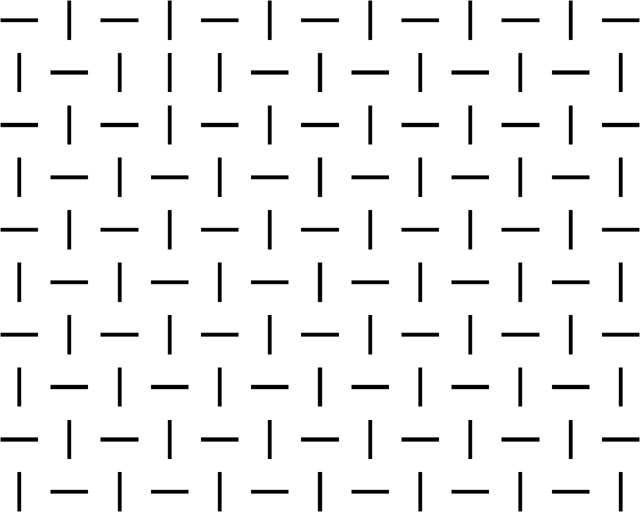}}
\\\makebox[1em]{15)}
\fbox{\includegraphics[width=.12\linewidth]{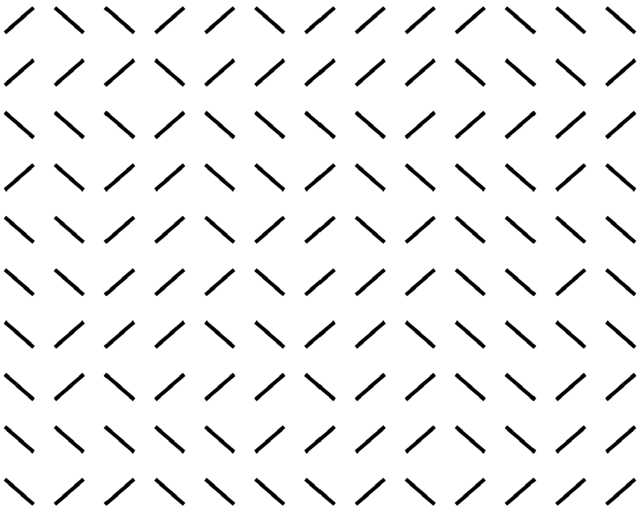}}
\fbox{\includegraphics[width=.12\linewidth]{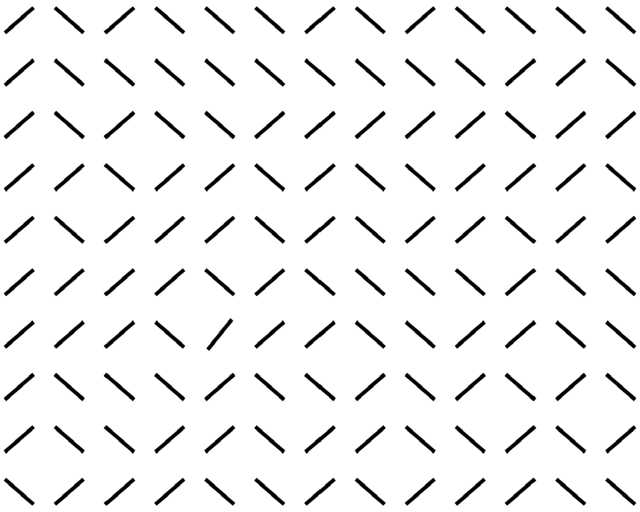}}
\fbox{\includegraphics[width=.12\linewidth]{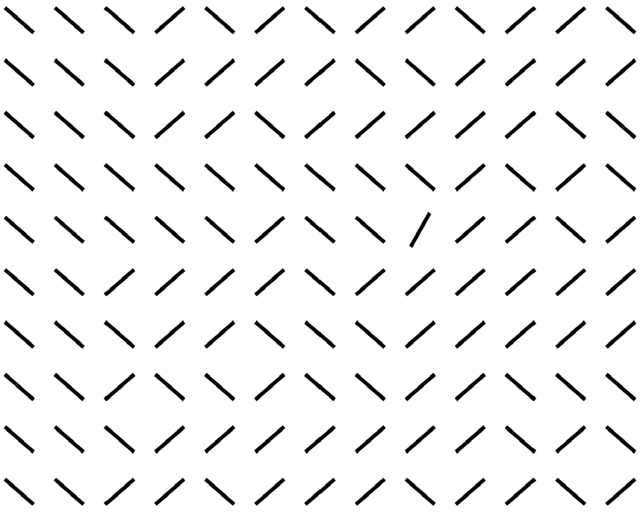}}
\fbox{\includegraphics[width=.12\linewidth]{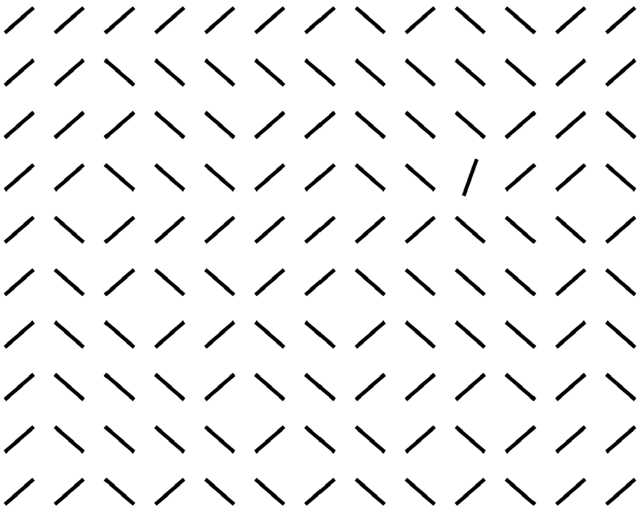}}
\fbox{\includegraphics[width=.12\linewidth]{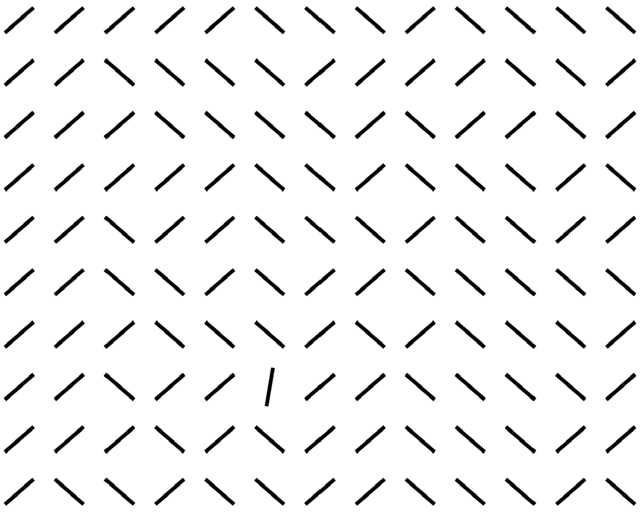}}
\fbox{\includegraphics[width=.12\linewidth]{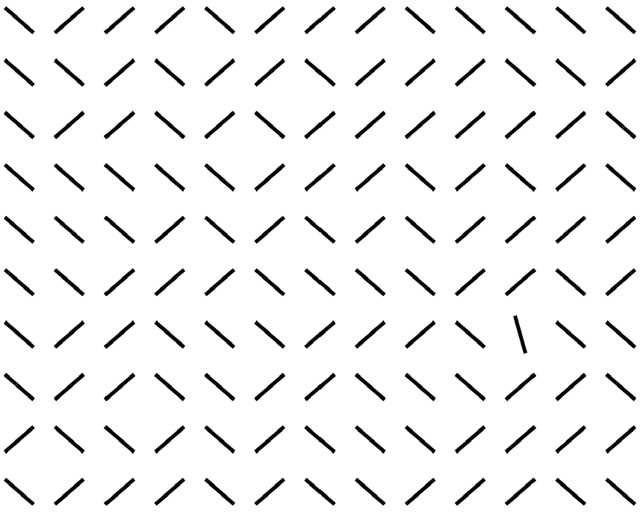}}
\fbox{\includegraphics[width=.12\linewidth]{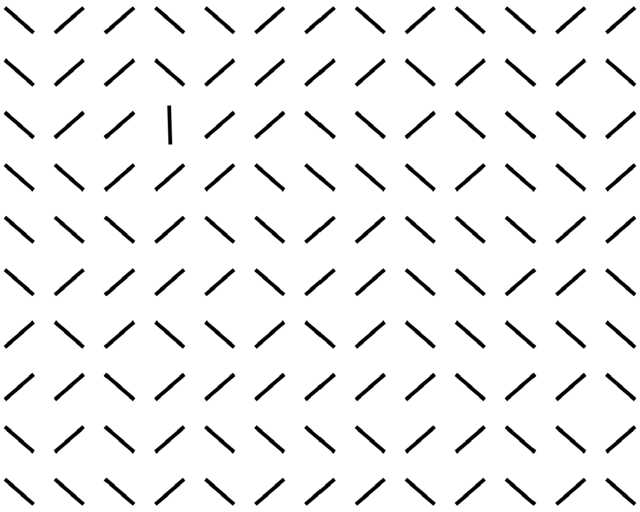}}
\\\hspace{1em}
\fbox{\includegraphics[width=.12\linewidth]{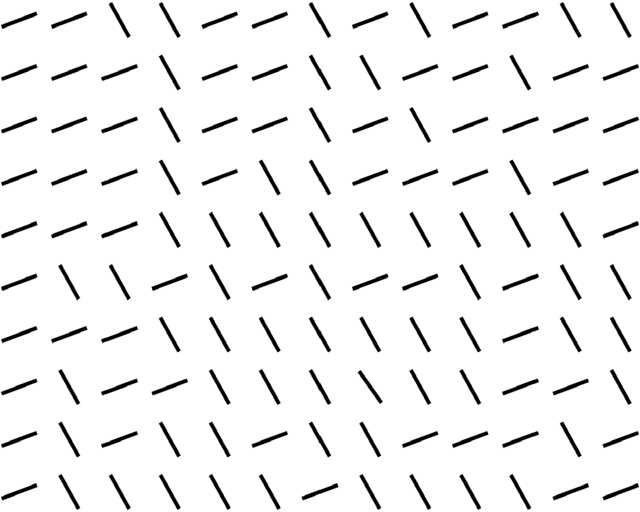}}
\fbox{\includegraphics[width=.12\linewidth]{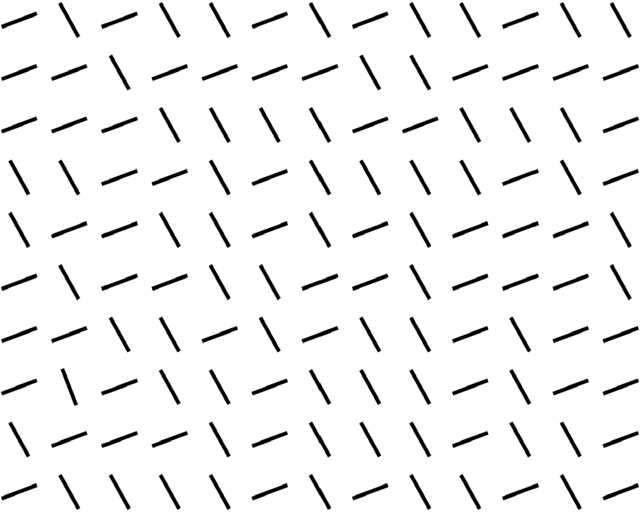}}
\fbox{\includegraphics[width=.12\linewidth]{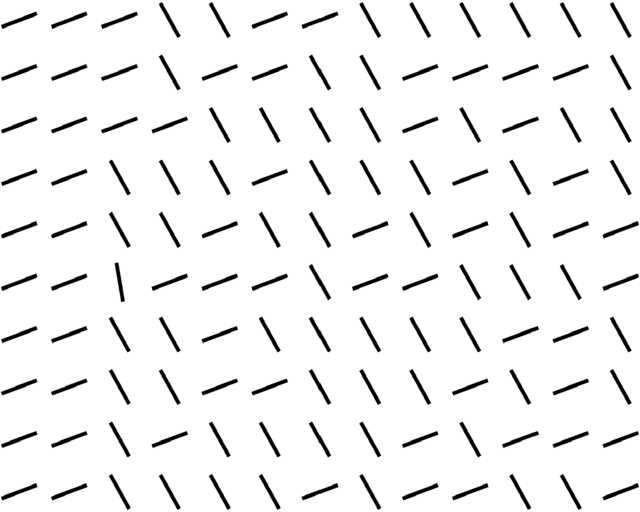}}
\fbox{\includegraphics[width=.12\linewidth]{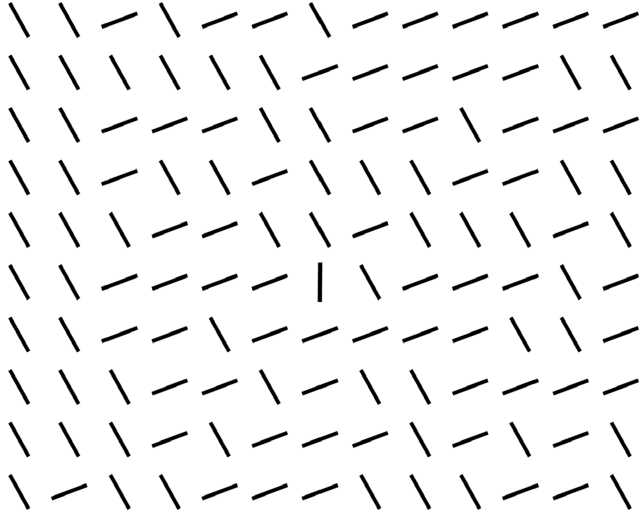}}
\fbox{\includegraphics[width=.12\linewidth]{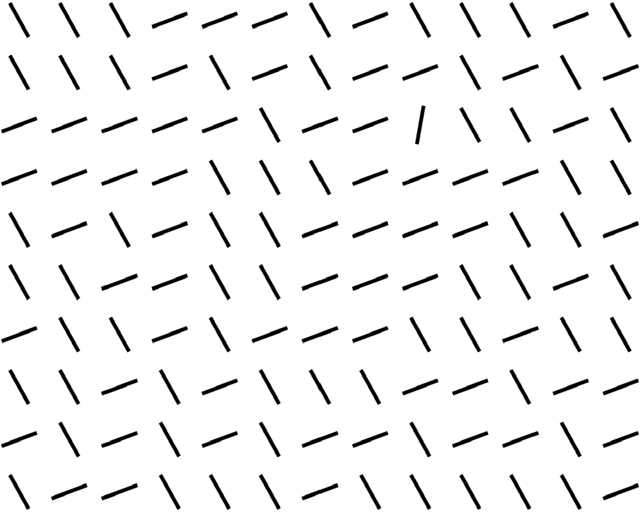}}
\fbox{\includegraphics[width=.12\linewidth]{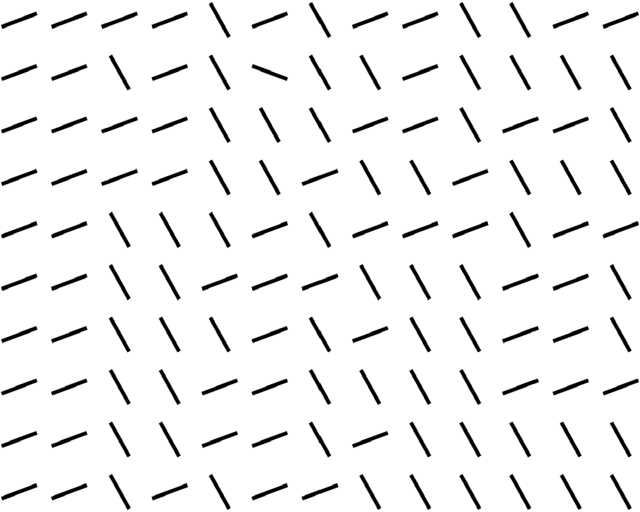}}
\fbox{\includegraphics[width=.12\linewidth]{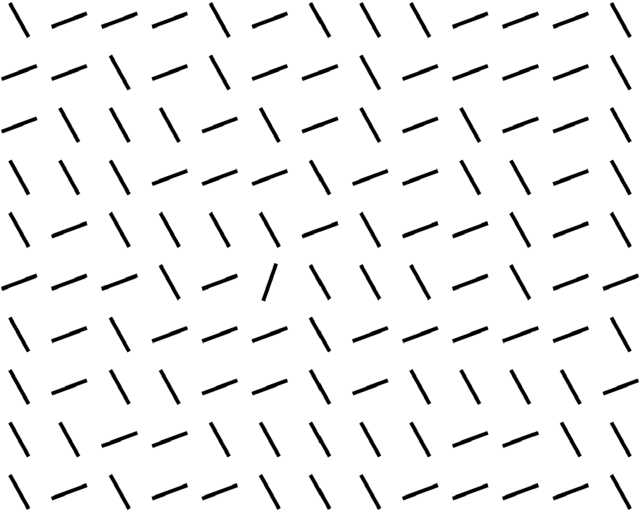}}
\\\hspace{1em}
\fbox{\includegraphics[width=.12\linewidth]{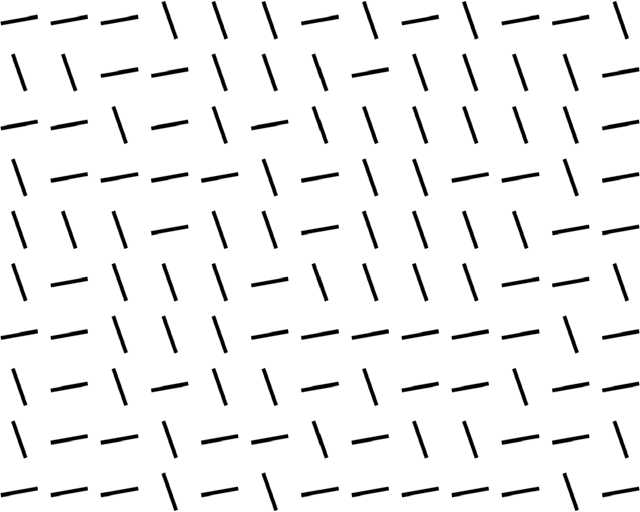}}
\fbox{\includegraphics[width=.12\linewidth]{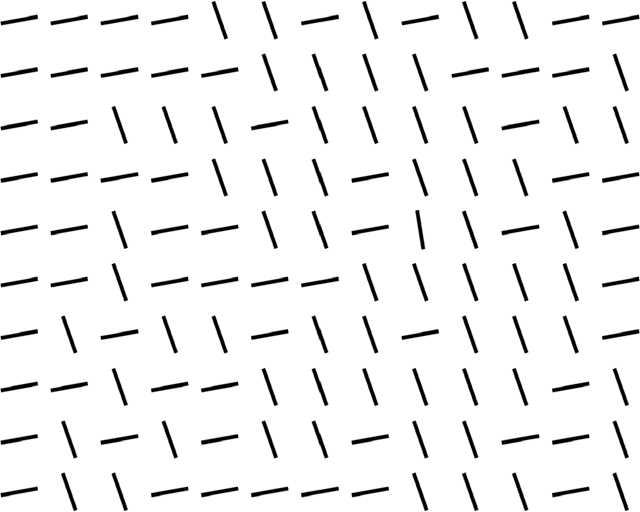}}
\fbox{\includegraphics[width=.12\linewidth]{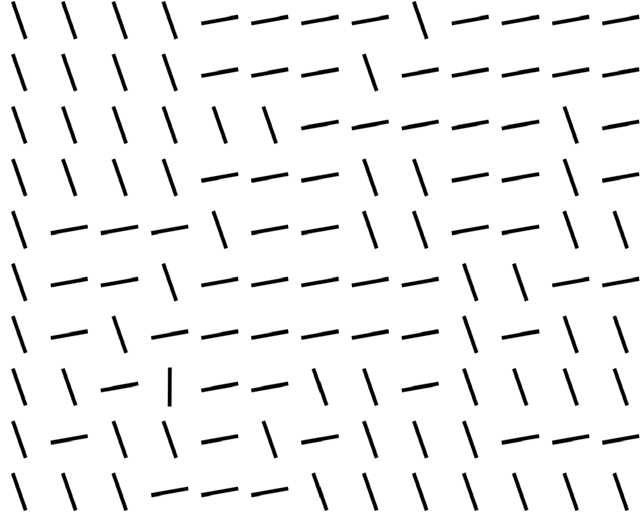}}
\fbox{\includegraphics[width=.12\linewidth]{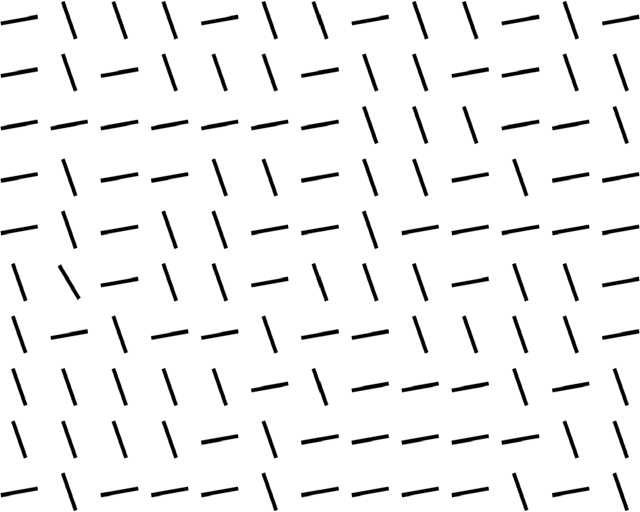}}
\fbox{\includegraphics[width=.12\linewidth]{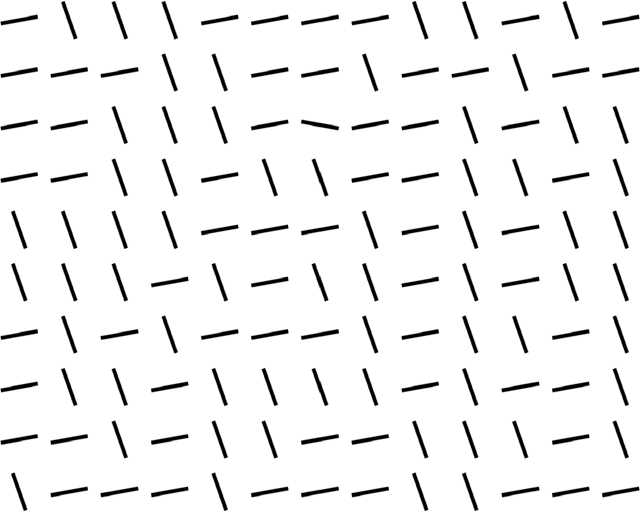}}
\fbox{\includegraphics[width=.12\linewidth]{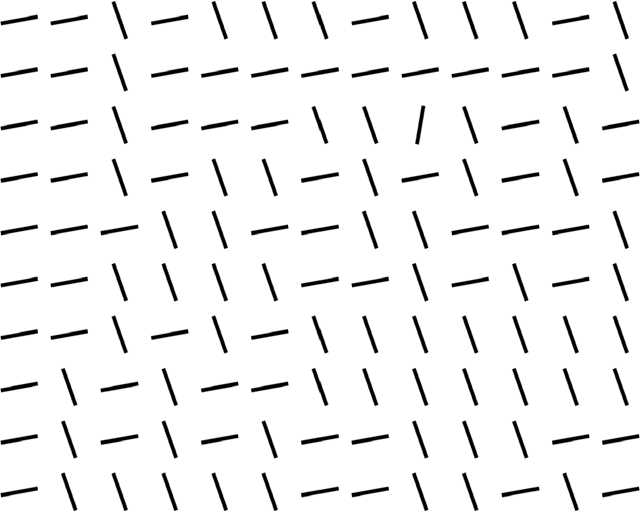}}
\fbox{\includegraphics[width=.12\linewidth]{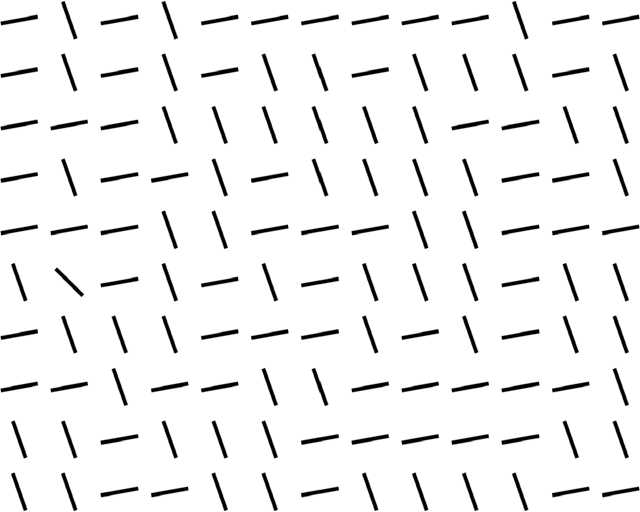}}
\makebox[0.05\linewidth]{ }\makebox[0.13\linewidth]{1}\makebox[0.13\linewidth]{2}\makebox[0.13\linewidth]{3}\makebox[0.13\linewidth]{4}\makebox[0.13\linewidth]{5}\makebox[0.13\linewidth]{6}\makebox[0.13\linewidth]{7}
\\
hard  $\longleftarrow$ $\Psi$ $\longrightarrow$ easy
\caption{Visual Search stimuli}
 \label{fig:vs}
\end{figure}

Fixations were collected from 34 participants in a dataset\footnote{Download the dataset in \url{http://www.cvc.uab.es/neurobit/?page_id=53}} of 230 images\cite{Berga2018a}. Images were displayed in a resolution of $1280\times1024$ px and fixations were captured at about $40$ pixels per degree of visual angle using SMI RED binocular eye tracker. The dataset had been splitted in two tasks: Free-Viewing (FV) and Visual Search (VS). For the FV task, participants had to freely look at the image during 5 seconds. On each stimuli there was a salient area of interest (AOI). For the VS task, participants had the instruction to visually locate the AOI, setting the salient region as the different object. For this task, the trigger for prompting the transition to next image was by gazing inside the AOI or pressing a key (for reporting absence of target). We can observe the stimuli generated for both tasks on \hyperref[fig:fv]{Figs. \ref*{fig:fv}-\ref*{fig:vs}}.

The dataset was divided in 15 stimulus types, 5 corresponding to FV and 10 to VS. Some of these blocks had distinct subsets of images (due to the alteration of either target or distractor shape, color, configuration and background properties), abling a total of 33 subtypes. Each of these blocks was individually generated as a low-level feature category, which had its own type of feature contrast between the salient region and the rest of distractors / background. FV categories were mainly based for analyzing preattentive effects (\hyperref[fig:fv]{Fig. \ref*{fig:fv}}): 1) Corner Salience, 2) Visual Segmentation by Bar Angle, 3) Visual Segmentation by Bar Length, 4) Contour Integration by Bar Continuity and 5) Perceptual Grouping by Distance. VS categories were based on a feature-singleton search stimuli, where there was a unique salient target and a set of distractors and/or altered background (\hyperref[fig:vs]{Fig. \ref*{fig:vs}}). These categories were: 6) Feature and Conjunctive Search, 7) Search Asymmetries, 8) Search in a Rough Surface, 9) Color Search, 10) Brightness Search, 11) Orientation Search, 12) Dissimilar Size Search, 13) Orientation Search with Heterogeneous distractors, 14) Orientation Search with Non-linear patterns, 15) Orientation search with distinct Categorization. Stimuli for SID4VAM's dataset was inspired by previous psychophysical experimentation \cite{Zhaoping2007,Wolfe2010a,Spratling2012}. 

Dataset stimuli were generated with 7 specific instances of feature contrast ($\Psi$), corresponding to hard ($\Psi_h=\{1..4\}$) and easy ($\Psi_e=\{5..7\}$) difficulies of finding the salient regions. These feature contrasts had their own parametrization (following Berga et al's psychophysical formulation \cite[Section~2.4]{Berga2018a}) corresponding to the feature differences between the salient target and the rest of distractors (e.g. differences of target orientation, size, saturation, brightness...) or global effects (e.g. overall distractor scale, shape, background color, background brightness).
\hfill\newline

\section{Experiments}

Fixation maps from eye tracking data are generated by distributing each fixation location to a binary map. Fixation density maps are created by convolving a gaussian filter to the fixation maps, this simulates a smoothing caused by the deviations of $\sigma$=$1$ deg given from eye tracking experimentation, recommended by LeMeur \& Baccino \cite{LeMeur2012}. 

Typically, location-based saliency metrics ($AUC_{Judd}$, $AUC_{Borji}$, NSS) increase their score fixation locations fall inside (TP) the predicted saliency maps. Conversely, scores decrease fixation locations are not captured by saliency maps (FN) or when saliency maps exist in locations with no present fixations (FP). In distribution-based metrics (CC, SIM, KL), saliency maps score higher when they have higher correlations with respect to fixation density map distributions. We have to point out that shuffled metrics (sAUC, InfoGain) consider FP values when saliency maps coincide with other fixation map locations or a baseline (here, corresponding to the center bias), which are not representative data for saliency prediction. Prediction metrics and its calculations are largely explained by Bylinskii et al. \cite{Bylinskii2018}. Our saliency metric scores and pre-processing used for this experimentation\footnote{Code for metrics: \url{https://github.com/dberga/saliency}} have been replicated from the official saliency benchmarking procedure \cite{mit-saliency-benchmark}. Psychometric evaluation of saliency predictions has been done with the Saliency Index (SI) \cite{Soltani2010,Spratling2012}. This metric evaluates the energy of a saliency map inside ($S_t$) a salient region (which would enclose a salient object) compared to the energy outside ($S_b$) the salient region. This metric allows evaluation of a saliency map when the salient region is known, considering in absolute terms the distribution of saliency of a particular AOI / mask. Here we show the formula of the SI 

%

\begin{equation*}
SI(S_t,S_b)=\frac{S_t-S_b}{S_b}.
\label{eq:sindex}
\end{equation*}

Saliency maps have been computed from models shown on \hyperref[table:models]{Table \ref*{table:models}}. Model evaluations have been divided according to its inspiration and prediction scores have been evaluated with saliency metrics and in psychophysical terms. 

\subsection{Model results on predicting fixations}
\begin{figure*}[h!]
\centering
\includegraphics[width=.40\linewidth]{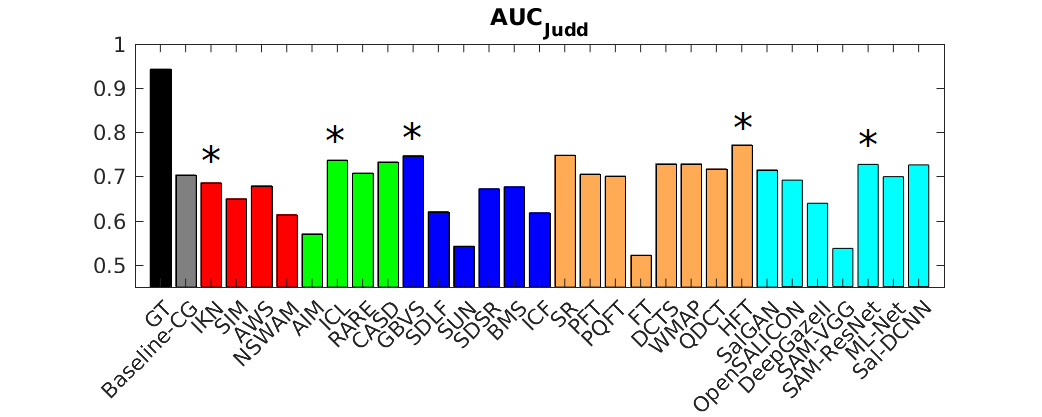}
\includegraphics[width=.40\linewidth]{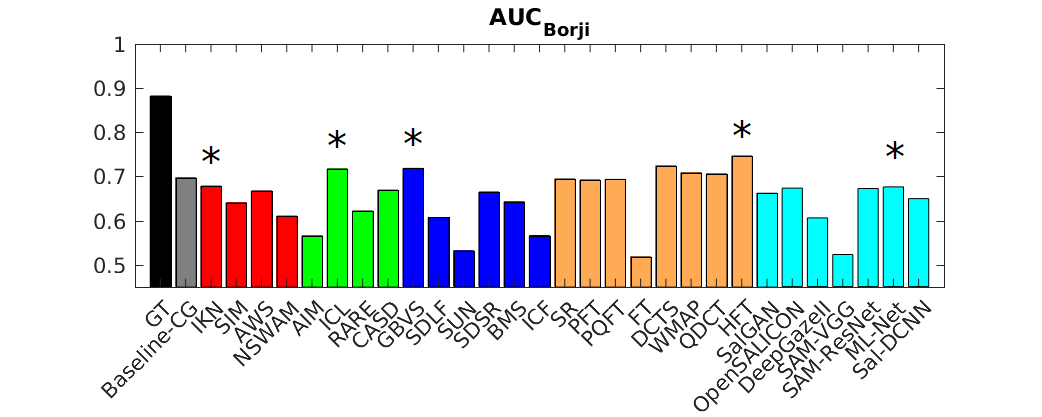}
\includegraphics[width=.40\linewidth]{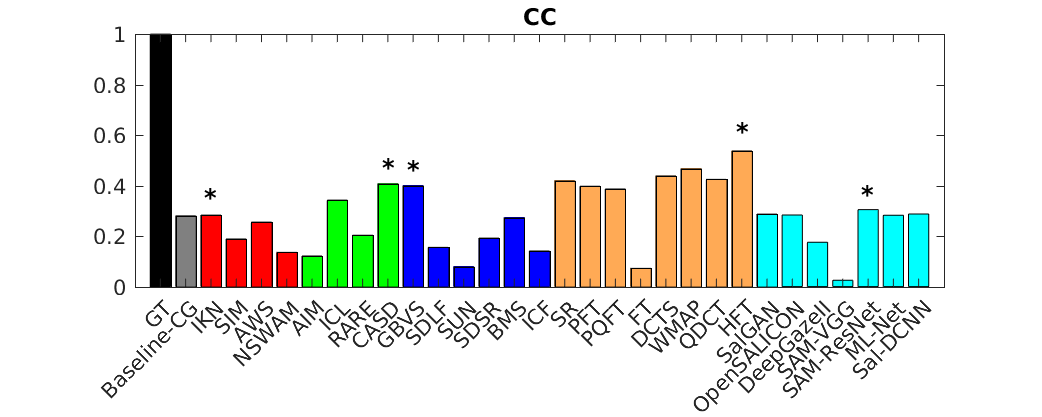}
\includegraphics[width=.40\linewidth]{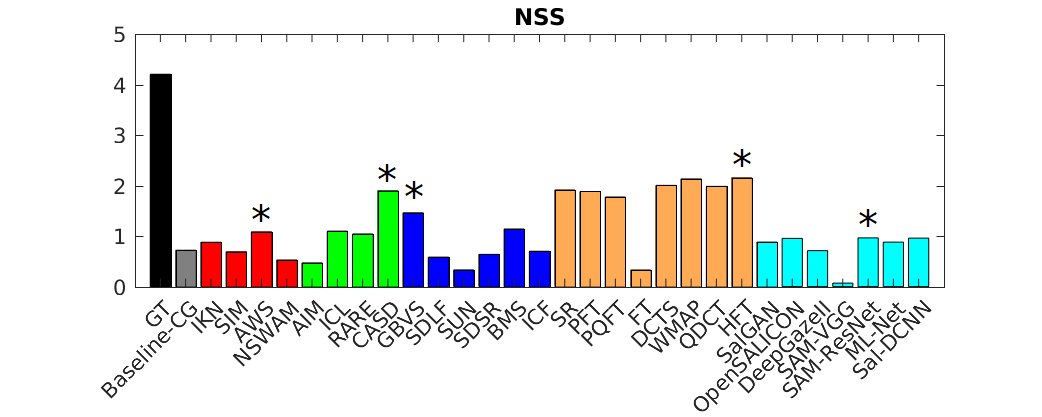}
\includegraphics[width=.40\linewidth]{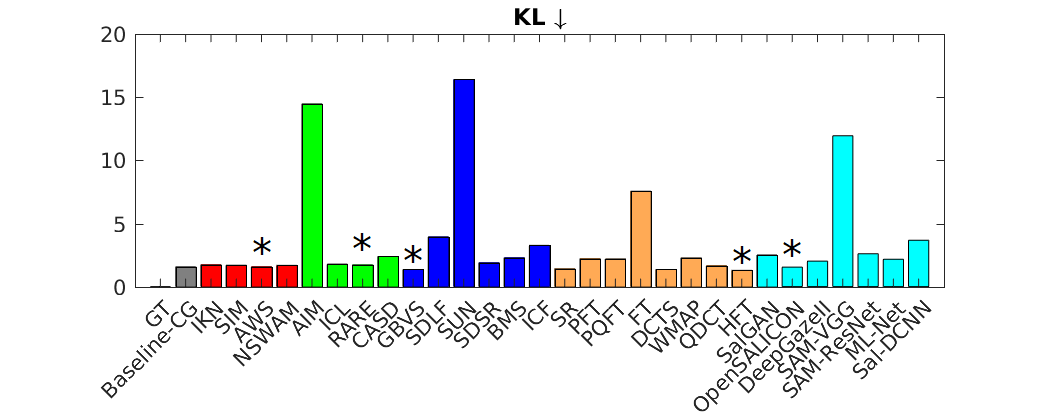}
\includegraphics[width=.40\linewidth]{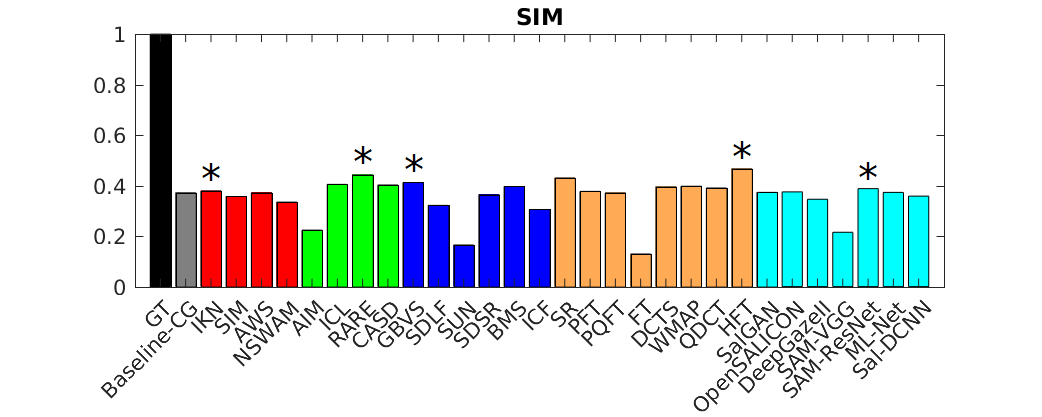}
\includegraphics[width=.40\linewidth]{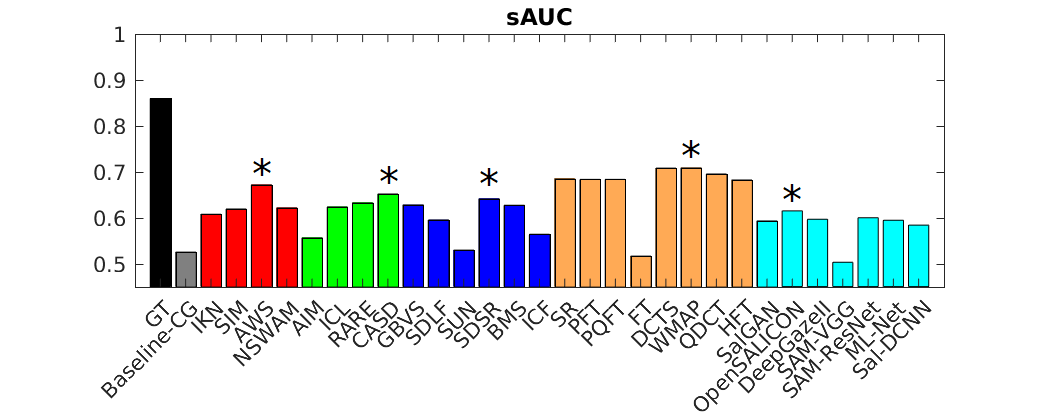}
\includegraphics[width=.40\linewidth]{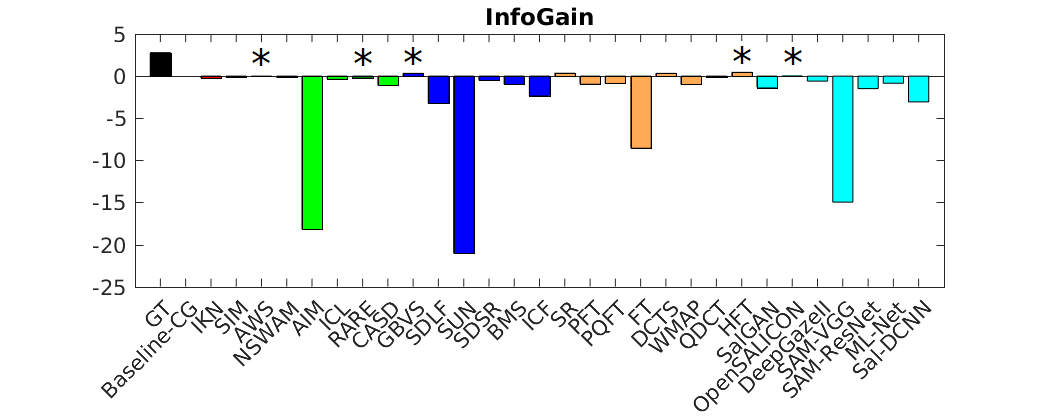}\\
\textbf{*}=Best performance for each model inspiration
\caption{Plots for saliency metric scores.}
\label{fig:results}
\end{figure*}

Previous saliency benchmarks \cite{Borji2013b,Riche2013a,Bylinskii2015,Bruce2015,mit-saliency-benchmark} reveal that Deep Learning models such as SALICON, ML-Net SAM-ResNet, SAM-VGG, DeepGazeII or SalGan score highest on both shuffled and unshuffled metrics. In this section we aim to evaluate whether saliency maps that scored highly on fixation prediction do so with a synthetic image dataset and if their inspiration influences on their performance. We present metric scores of saliency map predictions of the whole dataset in \hyperref[tab:results]{Table \ref*{tab:results}} and plots in \hyperref[fig:results]{Fig. \ref*{fig:results}}. Saliency metric scores reveal that overall Spectral/Fourier-based saliency models predict better fixations on a synthetic image dataset.


\pgfplotstableread[col sep=comma]{metrics.csv}\mytable

\begin{table}[h!]
\caption{Saliency metric scores for SID4VAM} 
\label{tab:results}
\begin{adjustbox}{width=.50\textwidth}
\pgfplotstabletypeset[
string type,
columns/model/.style={column name=\textbf{Model}},
columns/aucj/.style={column name=AUCj},
columns/aucb/.style={column name=AUCb},
columns/cc/.style={column name=CC},
columns/nss/.style={column name=NSS},
columns/kl/.style={column name=KL},
columns/sim/.style={column name=SIM},
columns/sauc/.style={column name=sAUC},
columns/infogain/.style={column name=InfoGain},
every first column/.style={},assign column name/.style={/pgfplots/table/column name={\textbf{#1}}},
every head row/.style={before row=\toprule, after row=\midrule},
every last row/.style={after row={\toprule}}, 
every row 0 column 0/.style={postproc cell content/.style={@cell content=\cellcolor{gray!20}\textit{##1}}}, 
every row 1 column 0/.style={postproc cell content/.style={@cell content=\cellcolor{gray!20}\textit{##1}}}, 
every row 1 column 1/.style={postproc cell content/.style={@cell content=\cellcolor{white}##1}},
every row 0 column 1/.style={postproc cell content/.style={@cell content=\cellcolor{white}\textbf{\textit{##1}}}},
every row 0 column 2/.style={postproc cell content/.style={@cell content=\cellcolor{white}\textbf{\textit{##1}}}},
every row 0 column 3/.style={postproc cell content/.style={@cell content=\cellcolor{white}\textbf{\textit{##1}}}},
every row 0 column 4/.style={postproc cell content/.style={@cell content=\cellcolor{white}\textbf{\textit{##1}}}},
every row 0 column 5/.style={postproc cell content/.style={@cell content=\cellcolor{white}\textbf{\textit{##1}}}},
every row 0 column 6/.style={postproc cell content/.style={@cell content=\cellcolor{white}\textbf{\textit{##1}}}},
every row 0 column 7/.style={postproc cell content/.style={@cell content=\cellcolor{white}\textbf{\textit{##1}}}},
every row 0 column 8/.style={postproc cell content/.style={@cell content=\cellcolor{white}\textbf{\textit{##1}}}},
every row 1 column 1/.style={postproc cell content/.style={@cell content=\cellcolor{white}\textbf{\textit{##1}}}},
every row 1 column 2/.style={postproc cell content/.style={@cell content=\cellcolor{white}\textbf{\textit{##1}}}},
every row 1 column 3/.style={postproc cell content/.style={@cell content=\cellcolor{white}\textbf{\textit{##1}}}},
every row 1 column 4/.style={postproc cell content/.style={@cell content=\cellcolor{white}\textbf{\textit{##1}}}},
every row 1 column 5/.style={postproc cell content/.style={@cell content=\cellcolor{white}\textbf{\textit{##1}}}},
every row 1 column 6/.style={postproc cell content/.style={@cell content=\cellcolor{white}\textbf{\textit{##1}}}},
every row 1 column 7/.style={postproc cell content/.style={@cell content=\cellcolor{white}\textbf{\textit{##1}}}},
every row 1 column 8/.style={postproc cell content/.style={@cell content=\cellcolor{white}\textbf{\textit{##1}}}},
every row 2 column 0/.style={postproc cell content/.style={@cell content=\cellcolor{red!20}##1}}, 
every row 3 column 0/.style={postproc cell content/.style={@cell content=\cellcolor{red!20}##1}}, 
every row 3 column 1/.style={postproc cell content/.style={@cell content=\cellcolor{white}##1}},
every row 4 column 0/.style={postproc cell content/.style={@cell content=\cellcolor{red!20}##1}}, 
every row 4 column 1/.style={postproc cell content/.style={@cell content=\cellcolor{white}##1}},
every row 5 column 0/.style={postproc cell content/.style={@cell content=\cellcolor{red!20}##1}}, 
every row 5 column 1/.style={postproc cell content/.style={@cell content=\cellcolor{white}##1}},
every row 2 column 1/.style={postproc cell content/.style={@cell content=\cellcolor{white}\underline{##1}}},
every row 2 column 2/.style={postproc cell content/.style={@cell content=\cellcolor{white}\underline{##1}}},
every row 2 column 3/.style={postproc cell content/.style={@cell content=\cellcolor{white}\underline{##1}}},
every row 4 column 4/.style={postproc cell content/.style={@cell content=\cellcolor{white}\underline{##1}}},
every row 4 column 5/.style={postproc cell content/.style={@cell content=\cellcolor{white}\underline{##1}}},
every row 2 column 6/.style={postproc cell content/.style={@cell content=\cellcolor{white}\underline{##1}}},
every row 4 column 7/.style={postproc cell content/.style={@cell content=\cellcolor{white}\underline{##1}}},
every row 4 column 8/.style={postproc cell content/.style={@cell content=\cellcolor{white}\underline{##1}}},
every row 6 column 0/.style={postproc cell content/.style={@cell content=\cellcolor{green!20}##1}}, 
every row 6 column 1/.style={postproc cell content/.style={@cell content=\cellcolor{white}##1}},
every row 7 column 0/.style={postproc cell content/.style={@cell content=\cellcolor{green!20}##1}}, 
every row 7 column 1/.style={postproc cell content/.style={@cell content=\cellcolor{white}##1}},
every row 8 column 0/.style={postproc cell content/.style={@cell content=\cellcolor{green!20}##1}}, 
every row 8 column 1/.style={postproc cell content/.style={@cell content=\cellcolor{white}##1}},
every row 9 column 0/.style={postproc cell content/.style={@cell content=\cellcolor{green!20}##1}}, 
every row 9 column 1/.style={postproc cell content/.style={@cell content=\cellcolor{white}##1}},
every row 7 column 1/.style={postproc cell content/.style={@cell content=\cellcolor{white}\underline{##1}}},
every row 7 column 2/.style={postproc cell content/.style={@cell content=\cellcolor{white}\underline{##1}}},
every row 9 column 3/.style={postproc cell content/.style={@cell content=\cellcolor{white}\underline{##1}}},
every row 9 column 4/.style={postproc cell content/.style={@cell content=\cellcolor{white}\underline{##1}}},
every row 8 column 5/.style={postproc cell content/.style={@cell content=\cellcolor{white}\underline{##1}}},
every row 8 column 6/.style={postproc cell content/.style={@cell content=\cellcolor{white}\underline{##1}}},
every row 9 column 7/.style={postproc cell content/.style={@cell content=\cellcolor{white}\underline{##1}}},
every row 8 column 8/.style={postproc cell content/.style={@cell content=\cellcolor{white}\underline{##1}}},
every row 10 column 0/.style={postproc cell content/.style={@cell content=\cellcolor{blue!20}##1}}, 
every row 10 column 1/.style={postproc cell content/.style={@cell content=\cellcolor{white}##1}},
every row 11 column 0/.style={postproc cell content/.style={@cell content=\cellcolor{blue!20}##1}}, 
every row 11 column 1/.style={postproc cell content/.style={@cell content=\cellcolor{white}##1}},
every row 12 column 0/.style={postproc cell content/.style={@cell content=\cellcolor{blue!20}##1}}, 
every row 12 column 1/.style={postproc cell content/.style={@cell content=\cellcolor{white}##1}},
every row 13 column 0/.style={postproc cell content/.style={@cell content=\cellcolor{blue!20}##1}}, 
every row 13 column 1/.style={postproc cell content/.style={@cell content=\cellcolor{white}##1}},
every row 14 column 0/.style={postproc cell content/.style={@cell content=\cellcolor{blue!20}##1}}, 
every row 14 column 1/.style={postproc cell content/.style={@cell content=\cellcolor{white}##1}},
every row 15 column 0/.style={postproc cell content/.style={@cell content=\cellcolor{blue!20}##1}}, 
every row 15 column 1/.style={postproc cell content/.style={@cell content=\cellcolor{white}##1}},
every row 10 column 1/.style={postproc cell content/.style={@cell content=\cellcolor{white}\underline{##1}}},
every row 10 column 2/.style={postproc cell content/.style={@cell content=\cellcolor{white}\underline{##1}}},
every row 10 column 3/.style={postproc cell content/.style={@cell content=\cellcolor{white}\underline{##1}}},
every row 10 column 4/.style={postproc cell content/.style={@cell content=\cellcolor{white}\underline{##1}}},
every row 10 column 5/.style={postproc cell content/.style={@cell content=\cellcolor{white}\underline{##1}}},
every row 10 column 6/.style={postproc cell content/.style={@cell content=\cellcolor{white}\underline{##1}}},
every row 13 column 7/.style={postproc cell content/.style={@cell content=\cellcolor{white}\underline{##1}}},
every row 10 column 8/.style={postproc cell content/.style={@cell content=\cellcolor{white}\underline{##1}}},
every row 16 column 0/.style={postproc cell content/.style={@cell content=\cellcolor{orange!20}##1}}, 
every row 16 column 1/.style={postproc cell content/.style={@cell content=\cellcolor{white}##1}},
every row 17 column 0/.style={postproc cell content/.style={@cell content=\cellcolor{orange!20}##1}}, 
every row 17 column 1/.style={postproc cell content/.style={@cell content=\cellcolor{white}##1}},
every row 18 column 0/.style={postproc cell content/.style={@cell content=\cellcolor{orange!20}##1}}, 
every row 18 column 1/.style={postproc cell content/.style={@cell content=\cellcolor{white}##1}},
every row 19 column 0/.style={postproc cell content/.style={@cell content=\cellcolor{orange!20}##1}}, 
every row 19 column 1/.style={postproc cell content/.style={@cell content=\cellcolor{white}##1}},
every row 20 column 0/.style={postproc cell content/.style={@cell content=\cellcolor{orange!20}##1}}, 
every row 20 column 1/.style={postproc cell content/.style={@cell content=\cellcolor{white}##1}},
every row 21 column 0/.style={postproc cell content/.style={@cell content=\cellcolor{orange!20}##1}}, 
every row 21 column 1/.style={postproc cell content/.style={@cell content=\cellcolor{white}##1}},
every row 21 column 7/.style={postproc cell content/.style={@cell content=\underline{\textbf{##1}}}}, 
every row 22 column 0/.style={postproc cell content/.style={@cell content=\cellcolor{orange!20}##1}}, 
every row 22 column 1/.style={postproc cell content/.style={@cell content=\cellcolor{white}##1}},
every row 23 column 0/.style={postproc cell content/.style={@cell content=\cellcolor{orange!20}##1}}, 
every row 23 column 1/.style={postproc cell content/.style={@cell content=\cellcolor{white}\underline{\textbf{##1}}}},
every row 23 column 2/.style={postproc cell content/.style={@cell content=\underline{\textbf{##1}}}},
every row 23 column 3/.style={postproc cell content/.style={@cell content=\underline{\textbf{##1}}}},
every row 23 column 4/.style={postproc cell content/.style={@cell content=\underline{\textbf{##1}}}},
every row 23 column 5/.style={postproc cell content/.style={@cell content=\underline{\textbf{##1}}}},
every row 23 column 6/.style={postproc cell content/.style={@cell content=\underline{\textbf{##1}}}},
every row 23 column 8/.style={postproc cell content/.style={@cell content=\underline{\textbf{##1}}}},
every row 24 column 0/.style={postproc cell content/.style={@cell content=\cellcolor{cyan!20}##1}}, 
every row 24 column 1/.style={postproc cell content/.style={@cell content=\cellcolor{white}##1}},
every row 25 column 0/.style={postproc cell content/.style={@cell content=\cellcolor{cyan!20}##1}}, 
every row 25 column 1/.style={postproc cell content/.style={@cell content=\cellcolor{white}##1}},
every row 26 column 0/.style={postproc cell content/.style={@cell content=\cellcolor{cyan!20}##1}}, 
every row 26 column 1/.style={postproc cell content/.style={@cell content=\cellcolor{white}##1}},
every row 27 column 0/.style={postproc cell content/.style={@cell content=\cellcolor{cyan!20}##1}}, 
every row 27 column 1/.style={postproc cell content/.style={@cell content=\cellcolor{white}##1}},
every row 28 column 0/.style={postproc cell content/.style={@cell content=\cellcolor{cyan!20}##1}}, 
every row 28 column 1/.style={postproc cell content/.style={@cell content=\cellcolor{white}##1}},
every row 29 column 0/.style={postproc cell content/.style={@cell content=\cellcolor{cyan!20}##1}}, 
every row 29 column 1/.style={postproc cell content/.style={@cell content=\cellcolor{white}##1}},
every row 30 column 0/.style={postproc cell content/.style={@cell content=\cellcolor{cyan!20}##1}}, 
every row 30 column 1/.style={postproc cell content/.style={@cell content=\cellcolor{white}##1}},
every row 28 column 1/.style={postproc cell content/.style={@cell content=\cellcolor{white}\underline{##1}}},
every row 29 column 2/.style={postproc cell content/.style={@cell content=\cellcolor{white}\underline{##1}}},
every row 28 column 3/.style={postproc cell content/.style={@cell content=\cellcolor{white}\underline{##1}}},
every row 28 column 4/.style={postproc cell content/.style={@cell content=\cellcolor{white}\underline{##1}}},
every row 25 column 5/.style={postproc cell content/.style={@cell content=\cellcolor{white}\underline{##1}}},
every row 28 column 6/.style={postproc cell content/.style={@cell content=\cellcolor{white}\underline{##1}}},
every row 25 column 7/.style={postproc cell content/.style={@cell content=\cellcolor{white}\underline{##1}}},
every row 25 column 8/.style={postproc cell content/.style={@cell content=\cellcolor{white}\underline{##1}}},
]\mytable
\end{adjustbox}
\centering
\footnotesize{\colorbox{red!20}{ Cognitive/Biological }, \colorbox{green!20}{ Information-Theoretic }, \colorbox{blue!20}{ Probabilistic}, \colorbox{orange!20}{   Fourier/Spectral}, \colorbox{cyan!20}{   Machine/Deep Learning}}
\end{table}

\begin{figure}[h!]
\centering 
\scriptsize
\makebox[7em]{Image}
\fbox{\includegraphics[width=0.10\linewidth]{images/d1Bfv1BdefaultB15.jpg}}
\fbox{\includegraphics[width=0.10\linewidth]{images/d2Bfv5BproximityGdissimilarB4p1667.jpg}}
\fbox{\includegraphics[width=0.10\linewidth]{images/d2Bvs2BcirclebarGcircleB2p5.jpg}}
\fbox{\includegraphics[width=0.10\linewidth]{images/d1Bvs4BrTGwBB1.jpg}}
\fbox{\includegraphics[width=0.10\linewidth]{images/d2Bvs5BwTGhBB0p5.jpg}}
\fbox{\includegraphics[width=0.10\linewidth]{images/d1Bvs6BdefaultB5.jpg}}
\fbox{\includegraphics[width=0.10\linewidth]{images/d1Bvs7BdefaultB41p8103.jpg}}
\\\makebox[7em]{Humans}
\fbox{\includegraphics[width=0.10\linewidth]{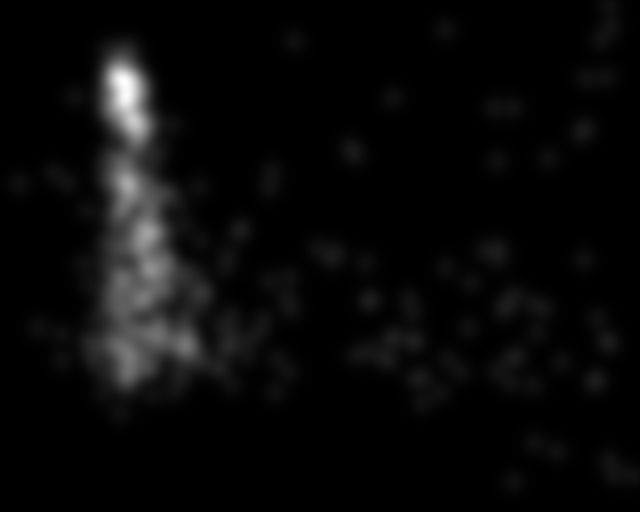}}
\fbox{\includegraphics[width=0.10\linewidth]{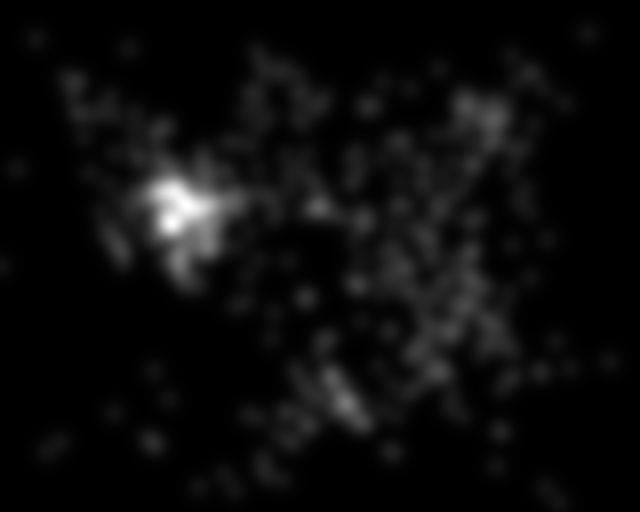}}
\fbox{\includegraphics[width=0.10\linewidth]{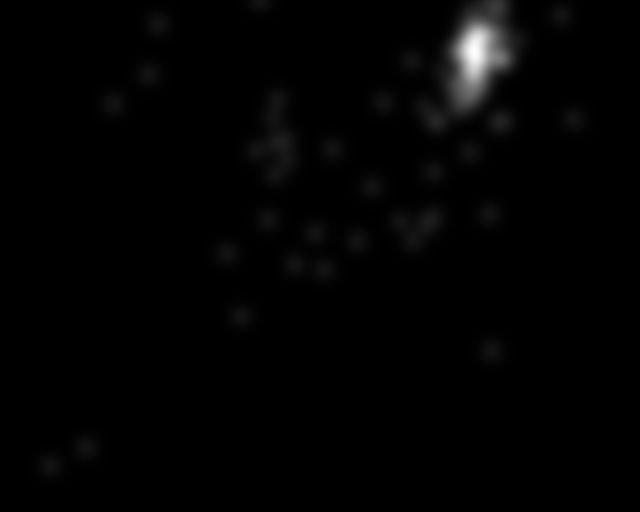}}
\fbox{\includegraphics[width=0.10\linewidth]{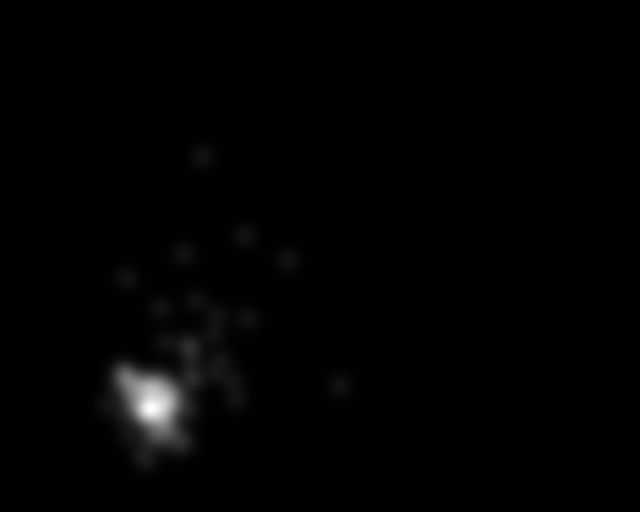}}
\fbox{\includegraphics[width=0.10\linewidth]{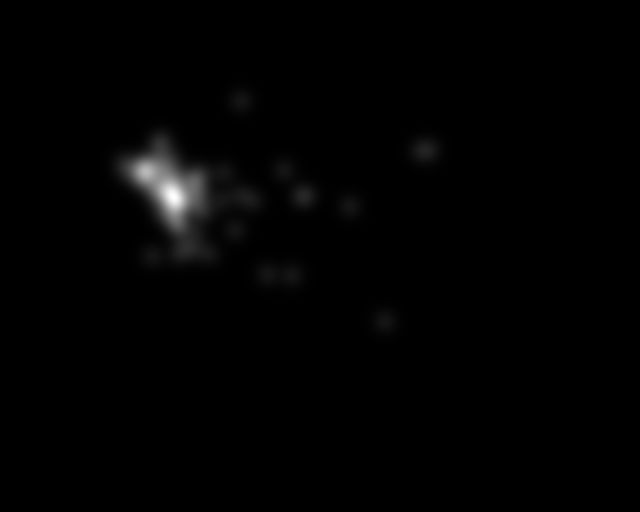}}
\fbox{\includegraphics[width=0.10\linewidth]{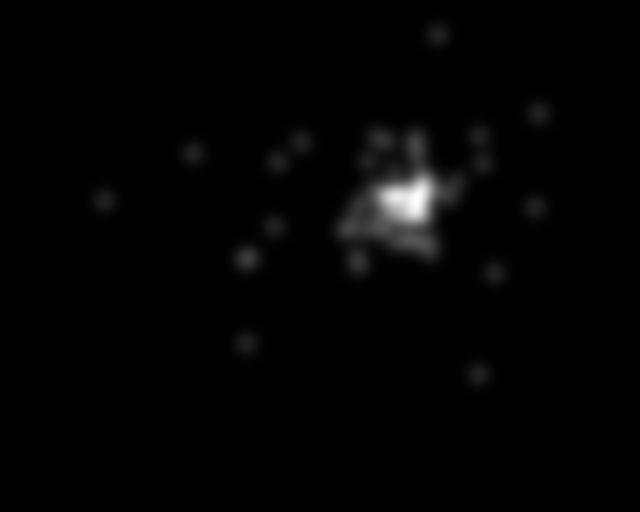}}
\fbox{\includegraphics[width=0.10\linewidth]{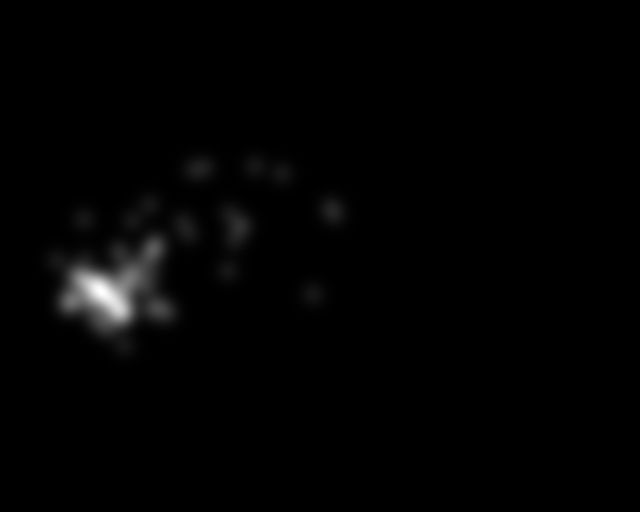}}
\\\makebox[7em]{AWS}
\fbox{\includegraphics[width=0.10\linewidth]{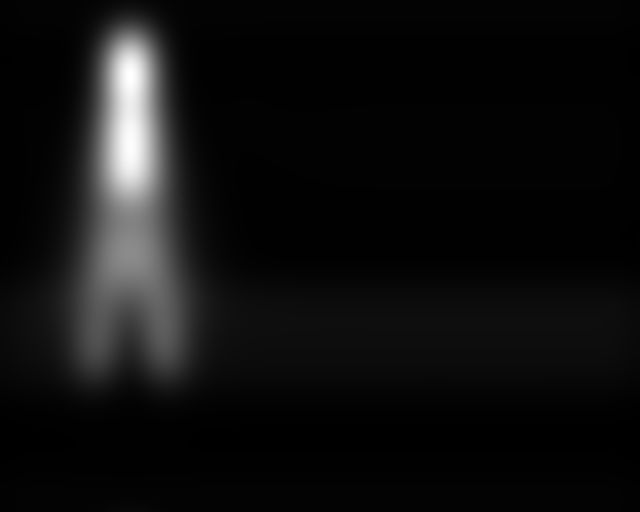}}
\fbox{\includegraphics[width=0.10\linewidth]{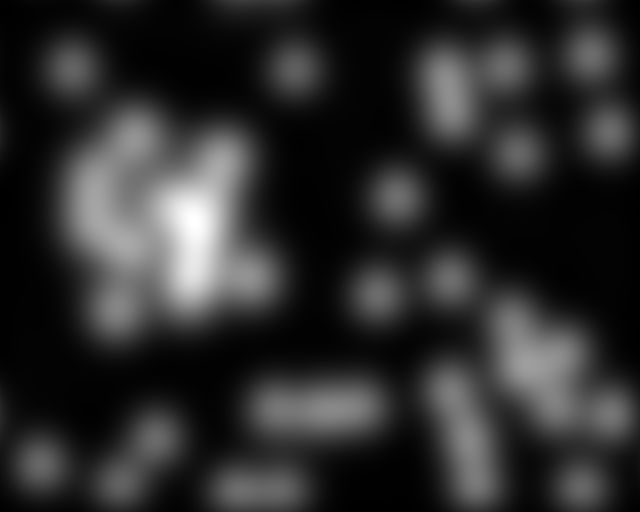}}
\fbox{\includegraphics[width=0.10\linewidth]{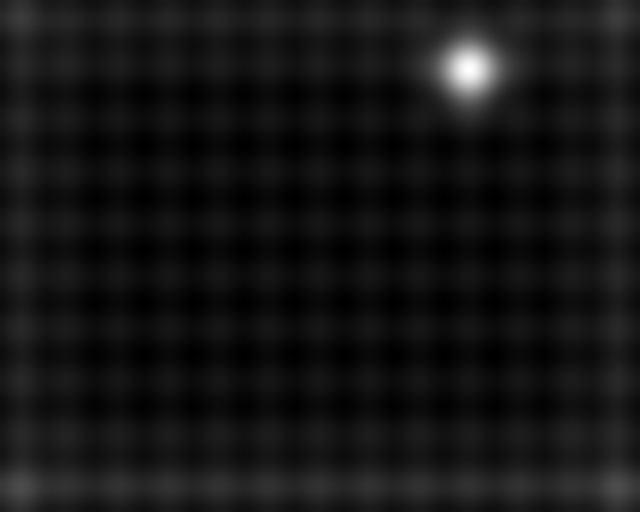}}
\fbox{\includegraphics[width=0.10\linewidth]{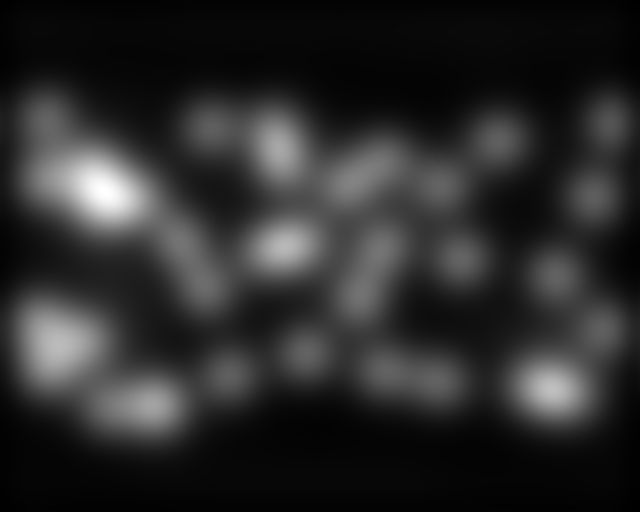}}
\fbox{\includegraphics[width=0.10\linewidth]{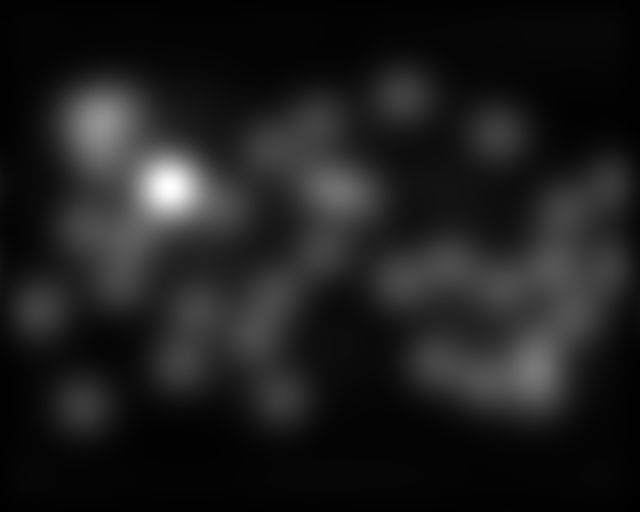}}
\fbox{\includegraphics[width=0.10\linewidth]{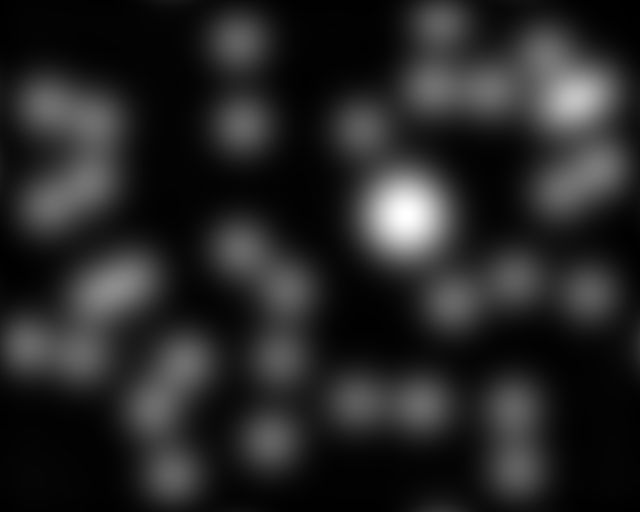}}
\fbox{\includegraphics[width=0.10\linewidth]{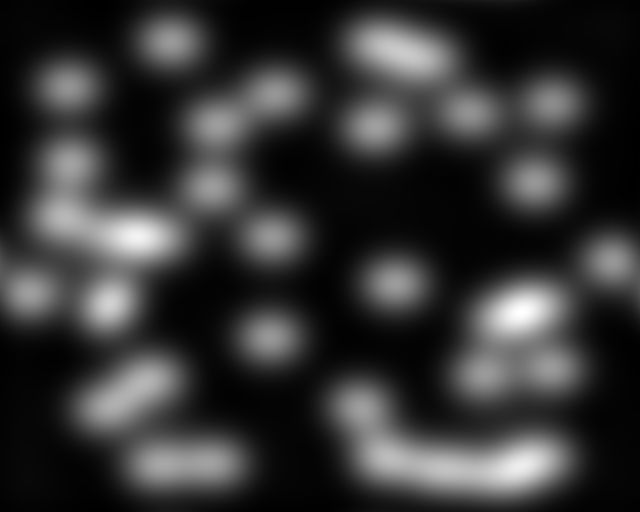}}
\\\makebox[7em]{NSWAM}
\fbox{\includegraphics[width=0.10\linewidth]{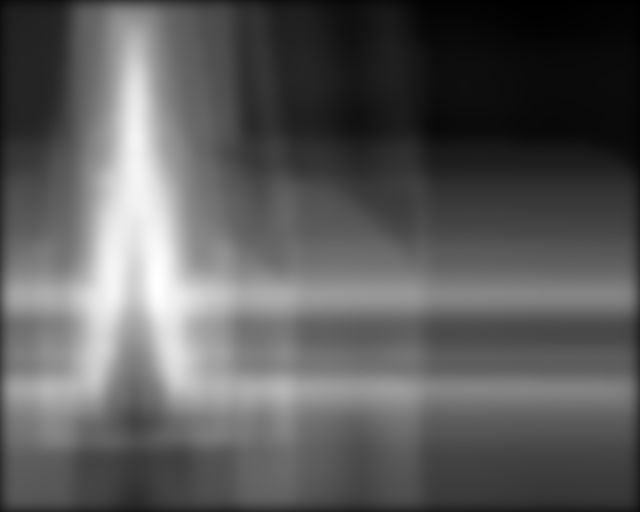}}
\fbox{\includegraphics[width=0.10\linewidth]{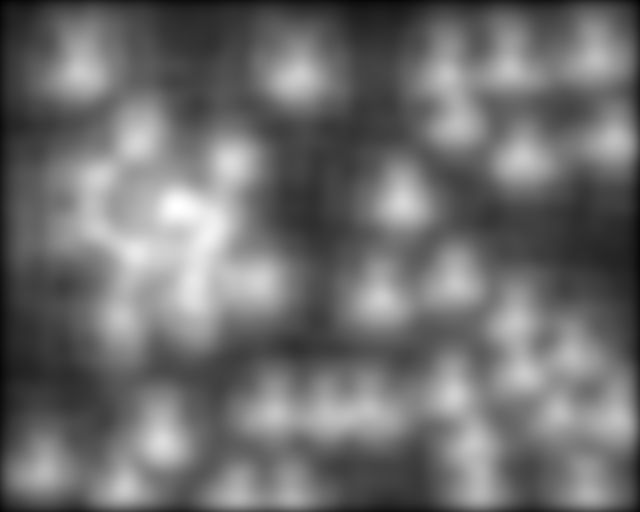}}
\fbox{\includegraphics[width=0.10\linewidth]{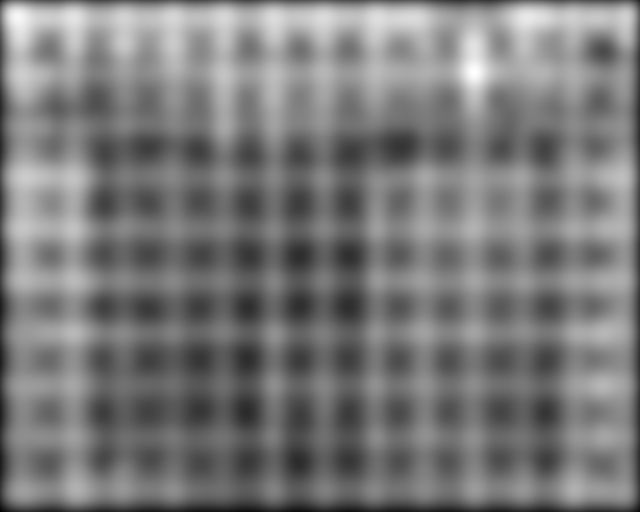}}
\fbox{\includegraphics[width=0.10\linewidth]{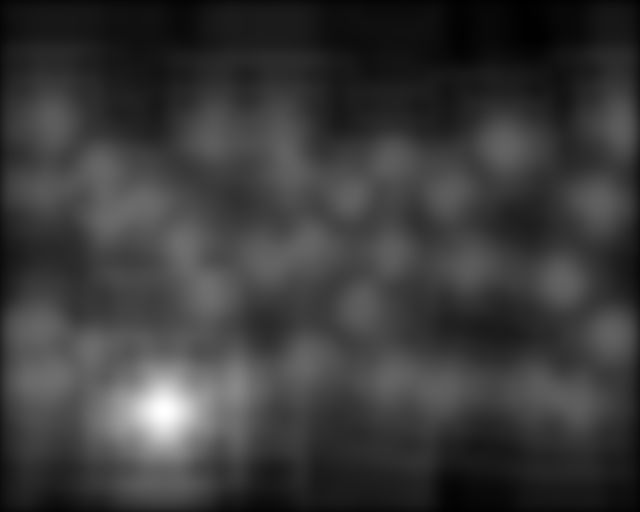}}
\fbox{\includegraphics[width=0.10\linewidth]{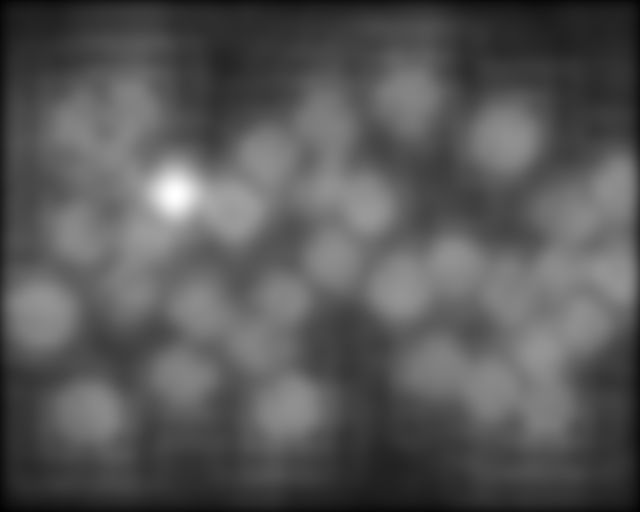}}
\fbox{\includegraphics[width=0.10\linewidth]{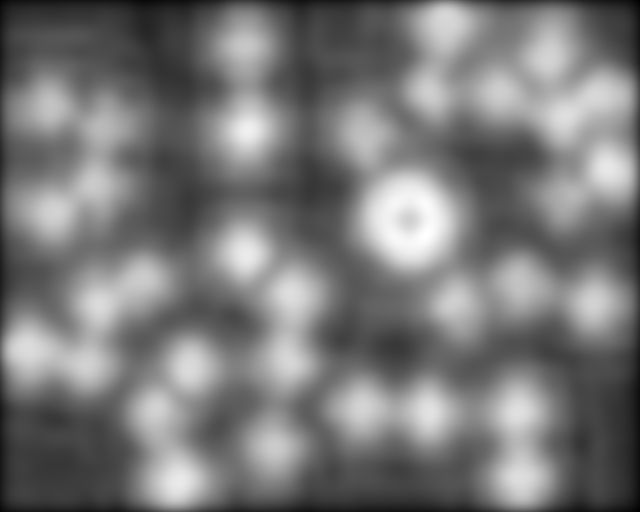}}
\fbox{\includegraphics[width=0.10\linewidth]{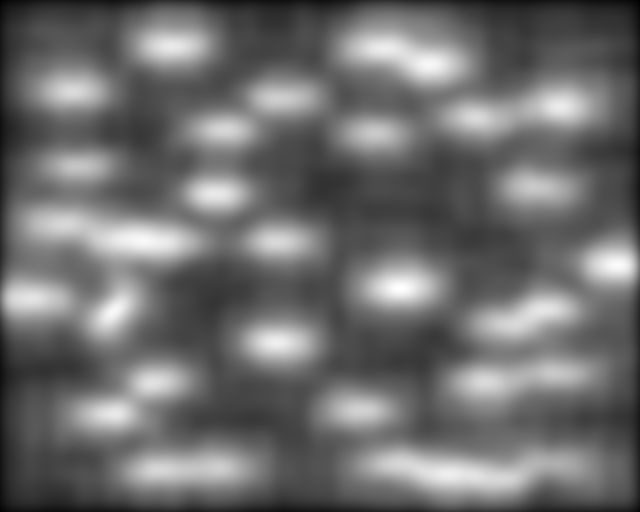}}
\\\makebox[7em]{RARE}
\fbox{\includegraphics[width=0.10\linewidth]{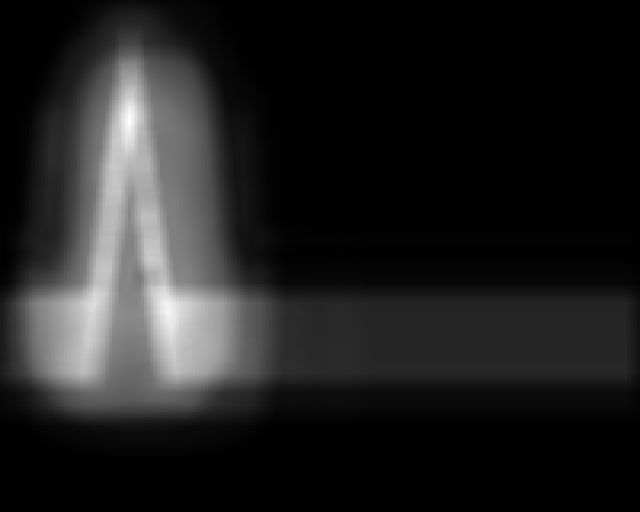}}
\fbox{\includegraphics[width=0.10\linewidth]{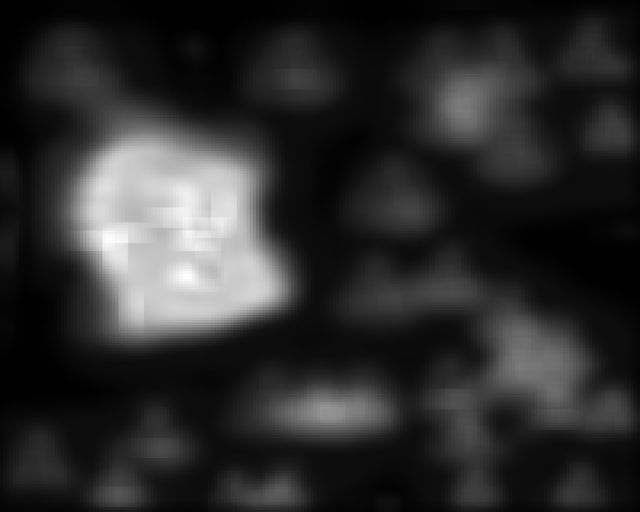}}
\fbox{\includegraphics[width=0.10\linewidth]{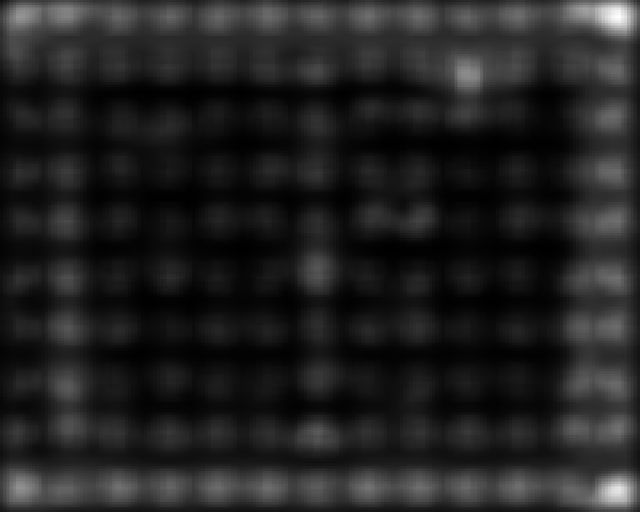}}
\fbox{\includegraphics[width=0.10\linewidth]{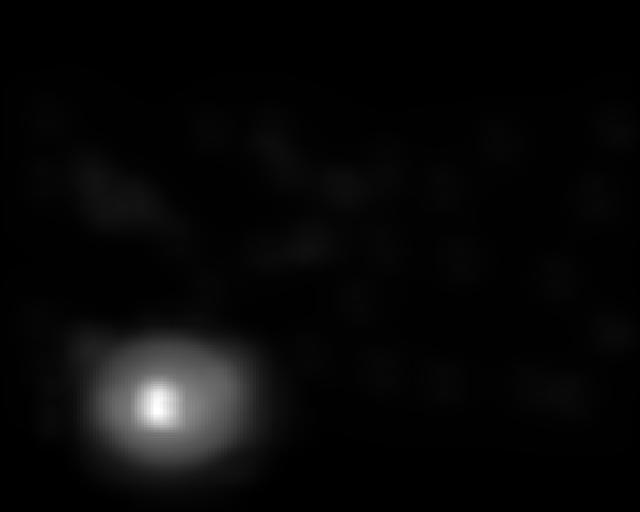}}
\fbox{\includegraphics[width=0.10\linewidth]{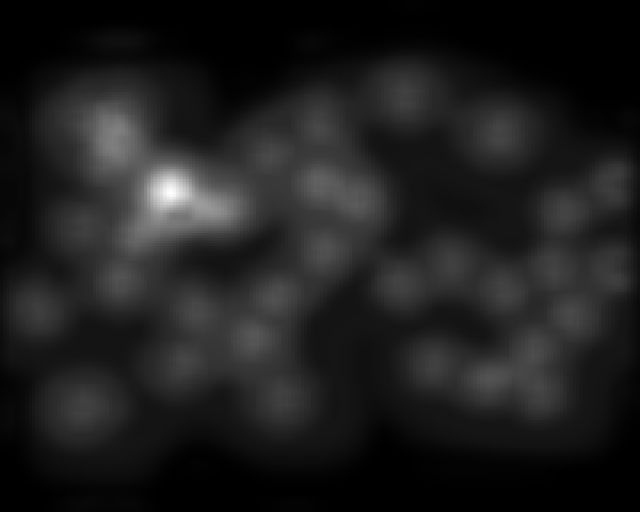}}
\fbox{\includegraphics[width=0.10\linewidth]{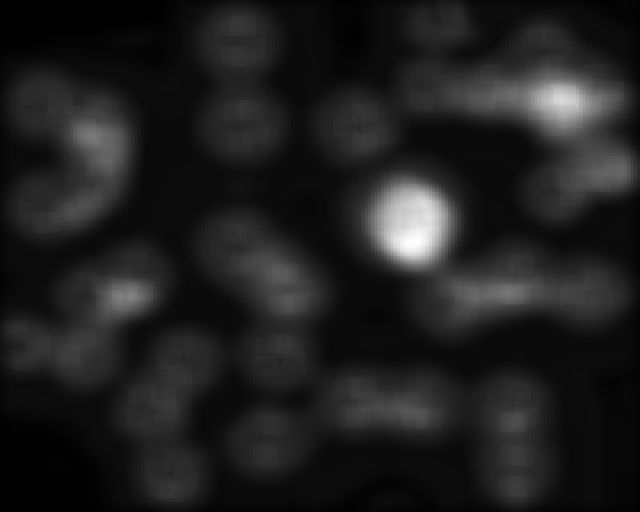}}
\fbox{\includegraphics[width=0.10\linewidth]{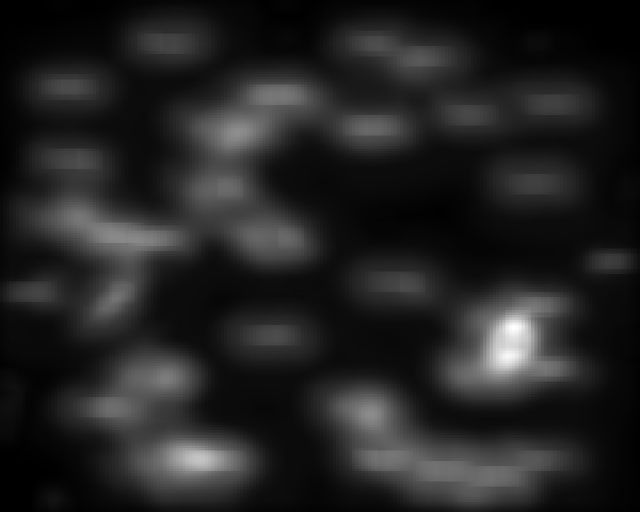}}
\\\makebox[7em]{CASD}
\fbox{\includegraphics[width=0.10\linewidth]{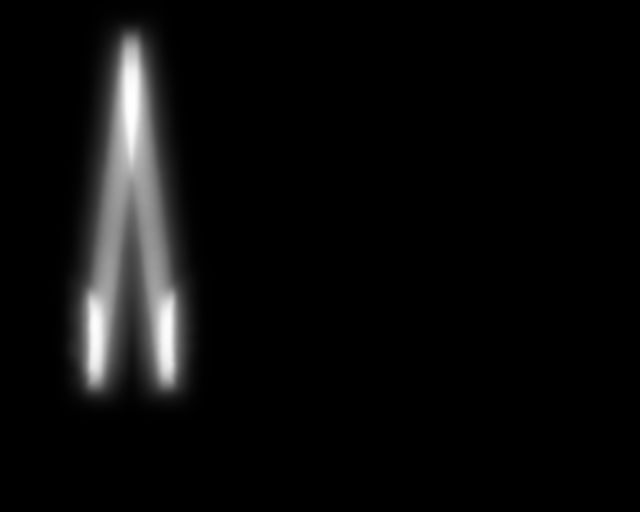}}
\fbox{\includegraphics[width=0.10\linewidth]{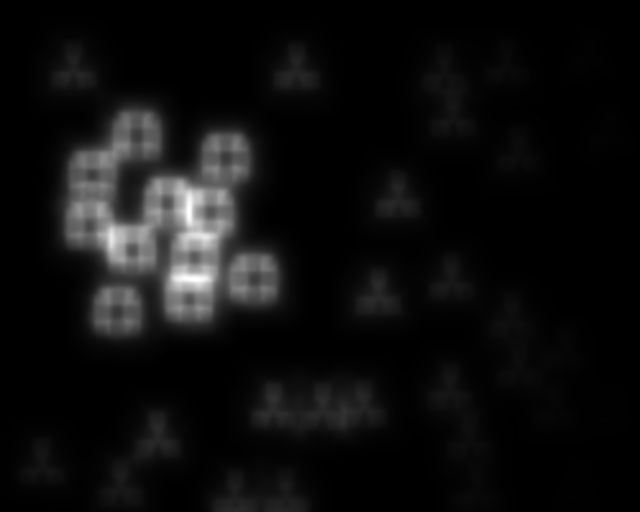}}
\fbox{\includegraphics[width=0.10\linewidth]{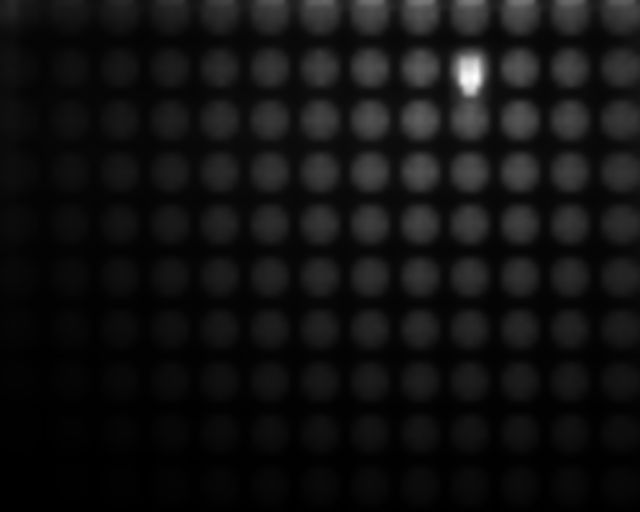}}
\fbox{\includegraphics[width=0.10\linewidth]{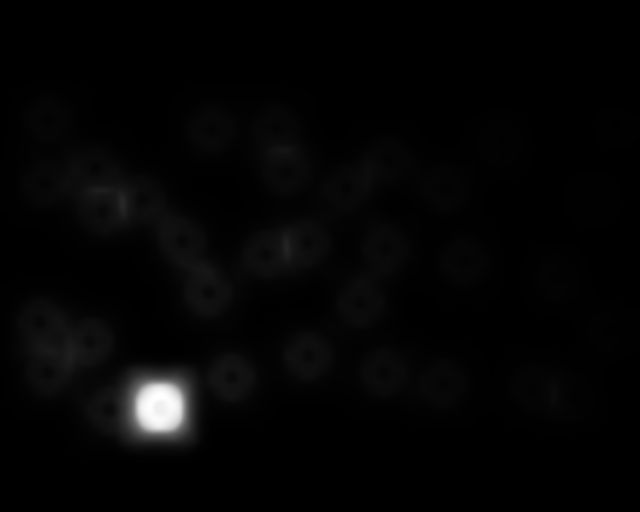}}
\fbox{\includegraphics[width=0.10\linewidth]{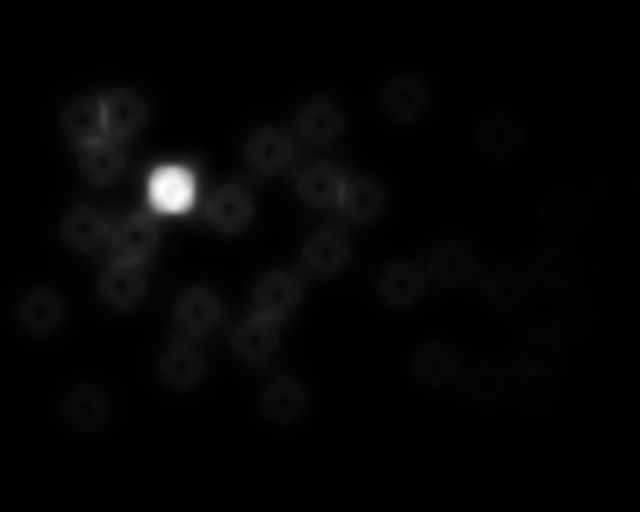}}
\fbox{\includegraphics[width=0.10\linewidth]{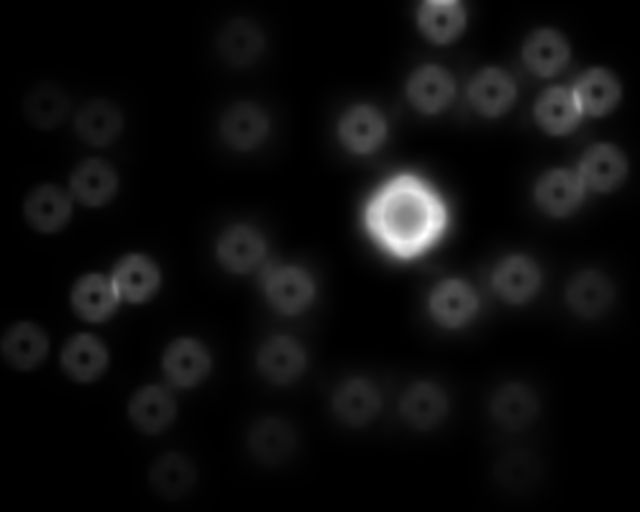}}
\fbox{\includegraphics[width=0.10\linewidth]{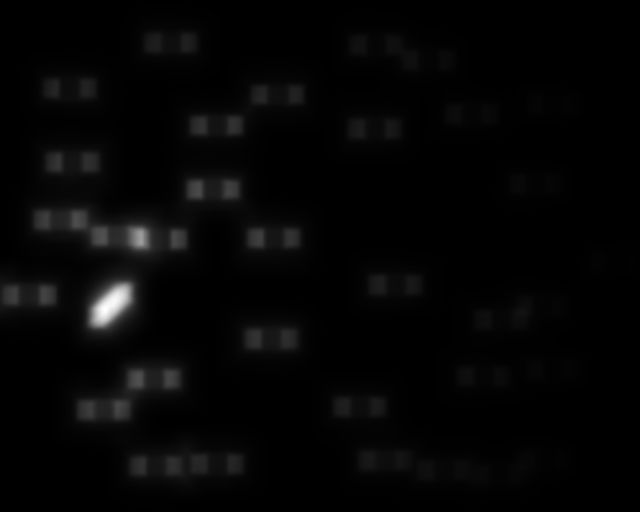}}
\\\makebox[7em]{GBVS}
\fbox{\includegraphics[width=0.10\linewidth]{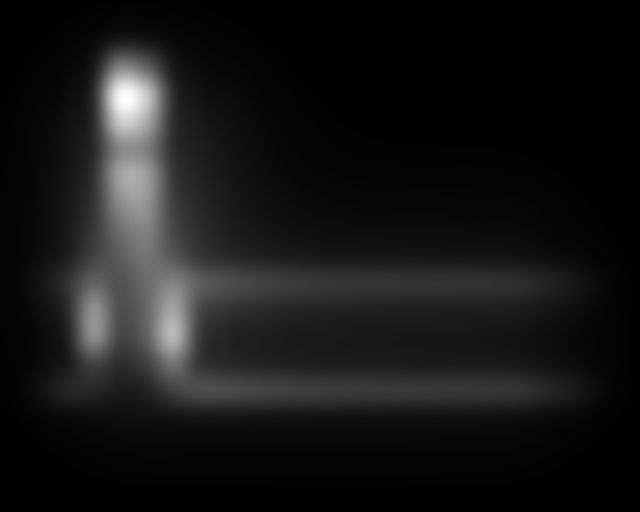}}
\fbox{\includegraphics[width=0.10\linewidth]{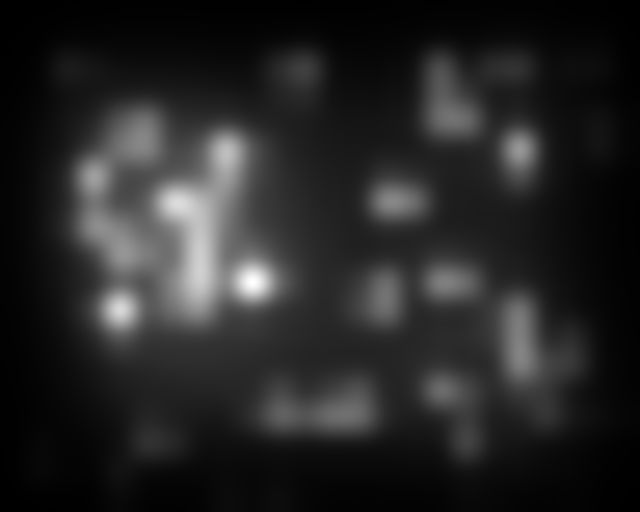}}
\fbox{\includegraphics[width=0.10\linewidth]{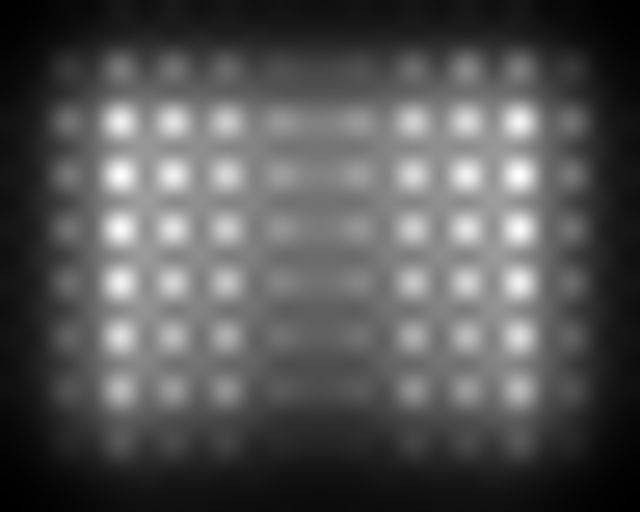}}
\fbox{\includegraphics[width=0.10\linewidth]{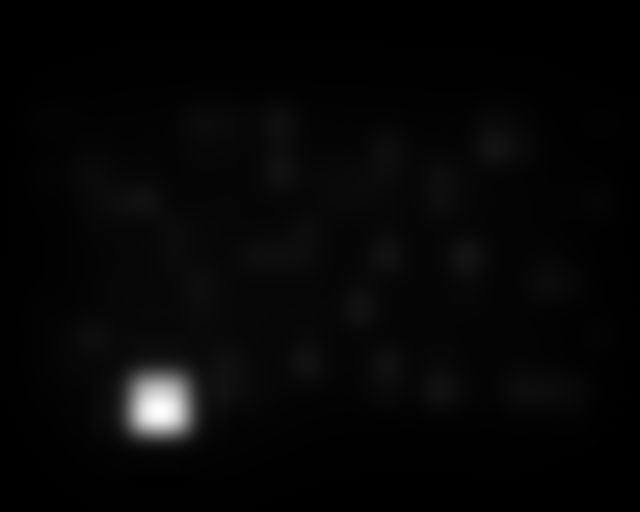}}
\fbox{\includegraphics[width=0.10\linewidth]{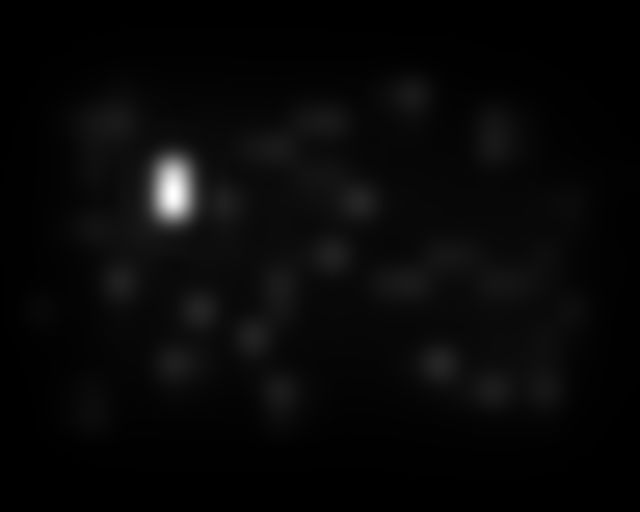}}
\fbox{\includegraphics[width=0.10\linewidth]{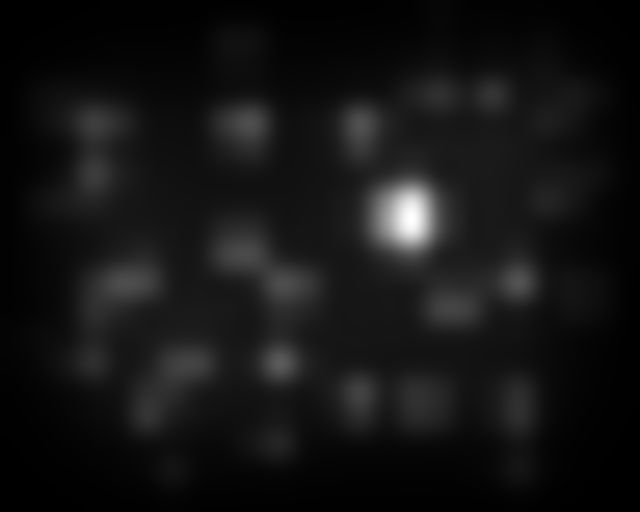}}
\fbox{\includegraphics[width=0.10\linewidth]{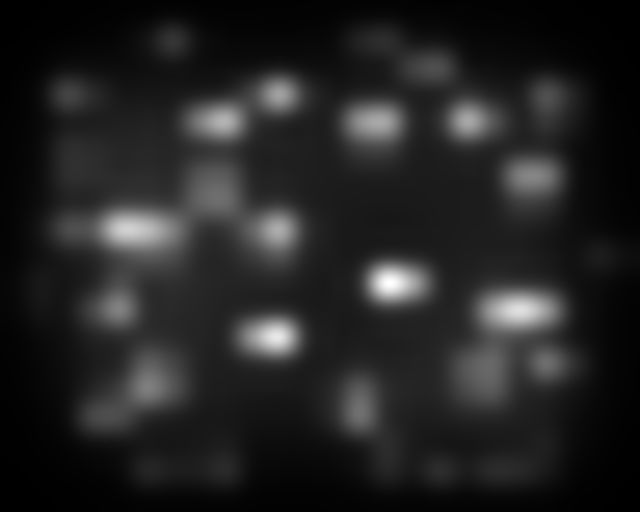}}
\\\makebox[7em]{SDSR}
\fbox{\includegraphics[width=0.10\linewidth]{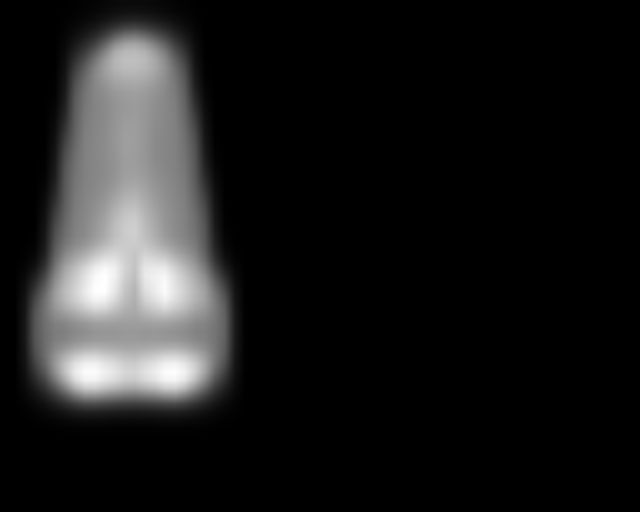}}
\fbox{\includegraphics[width=0.10\linewidth]{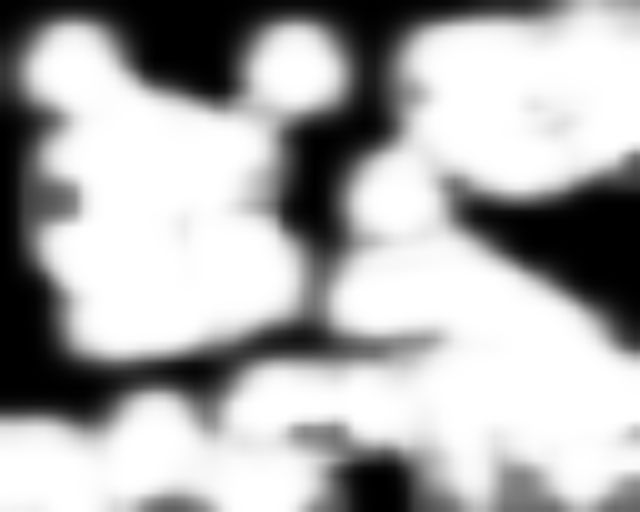}}
\fbox{\includegraphics[width=0.10\linewidth]{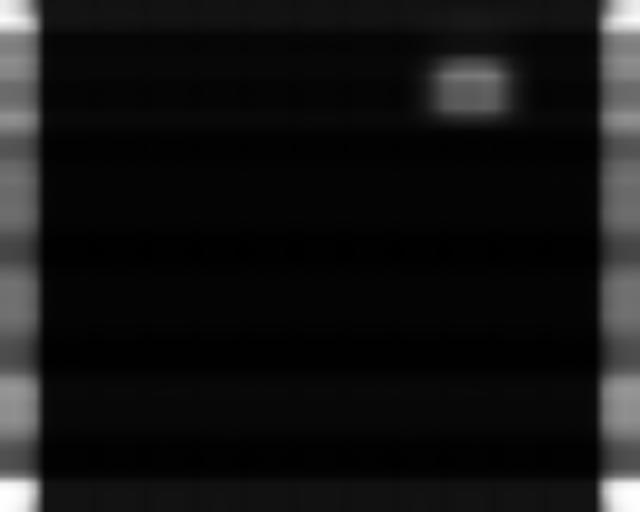}}
\fbox{\includegraphics[width=0.10\linewidth]{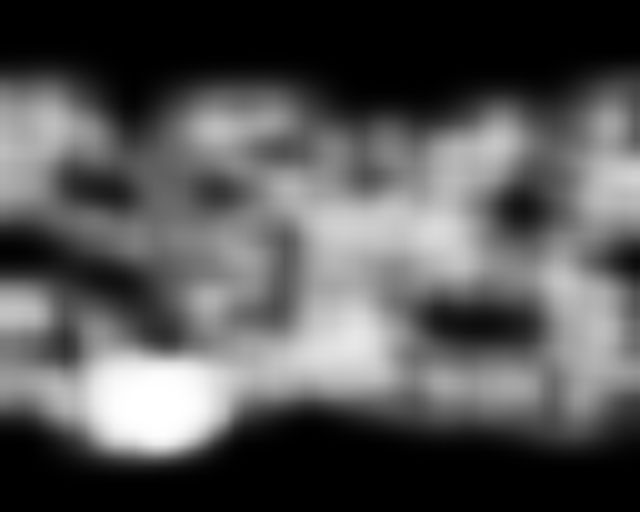}}
\fbox{\includegraphics[width=0.10\linewidth]{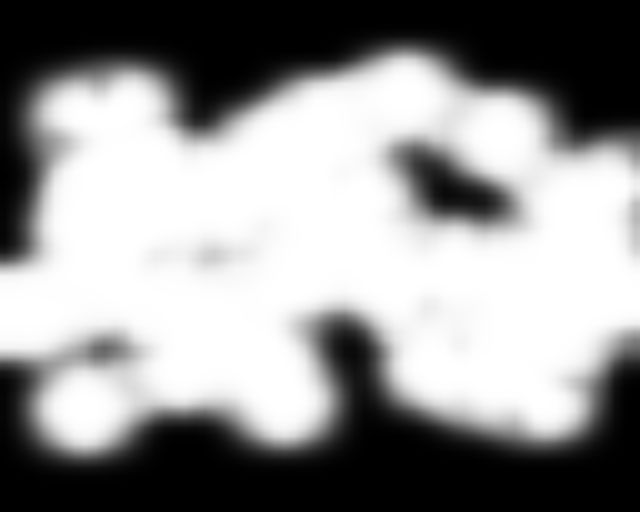}}
\fbox{\includegraphics[width=0.10\linewidth]{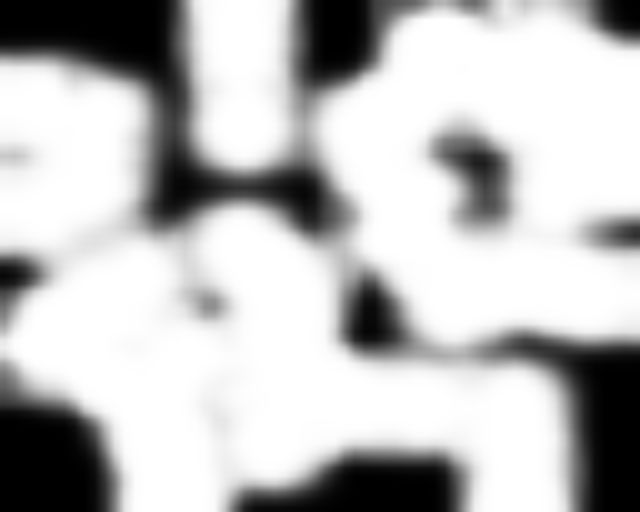}}
\fbox{\includegraphics[width=0.10\linewidth]{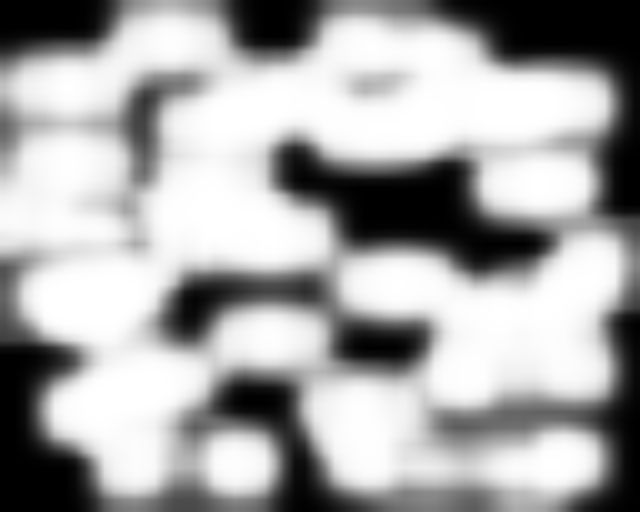}}
\\\makebox[7em]{WMAP}
\fbox{\includegraphics[width=0.10\linewidth]{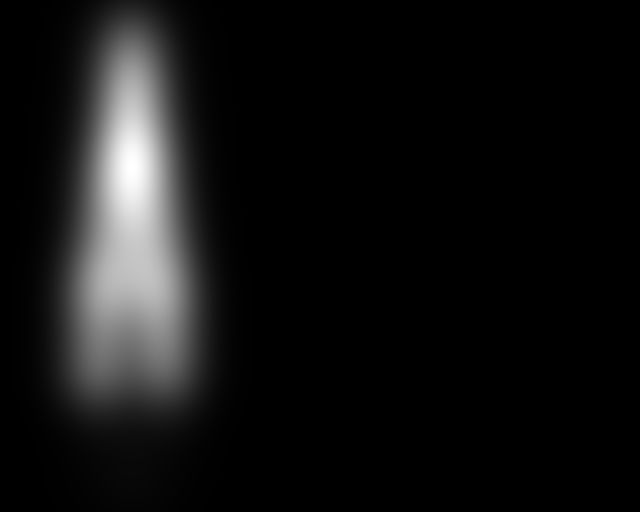}}
\fbox{\includegraphics[width=0.10\linewidth]{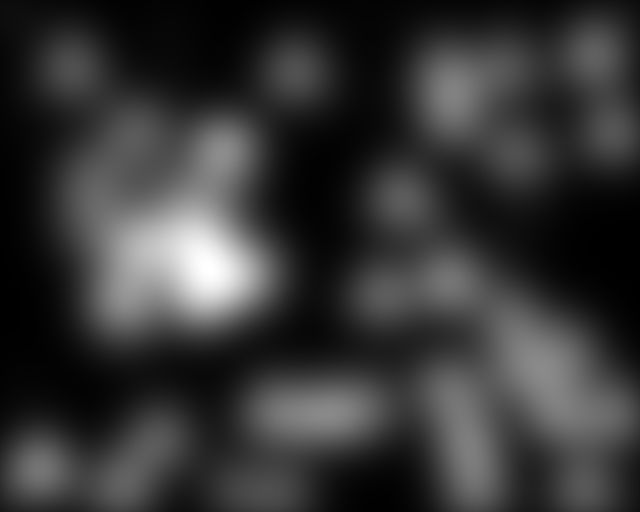}}
\fbox{\includegraphics[width=0.10\linewidth]{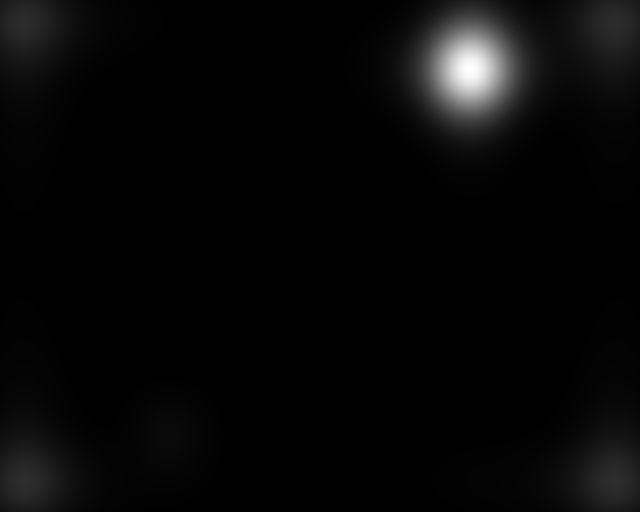}}
\fbox{\includegraphics[width=0.10\linewidth]{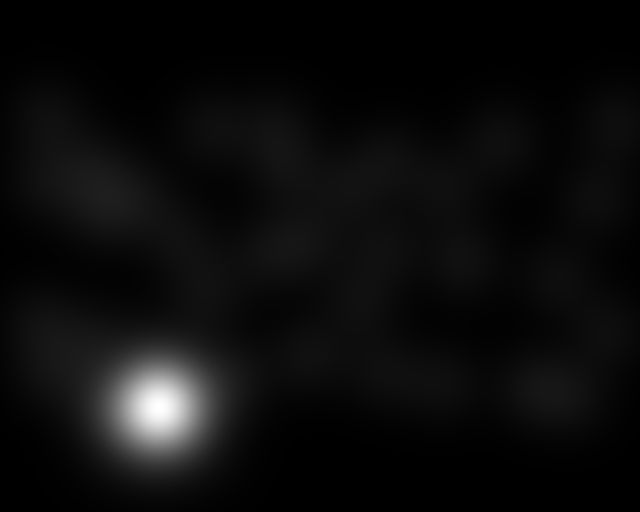}}
\fbox{\includegraphics[width=0.10\linewidth]{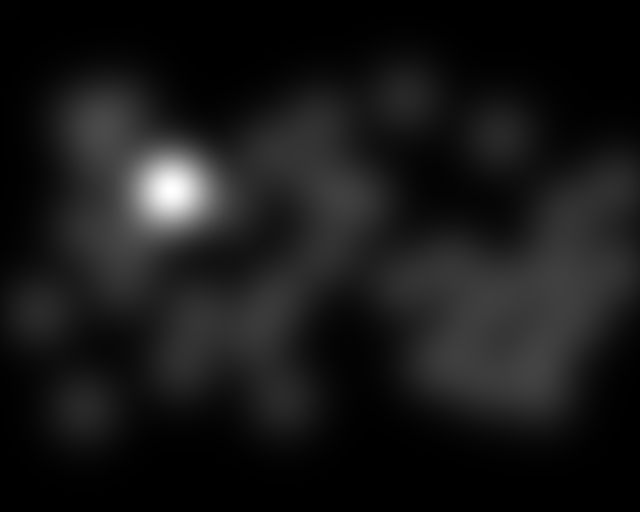}}
\fbox{\includegraphics[width=0.10\linewidth]{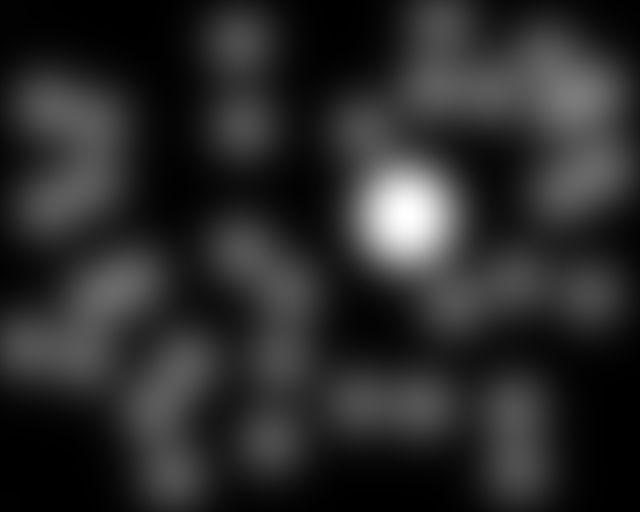}}
\fbox{\includegraphics[width=0.10\linewidth]{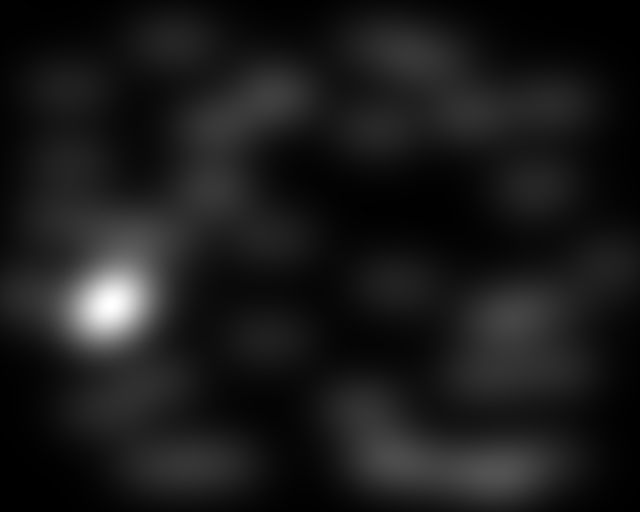}}
\\\makebox[7em]{HFT}
\fbox{\includegraphics[width=0.10\linewidth]{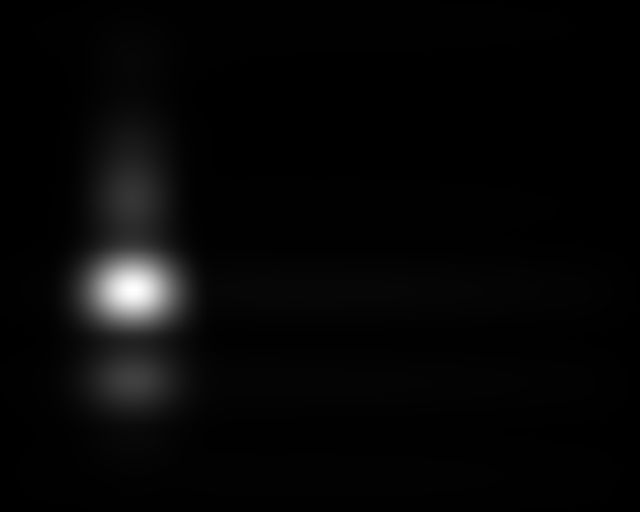}}
\fbox{\includegraphics[width=0.10\linewidth]{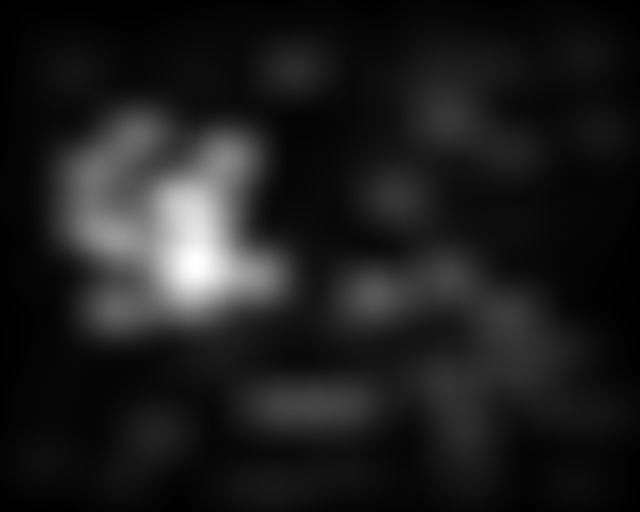}}
\fbox{\includegraphics[width=0.10\linewidth]{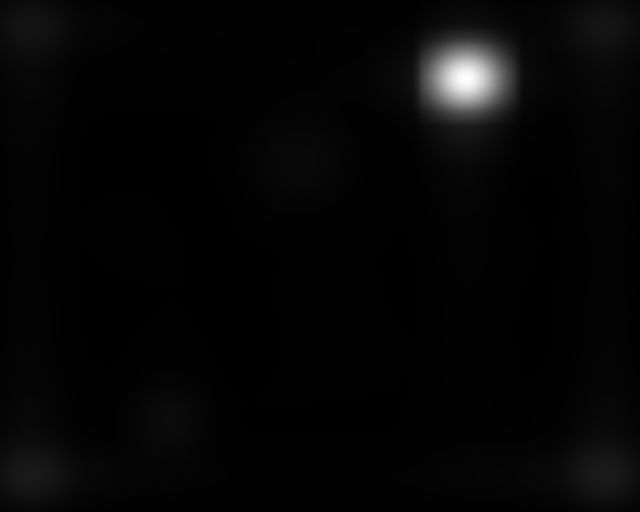}}
\fbox{\includegraphics[width=0.10\linewidth]{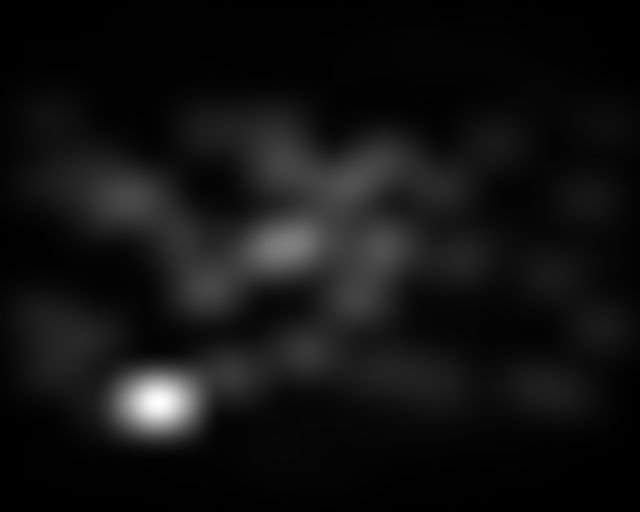}}
\fbox{\includegraphics[width=0.10\linewidth]{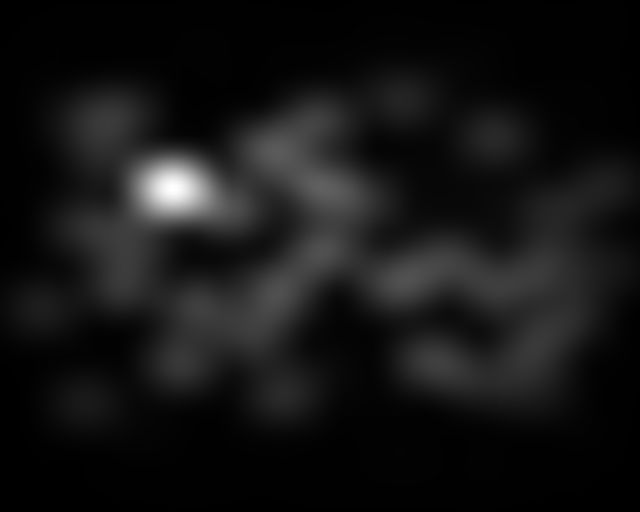}}
\fbox{\includegraphics[width=0.10\linewidth]{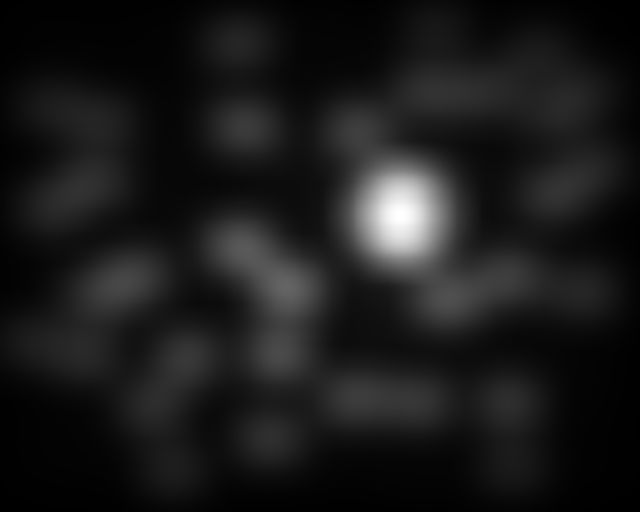}}
\fbox{\includegraphics[width=0.10\linewidth]{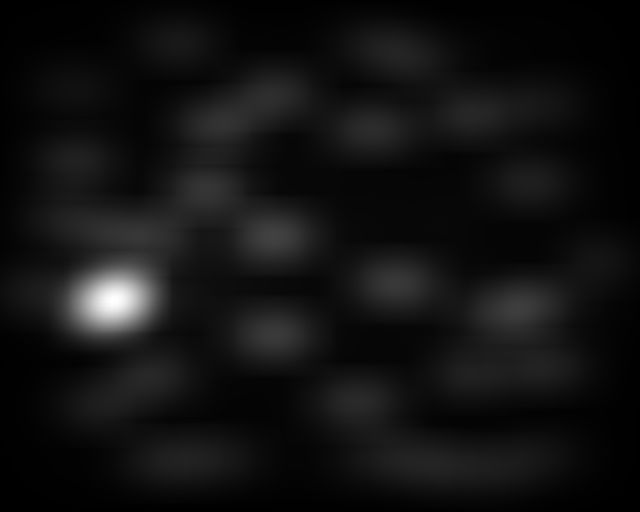}}
\\\makebox[7em]{OpenSALICON}
\fbox{\includegraphics[width=0.10\linewidth]{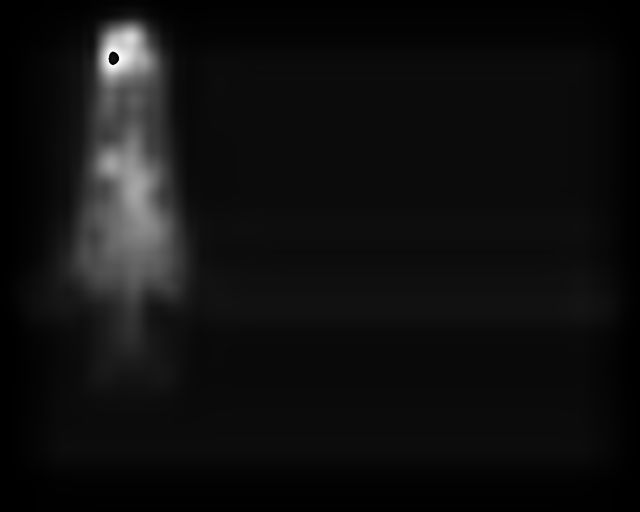}}
\fbox{\includegraphics[width=0.10\linewidth]{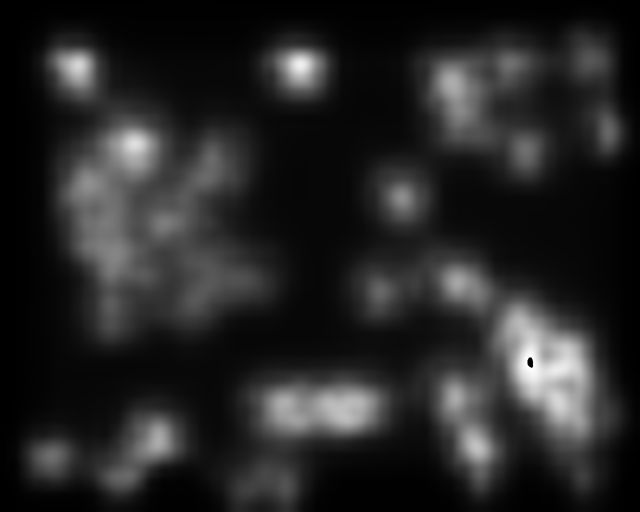}}
\fbox{\includegraphics[width=0.10\linewidth]{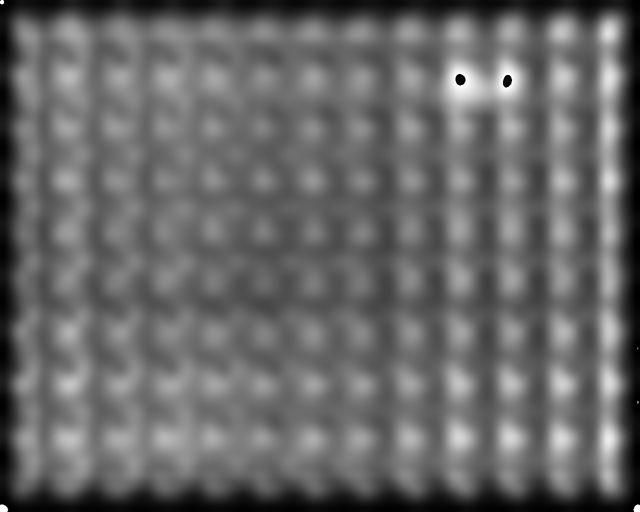}}
\fbox{\includegraphics[width=0.10\linewidth]{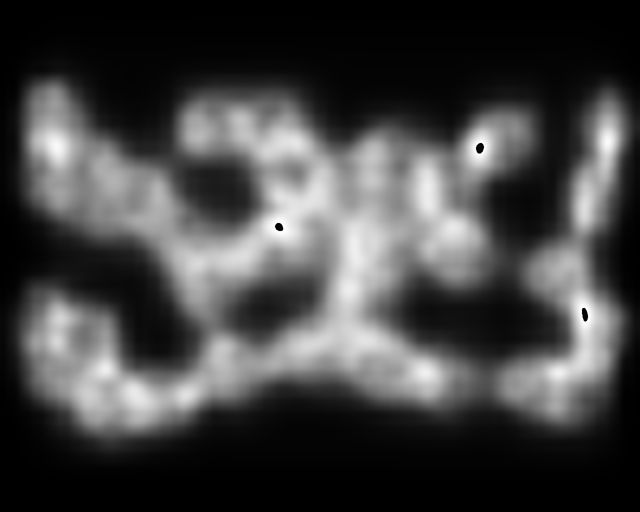}}
\fbox{\includegraphics[width=0.10\linewidth]{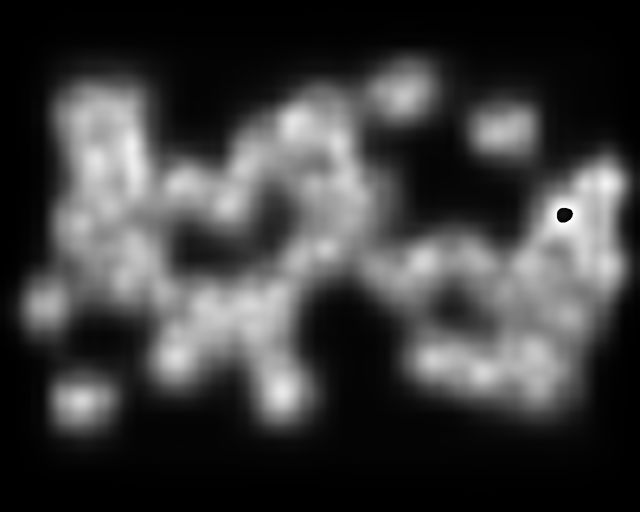}}
\fbox{\includegraphics[width=0.10\linewidth]{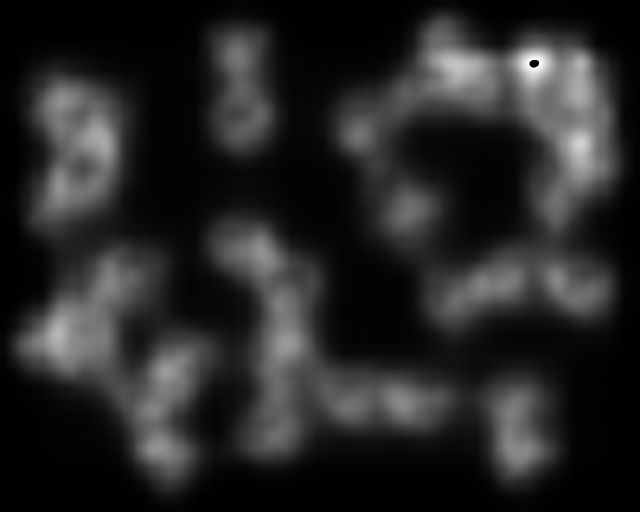}}
\fbox{\includegraphics[width=0.10\linewidth]{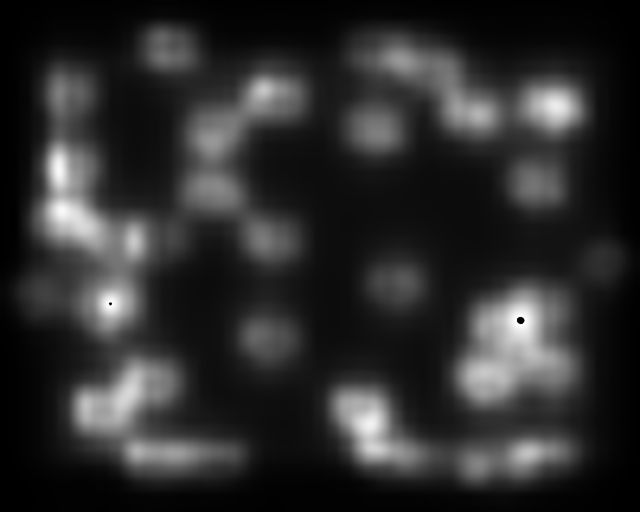}}
\\\makebox[7em]{SAM-ResNet}
\fbox{\includegraphics[width=0.10\linewidth]{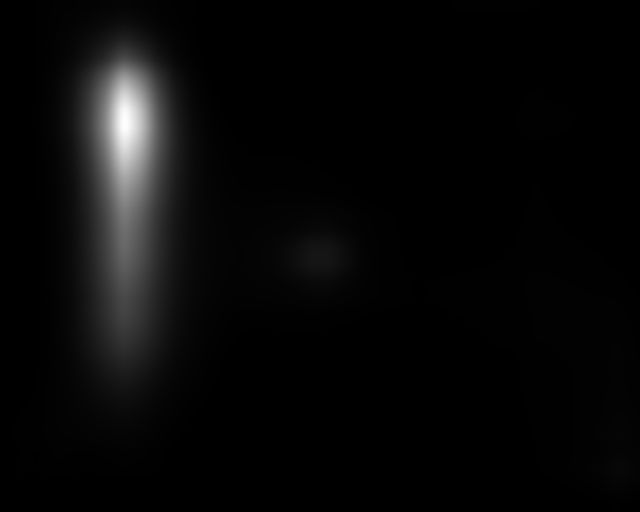}}
\fbox{\includegraphics[width=0.10\linewidth]{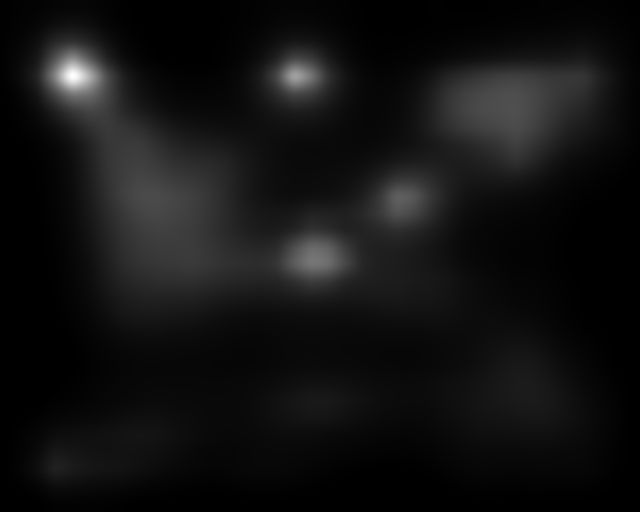}}
\fbox{\includegraphics[width=0.10\linewidth]{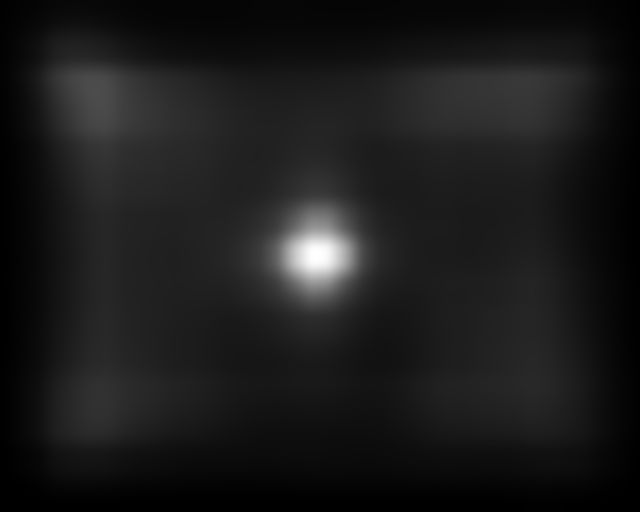}}
\fbox{\includegraphics[width=0.10\linewidth]{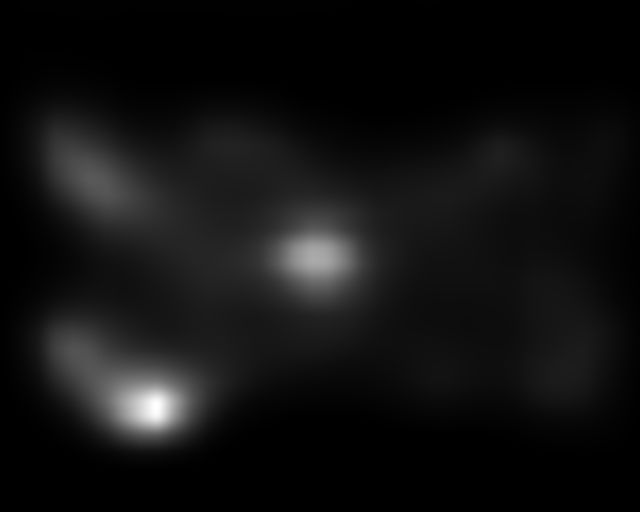}}
\fbox{\includegraphics[width=0.10\linewidth]{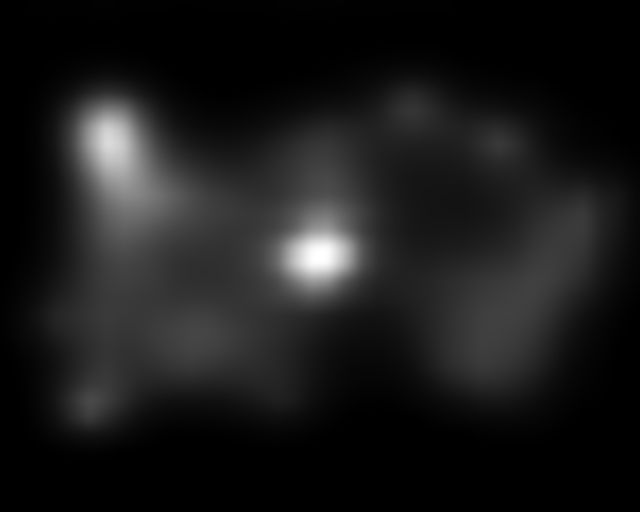}}
\fbox{\includegraphics[width=0.10\linewidth]{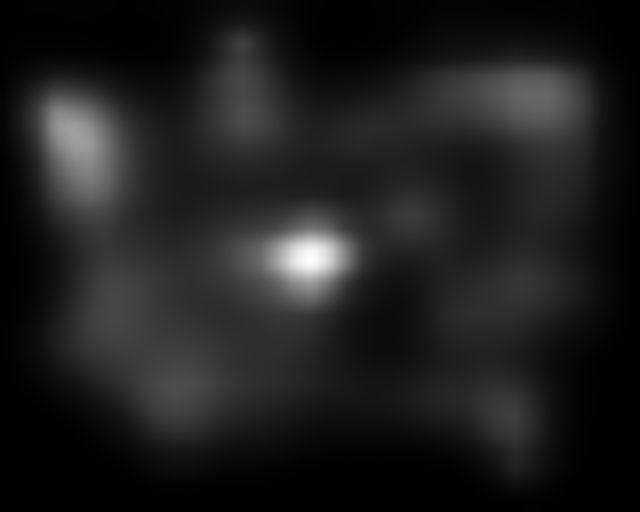}}
\fbox{\includegraphics[width=0.10\linewidth]{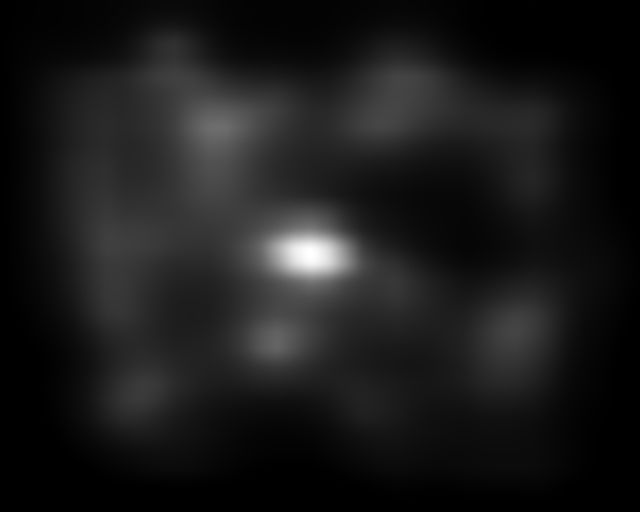}}
\caption{Examples of dataset stimuli and saliency map predictions. Only two models for each inspiration category that presented highest performance with shuffled saliency metric scores (sAUC and InfoGain) are shown.}
\label{fig:qualitative}
\end{figure}

\begin{figure}[h!]
\centering
\includegraphics[width=1\linewidth]{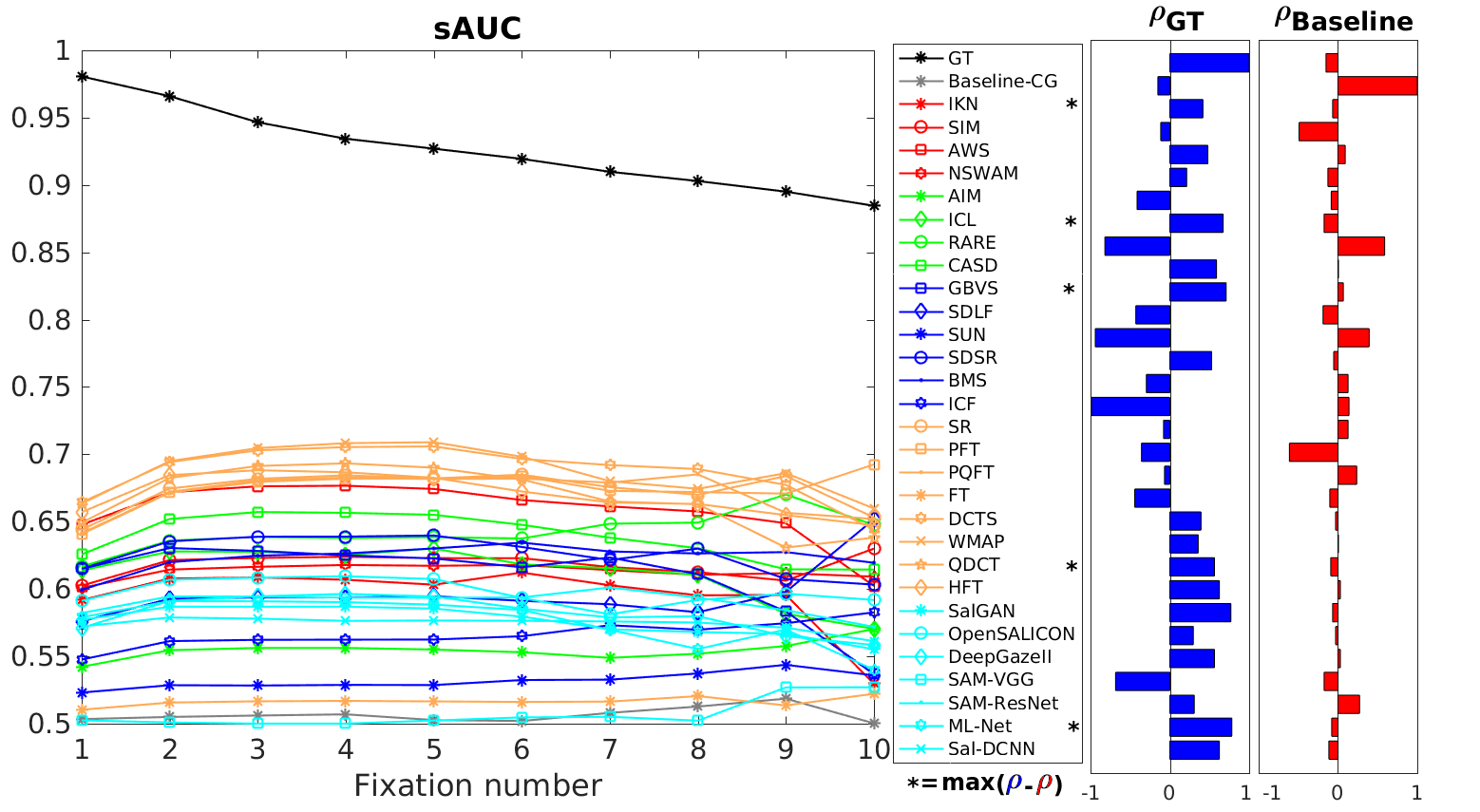} 
\caption{sAUC gaze-wise prediction scores.}
\label{fig:results_auc_gazewise}
\end{figure}

Models such as HFT and WMAP remarkably outpeform other saliency models. From other model inspirations, AWS score higher than other Cognitive/Biologically-inspired models, GBVS and CASD outperform other Probabilistic/Bayesian and Information-theoretic saliency models respectively. For Deep Learning models, SAM$_{ResNet}$ and OpenSALICON are the ones with highest scores. Although there are present differences in terms of model performances and model inspiration, similarities in model mechanisms can reveal phenomena of increasing and decreasing prediction statistics. This phenomena is present for Spectral/Fourier-based and Cognitive/Biologically-inspired models, withwhom all present similar performance and balanced scores throughout the distinct metric scores. It is to consider that sAUC and InfoGain metrics are more reliable compared to other metrics (which the baseline center gaussian sometimes acquires higher performance than most saliency models). In these terms, models shown on \hyperref[fig:qualitative]{Fig. \ref*{fig:qualitative}} are efficient saliency predictors for this dataset. We can also point out that models which process uniquely local feature conspicuity scored lower on SID4VAM fixation predictions, whereas the ones that processed global conspicuity scored higher. This phenomena might be related with the distinction of foveal (near the fovea) and ambient (away from the fovea) fixations, relative to the fixation order and the spatial locations of fixations  \cite{Eisenberg2016}. The evaluation of gaze-wise model predictions has been done by grouping fixations of every instance separately. We have plotted results of the $sAUC$ saliency metric for each model (\hyperref[fig:results_auc_gazewise]{Fig. \ref*{fig:results_auc_gazewise}}) and it is observable that model performance decrease upon fixation number, meaning that saliency is more likely to be predicted during first fixations. For evaluating the temporal relationship between human and model performance ($sAUC$), we have performed Spearman's ($\rho$) correlation tests for each fixation and it can be observed that IKN, ICL, GBVS, QDCT and ML-Net follow a similar slope as the GT, contrary to the case of the baseline center gaussian. 

\subsection{Model results on psychophysical consistency }

\begin{figure}[!]
\centering
\textbf{a)}\includegraphics[width=1\linewidth]{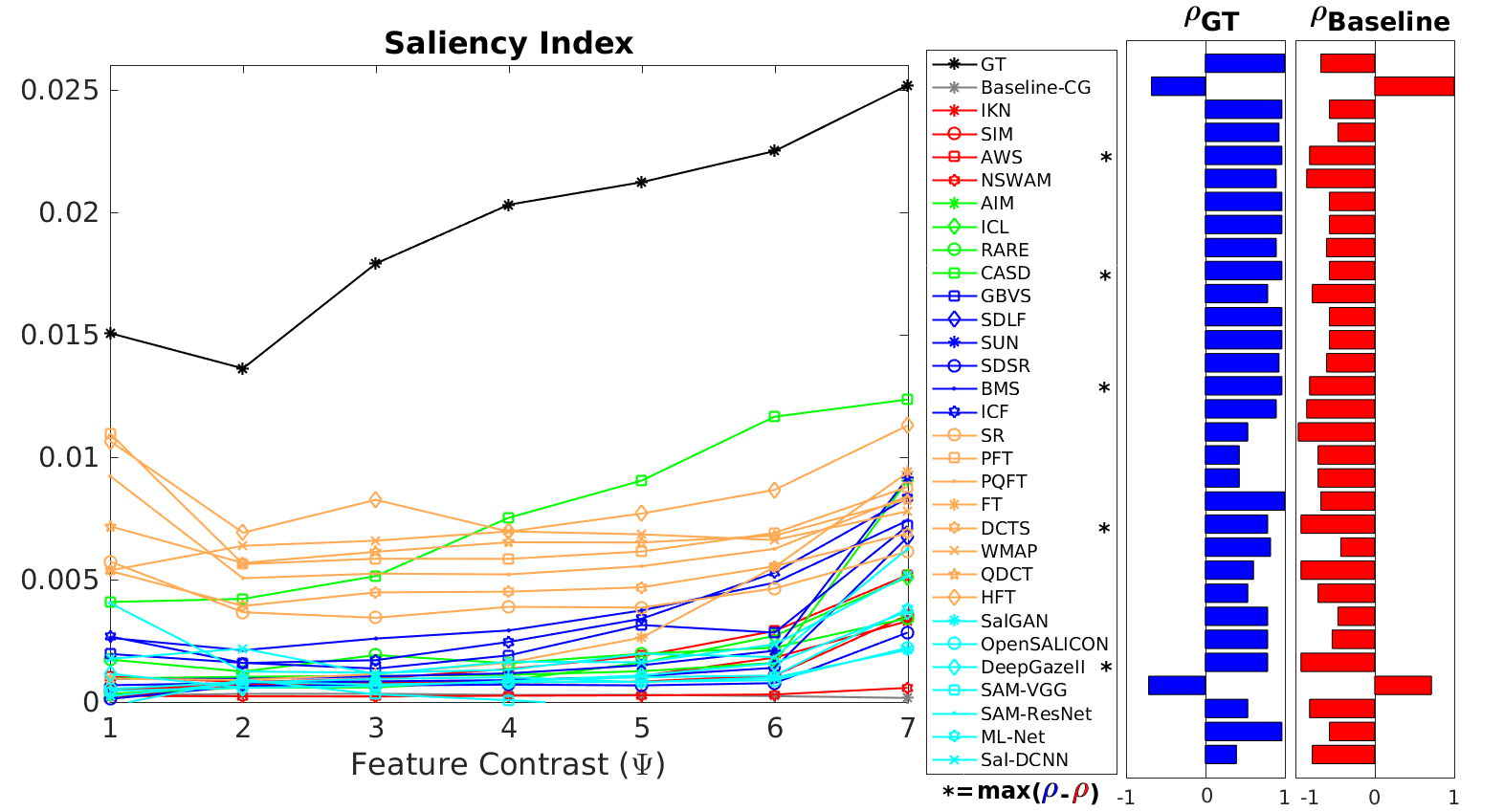}\\
\textbf{b)}\includegraphics[width=0.8\linewidth]{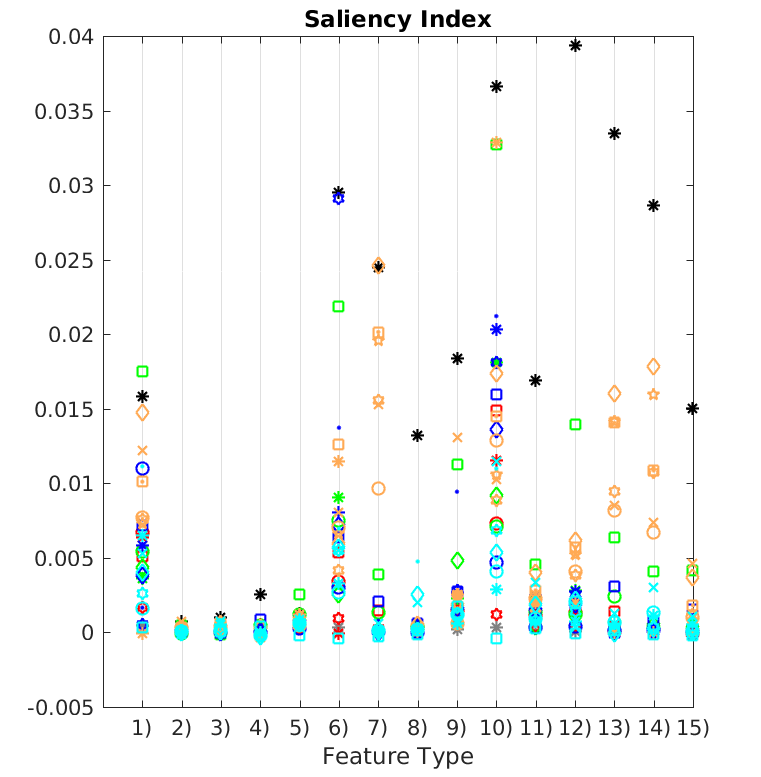}
\caption{Results of Saliency Index of model predictions upon Feature Contrast \textbf{(a)} and Feature Type \textbf{(b)}.}
\label{fig:results_sindex_contrast}
\end{figure}

Previous studies \cite{Borji2013c,Bruce2015,Berga2018a} found that several factors such as feature type, feature contrast, task, temporality of fixations and the center bias alternatively contribute to eye movement guidance. The HVS has specific contrast sensitivity to each stimulus feature, so that saliency models should adapt in the same way in order to be plausible in psychometric parameters. Here we will show how saliency prediction varies significantly upon feature contrast and the type of low-level features found in images. In \hyperref[fig:results_sindex_contrast]{Fig. \ref*{fig:results_sindex_contrast}a} is found that saliency models increase SI with feature contrast ``$\Psi$" following the distribution of human fixations. Most prediction SI scores show a higher slope with easy targets (salient objects with higher contrast with respect the rest, when $\Psi>4$), being CASD and HFT the models that have higher SI at higher contrasts.

\begin{figure*}[!]
\centering
\begin{subfigure}[c]{0.3\linewidth}
\centering
\includegraphics[width=1\linewidth]{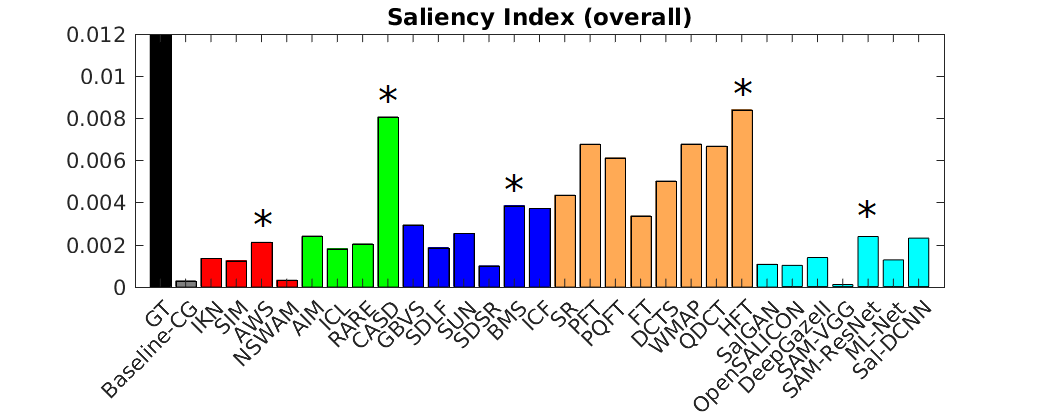}
\caption*{\textbf{(a)}}
\end{subfigure}
\begin{subfigure}[c]{0.3\linewidth}
\centering
\includegraphics[width=1\linewidth]{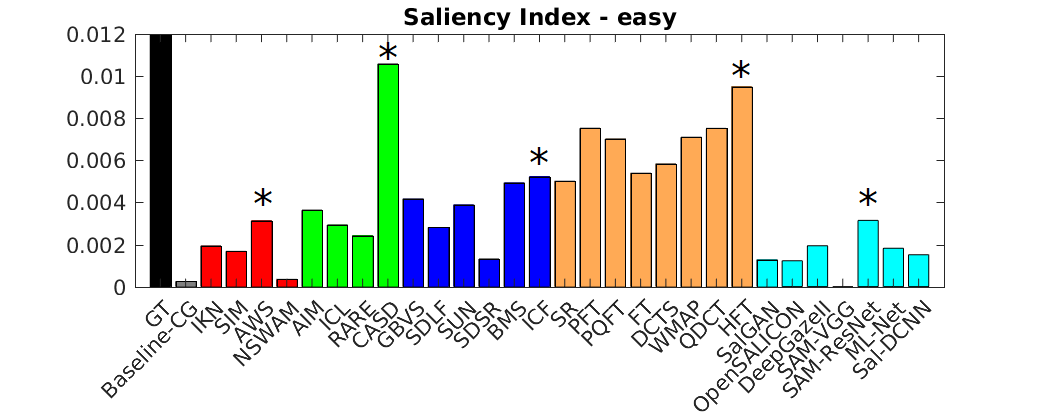}
\caption*{\textbf{(b)}}
\end{subfigure}
\begin{subfigure}[c]{0.3\linewidth}
\centering
\includegraphics[width=1\linewidth]{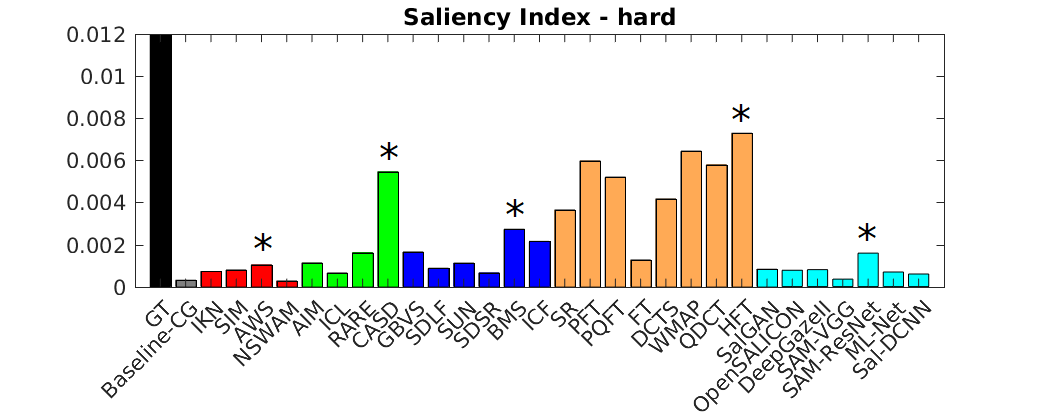}
\caption*{\textbf{(c)}}
\end{subfigure}
\makebox[0.3\linewidth]{ }
\begin{subfigure}[c]{0.3\linewidth}
\centering
\includegraphics[width=1\linewidth]{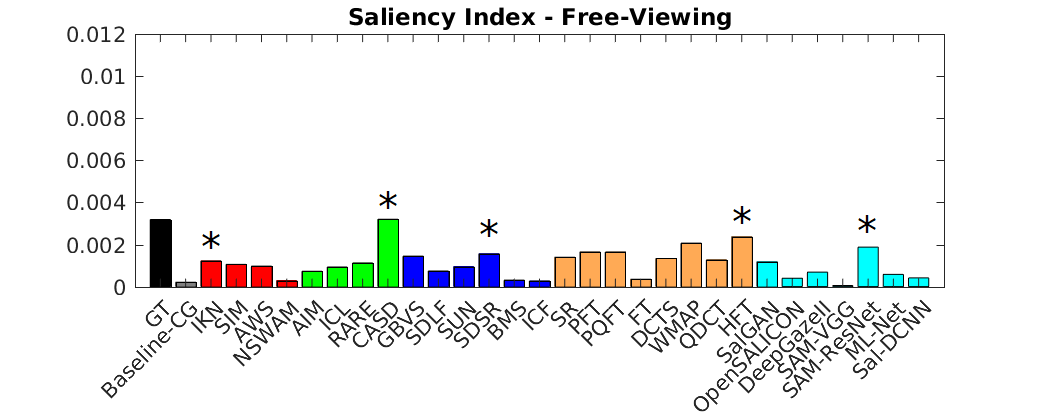}
\caption*{\textbf{(d)}}
\end{subfigure}
\begin{subfigure}[c]{0.3\linewidth}
\centering
\includegraphics[width=1\linewidth]{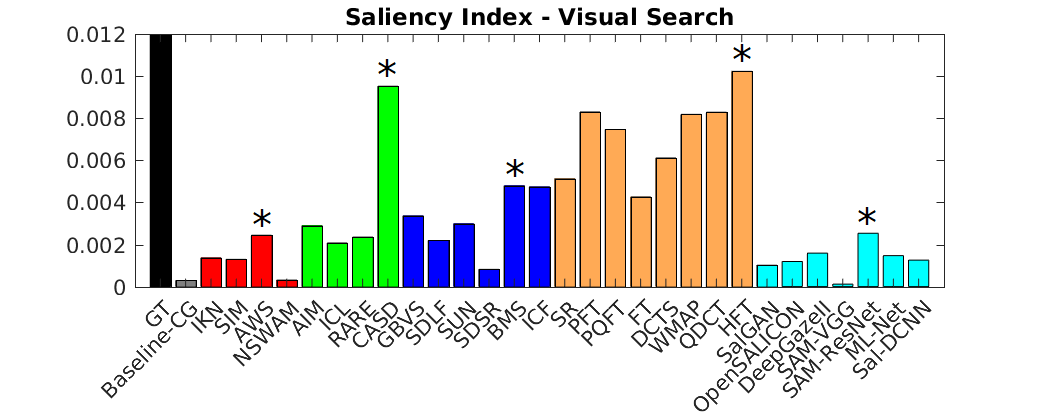}
\caption*{\textbf{(e)}}
\end{subfigure}
\caption{Results of Saliency Index metric scores from dataset model predictions \textbf{(a)}, for easy/hard difficulties \textbf{(b-c)} and Free-Viewing/Visual Search tasks \textbf{(d-e)}.}
\label{fig:results_sindex}
\end{figure*}

Contextual influences (here represented as distinct low-level features that appear in the image) contribute distinctively on saliency induced from objects that appear on the scene \cite{Hwang2011}. We suggest that not only the semantic content that appears on the scene affects saliency but the feature characteristics do significantly impact how salient objects are. This phenomena is observable in \hyperref[fig:results_sindex_contrast]{Fig. \ref*{fig:results_sindex_contrast}b} and occurs for both human fixations and model predictions, specifically with highest SI for human fixations in 1) Corner Salience, 6) Feature and Conjunctive Search, 7) Search Asymmetries, 10) Brightness Search, 12) Dissimilar Size Search and 13) Orientation Search with Heterogeneous distractors. HFT and CASD have highest SI when GT is higher (when human fixations are more probable to fall inside the AOI), even outperforming GT probabilities for the cases of 1) and 7). We show in \hyperref[fig:results_sindex]{Fig. \ref*{fig:results_sindex}a} that overall Saliency Index of most saliency models is distinct when we vary the type of feature contrast (easy vs hard) and the performed stimulus task (free-viewing vs visual search). Spectral/Fourier based models outperform other saliency models also in SI metric. Similarly with saliency metrics shown on previous subsection, AWS, CASD, BMS, HFT and SAM-ResNet are the most efficient models for each model inspiration category respectively. It is observable in \hyperref[fig:results_sindex]{Fig. \ref*{fig:results_sindex}b} that saliency models have higher performance for easy targets, with increased overall model performance differences with respect hard targets (\hyperref[fig:results_sindex]{Fig. \ref*{fig:results_sindex}c}). Similarly, visual search targets show lower difficulty (higher SI) to find predicted fixations inside the AOI than the free-viewing cases (\hyperref[fig:results_sindex]{Fig. \ref*{fig:results_sindex}d-e}). Also distinct SI curves upon feature contrast are reported, revealing that contrast sensitivies are distinct for each low-level feature. Spearman's correlation tests on \hyperref[fig:results_sindex_contrast]{Fig. \ref*{fig:results_sindex_contrast}b} show which models correlate with human performance over feature contrast and which one do so with the baseline (designating higher center biases). These results show that models such as AWS, CASD, BMS, DCTS or DeepGazeII highly correlate with human contrast sensitivities and do not correlate with the baseline center gaussian. Matching human contrast sensitivities on low-level visual features would be an interesting point of view to make future saliency models accurately predict saliency as well as to better understand how the HVS processes visual scenes. 

\section{SIG4VAM: Generating synthetic image patterns for training saliency models}

\begin{figure}
    \centering
    \begin{subfigure}{.30\linewidth}
        \includegraphics[width=\linewidth]{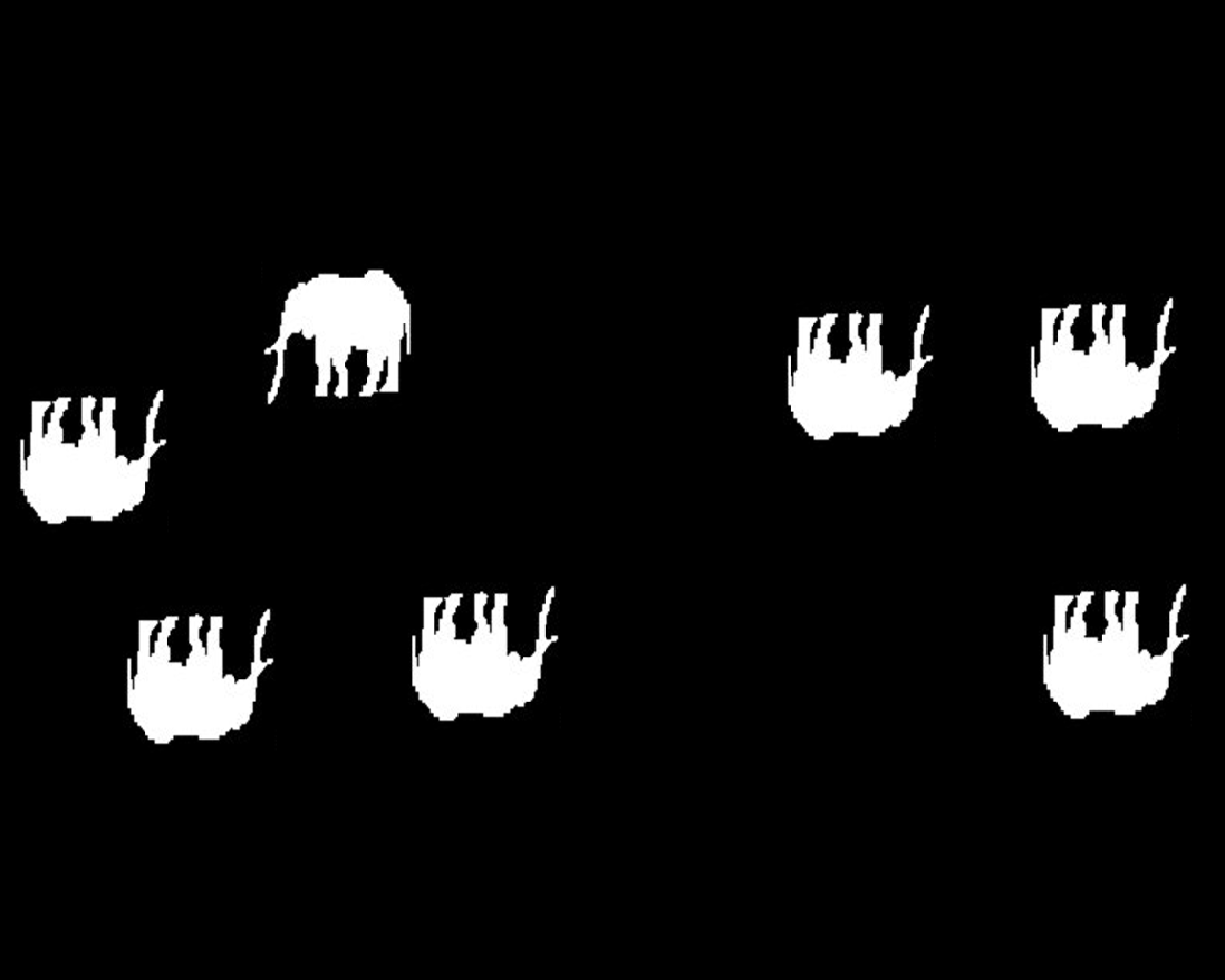}
    \end{subfigure}
    \begin{subfigure}{.30\linewidth}
        \includegraphics[width=\linewidth]{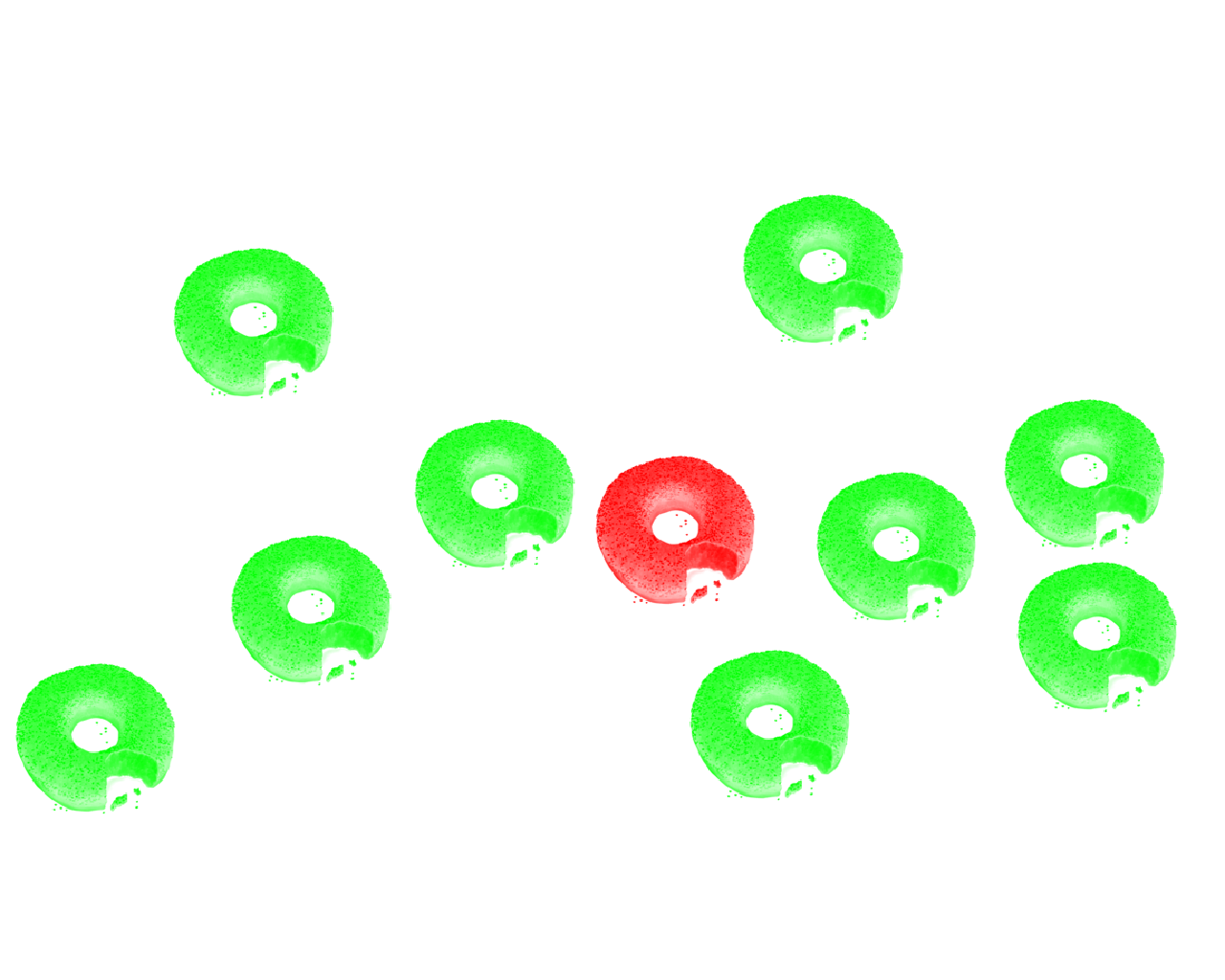}
    \end{subfigure}
    \begin{subfigure}{.30\linewidth}
        \includegraphics[width=\linewidth]{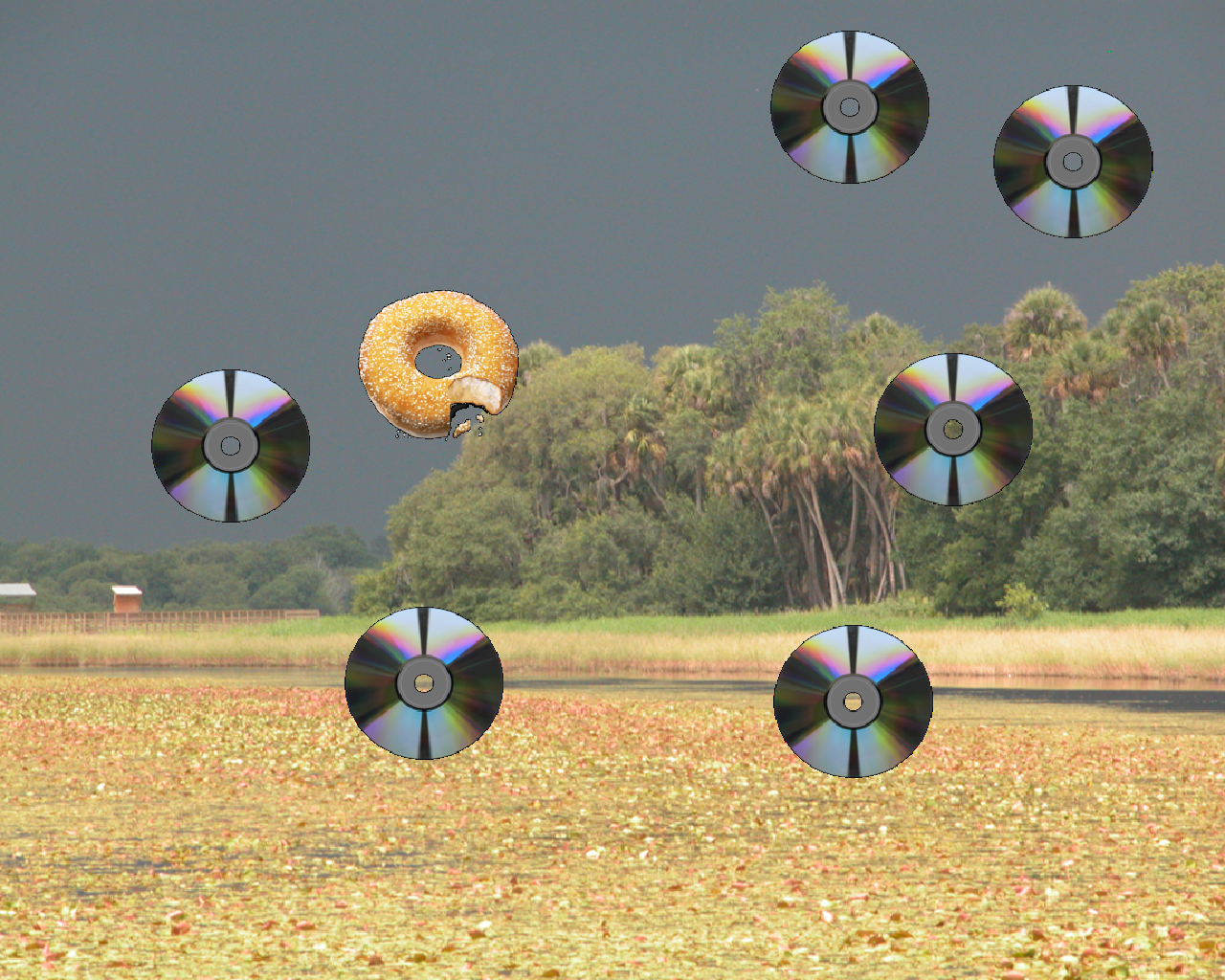}
    \end{subfigure}
    \caption{Examples of generating synthetic images with high-level features (i.e. objects as target/distractors), changing low-level feature properties \textbf{(a-b)} or background \textbf{(c)}.}
    \label{fig:SIG4VAM}
\end{figure}

We have also provided a synthetic image generator (SIG4VAM)\footnote{Code for generating synthetic stimuli: \url{https://github.com/dberga/sig4vam}}, able to generate similar psychophysical images with other types of patterns. A larger set of images can be created by parametrizing factors such as stimulus size, number of distractors, feature contrast, etc. For instance, if the same 15 types (33 subtypes) of stimuli are selected instead with 28 contrast ($\Psi$) instances, then is generated a dataset with $33\times28=924$ stimulus. Adding to that, synthetic images with high-level features can be created using SIG4VAM (\hyperref[fig:SIG4VAM]{Fig. \ref*{fig:SIG4VAM}}), by changing background properties, setting specific object instances for targets/distractors, as well as their low-level properties (orientation, brightness, color, etc.). SID4VAM has been proposed as a possible initial test set for saliency prediction, where data of fixations and binary masks are available for benchmarking. Training sets can be obtained with SIG4VAM (GT of binary masks of pop-out/salient regions are automatically-generated), abling to fit contrast sensitivies and obtaining loss functions upon scores of fixation probability distribution \cite{Bylinskii2018} and salient region detection metrics \cite{Wang2019} (e.g. SI, PR, MAE, S-/F-measures, etc.). Latest strategies \cite{Rosenfeld2018} that synthetically modify real scenes have shown dramatic changes in scores of object detection tasks, using ``object transplanting" (superposing an object on distinct locations on the scene). In these terms, SIG4VAM could be extended for evaluating predictions of models over distinct contexts and tasks.

\section{Discussion}


Previous saliency benchmarks show that eye movements are efficiently predicted with latest Deep Learning saliency models. This is not the case with synthetic images, also for models pre-trained with sets of psychophysical patterns (e.g. SAM with CAT2000). This suggest that their computations of saliency do not arise as a general mechanism. These methods have been trained with eye tracking data (real images containing high-level features) and although several factors guide eye movements have been shown \cite{Wolfe2010a} that low-level saliency (i.e. pop-out effects) is one of the most influential for determining bottom-up attention. Another possibility is that we randomly parametrized salient object location, lowering the center bias effect. With this benchmark we can evaluate how salient is a particular object by parametrizing its low-level feature contrast with respect to the rest of distractors and/or background. Therefore, the evaluation of saliency can be done accounting for feature contrast, analyzing the importance to the objects that are easier to detect or preattetively. Previous saliency benchmarks usually evaluate eye tracking data spatially across all fixations, we also propose the evaluation of saliency across fixations, which is an issue of further study. Future steps for this study would include the evaluation of saliency in dynamic scenes \cite{Riche2016b,Leboran2017} using synthetic videos with both static or dynamic camera. This would allow us to investigate the impact of temporally-variant features (e.g. flicker and motion) over saliency predictions. Another analysis to consider is the impact of the spatial location of salient features (in eccentricity terms towards the image center), which might affect each model distinctively. Each of the steps in saliency modelization (i.e. feature extraction, conspicuity computation and feature fusion) might have a distinct influence over eye movement predictions. Acknowledging that conspicuity computations are the key factor for computing saliency, a future evaluation of how each mechanism contributes to model performance might be of interest. 

%

\section{Conclusion}

Contrary to the current state-of-the-art, we reveal that saliency models are far away from acquiring HVS performance in terms of predicting bottom-up attention. We prove this with a novel dataset SID4VAM, which contains uniquely synthetic images, generated with specific low-level feature contrasts. In this study, we show that overall Spectral/Fourier-based saliency models (i.e. HFT and WMAP) clearly outperform other saliency models when detecting a salient region with a particular conspicuous object. Other models such as AWS, CASD, GBVS and SAM-ResNet are the best predictor candidates for each saliency model inspiration categories respectively (Cognitive/Biological, Information-Theoretic, Probabilistic and Deep Learning). In particular, visual features learned with deep learning models might not be suitable for efficiently predicting saliency using psychophysical images. Here we pose that saliency detection might not be directly related to object detection, therefore training upon high-level object features might not be significatively favorable for predicting saliency in these terms. Future saliency modelization and evaluation should account for low-level feature distinctiveness in order to accurately model bottom-up attention. Here we remark the need for analyzing other factors such as the order of fixations, the influences of the task and the psychometric parameters of the salient regions.

\section{Acknowledgements}

This work was funded by the MINECO (DPI2017-89867-C2-1-R, TIN2015-71130-REDT), AGAUR (2017-SGR-649), CERCA Programme / Generalitat de Catalunya, in part by Xunta de Galicia under Project ED431C2017/69, in part by the Conseller\'ia de Cultura, Educaci\'on e Ordenaci\'on Universitaria (accreditation 2016–2019, ED431G/08) and the European Regional Development Fund, and in part by Xunta de Galicia and the European Union (European Social Fund). We also acknowledge the generous GPU support from NVIDIA.

{\small
\bibliographystyle{iccv_format/ieee_fullname}
\bibliography{main}
}

\end{document}